\documentclass[sigconf,screen]{acmart} 
\usepackage{subcaption}
\usepackage{longtable}
\usepackage{multicol}

\begin{document}

\title{ReLI: Cross-Lingual Language-to-Action Grounding for Human–Robot Interaction}

\author{Linus Nwankwo}
\authornote{Corresponding author.}
\orcid{0000-0002-1767-2140}
\correspondingauthor
\affiliation{%
  \institution{Chair of Cyber-Physical Systems, Montanuniversität}
  \city{Leoben}
  \country{Austria}
}

\author{Bjoern Ellensohn}
\affiliation{%
 \institution{Chair of Cyber-Physical Systems, Montanuniversität}
  \city{Leoben}
  \country{Austria}
  }

\author{Ozan Özdenizci}
\affiliation{%
  \institution{Institute of Machine Learning and Neural Computation, Graz University
of Technology}
  \city{Graz}
  \country{Austria}
}

\author{Elmar Rueckert}
\affiliation{%
 \institution{Chair of Cyber-Physical Systems, Montanuniversität}
  \city{Leoben}
  \country{Austria}
  }

\renewcommand{\shortauthors}{Nwankwo et al.}

\begin{abstract}
  Adapting autonomous agents for real-world industrial, domestic, and other daily tasks is currently gaining momentum. However, in global or cross-lingual application contexts, the ability to instruct these agents in one's native language remains until today a formidable challenge. Existing language-conditioned human-robot interaction frameworks typically support only a handful of high-resource languages, e.g., English and Chinese, limiting accessibility for billions of potential end users. To address this gap, we propose ReLI, a cross-lingual framework that enables autonomous agents to converse naturally, reason semantically about their environment, and execute downstream tasks, regardless of the tasks' instruction linguistic origin or input modalities. First, we ground large-scale pre-trained foundation models and transform them into language-to-action models that can directly provide common-sense reasoning and high-level robot control through free-form conversational interactions. We then perform an implicit language-conditioned cross-lingual adaptation of the models to ensure that ReLI generalises effectively across diverse global languages. To demonstrate ReLI's robustness, we conducted extensive empirical evaluation on a diverse set of short- and long-horizon tasks, including zero-shot and few-shot spatial navigation, scene information retrieval, and query-oriented tasks, and then benchmarked the performance across more than $70K+$ multi-turn conversations in over $140$ languages spanning high-resource, low-resource, and vulnerable/creole tiers. Across the benchmarked languages, ReLI achieved consistently high instruction-parsing accuracy, task success rate, and rapid response time. Further, we conducted a user study on ReLI with $34$ human raters interacting in $13$ native languages, and they reported high perceived naturalness and responsiveness, with no self-reported language-induced performance gap in their interaction. These results demonstrate ReLI's potential to advance and champion inclusive human–robot interaction in linguistically diverse real-world settings. Demos and resources:~\url{https://linusnep.github.io/ReLI/}.
\end{abstract}

\maketitle

\section{Introduction}
    Nowadays, physical autonomous agents, such as robots, are increasingly being deployed for real-world tasks, including industrial inspection, domestic chores, and other daily tasks, where they are expected to interact naturally with both expert and non-expert end users. In these settings, users can provide task instructions in different languages, dialects, scripts, and modalities. Effective human-robot interaction (HRI) therefore requires the agents to interpret these instructions and ground them in perceptual observations, spatial context, the agent's internal state, and feasible control actions. As the tasks presented to these agents grow more complex and the deployment environments more unpredictable and linguistically diverse, robust cross-lingual HRI frameworks become increasingly important~\cite{8307144,slade2024human}.

    Until today, language has posed a substantial barrier to truly universal and realistic natural human-robot collaboration, particularly for speakers of under-resourced languages and users in unconstrained everyday environments~\cite{nwankwo2024conversation,gonen2025socially,marge2022spoken}. Of roughly $7,000$ living languages worldwide~\cite{campbell2008ethnologue}, current language-conditioned HRI frameworks~\cite{lynch2023interactive,brohan2023can, nwankwo2024conversation,allgeuer2024robots} and physical robot platforms have remained constrained in high-resource languages, such as English, Chinese, and a handful of Indo-European languages. 
    This unilateral, language-specific training restricts equitable access to robotic systems and to natural interaction for billions of potential robot users and communities whose languages remain under-represented in contemporary AI resources. Therefore, to preserve linguistic diversity and foster inclusive and accessible HRI in the real world, enabling autonomous agents to converse across multiple languages is essential.
    
\begin{figure*}[htp]
    \centering
    \includegraphics[width=1.0\linewidth]{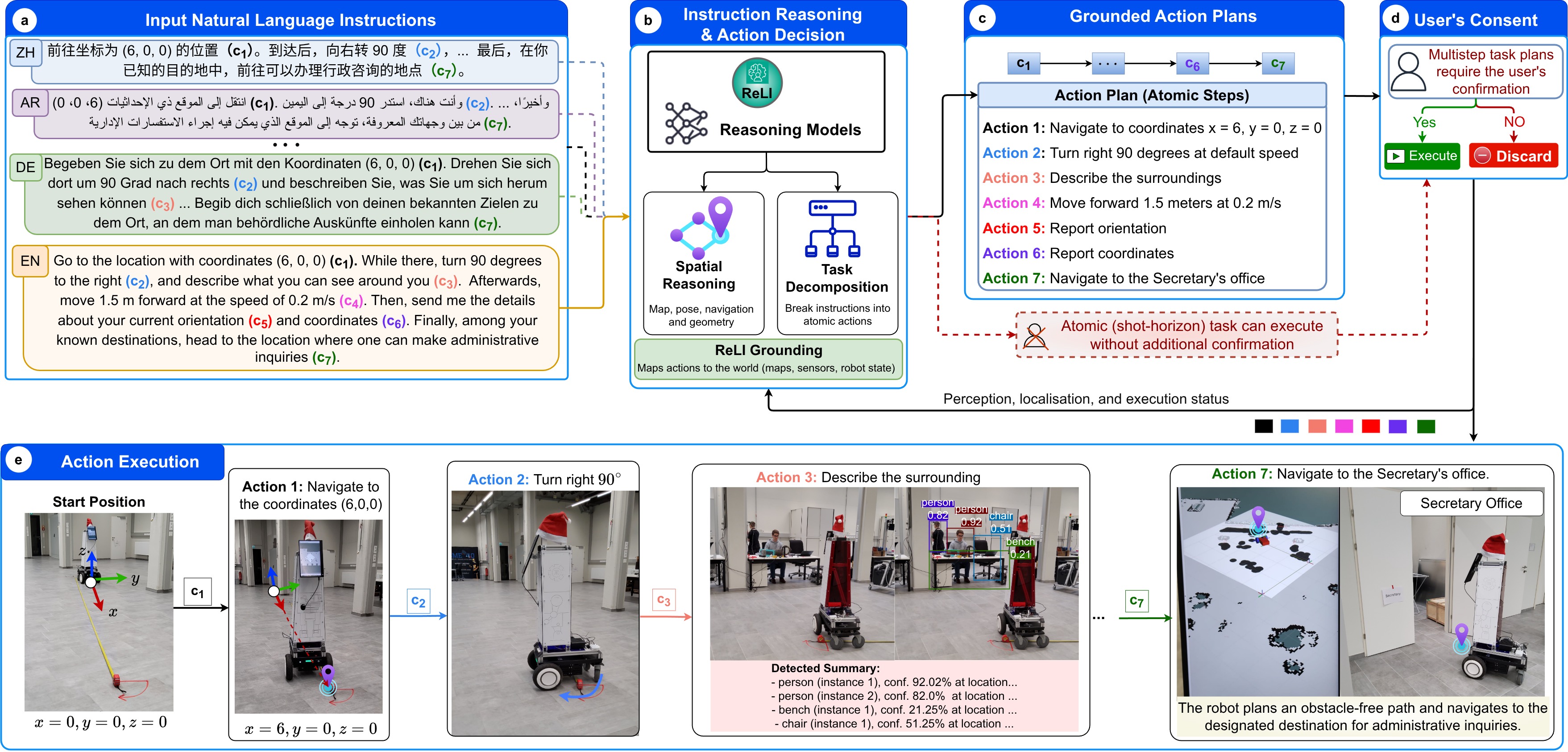}
    \small \caption{Overview of ReLI for short- and long-horizon task execution. (a) A natural-language instruction $c \in \mathcal{C}_T$, composed of sub-instructions $(c_1,\ldots,c_7)$, is provided in a language $\ell \in \mathcal{L}$. (b) ReLI reasons over the task instruction and grounds it in the available spatial, perceptual, and task context. (c) It generates an ordered grounded action plan, $\pi = (a_1,\ldots,a_7)$, where each action corresponds to a sub-instruction. (d) For multistep executable plans, ReLI requests user approval before dispatching the action plans to the robot controller. (e) Approved plans are executed; otherwise, they are discarded. See Section~\ref{sec:method} for the details.}
    \label{fig:ReLIexample}
\end{figure*}
    The HRI community has undoubtedly made substantial progress towards natural language-conditioned robotics~\cite{brohan2023can,lynch2023interactive, brohan2022rt, team2025gemini, nwankwo2024multimodalhumanautonomousagentsinteraction}. With the current progress, today's autonomous agents can interpret natural-language instructions, reason over perceptual observations, and execute increasingly complex tasks.
    However, to our knowledge, a robust framework that simultaneously supports physical agents in natural multimodal interaction, physical-environment grounding, and downstream task execution regardless of languages and input conversation modalities remains underexplored.

    Motivated by these limitations, we propose~\textbf{Re}gardless of the \textbf{L}anguage of task~\textbf{I}nstructions (\textbf{ReLI}), a cross-lingual language-to-action framework for embodied agents. ReLI pushes the boundaries of language-conditioned HRI across users with diverse linguistic backgrounds and levels of technical expertise, including under-resourced languages, endangered languages, creoles, and vernaculars such as African Pidgin and Cherokee.
    To achieve these novel objectives, we extensively exploit the inherent cross-lingual generalisation capabilities 
    of large-scale pre-trained foundation models, e.g., GPT models~\cite {openai2023reasoning,singh2025openai}, to capture semantic and syntactic aspects across languages without explicit language-specific supervision, data collection, or model retraining.
    We employed the models off-the-shelf to avoid the additional computational and data demands associated with task-specific fine-tuning, and to mitigate the risks of catastrophic forgetting~\cite{zhai2024investigating}, common with fine-tuned models, where the model loses general knowledge or capabilities in favour of task-specific retraining.
    Fig.~\ref{fig:ReLIexample} illustrates how ReLI can empower physical agents to abstract human-specified instructions into grounded action plans for both short- and long-horizon tasks. 
    The key contributions are:
\begin{itemize}
    \item \textbf{Cross-lingual language-to-action grounding for HRI.} We introduce ReLI, a robust framework to drive inclusivity and diversity in real-world human-robot interactions and task collaborations. Unlike the existing approaches that either depend on code-level methods~\cite{liang2023code} or on unilingual high-resource languages \cite{lynch2023interactive, brohan2023can, shah2023lm}, ReLI is, to the best of our knowledge, the first language-conditioned HRI framework to abstract natural free-form human instructions into robot-actionable commands, across diverse languages, input instruction modalities, technical expertise, and without task-specific retraining of the orchestrating models.
    \item \textbf{Extensive evaluation across embodied tasks.} We evaluate ReLI in both real-world and simulated environments on several short- and long-horizon tasks, including zero- and few-shot instruction following, open-vocabulary object navigation, spatial navigation, scene-information retrieval, and query-oriented reasoning.
    \item \textbf{Large-scale multilingual benchmark.} We benchmarked ReLI's multilingual instruction parsing accuracy on over $140$ living languages drawn from across the continents, involving over $70$K multi-turn conversations. Across all benchmarked languages, ReLI achieved, on average, $90\%\pm0.2$ accuracy in multilingual instruction parsing and task execution success rates. These results provide strong empirical evidence for ReLI and demonstrate the feasibility of cross-lingual language-to-action grounding across a broad range of linguistic conditions.
    \item \textbf{Human-rater study and open-source data.}
    We conducted a user study with $34$ raters interacting in $13$ native languages, assessing perceived responsiveness, naturalness, and language-induced interaction gaps in real-world deployment.
    We release our evaluation data, experimental protocols, and model configurations to facilitate reproducibility and further extensions of the ReLI framework.
\end{itemize}

\section{Background and Related Work}
    The last few years have witnessed tremendous advancement in generative AI~\cite{10109305, doi:10.1177/02666669231200628} and natural language processing (NLP)~\cite{Chowdhary2020, JUST2024102883, 10.1007/978-981-15-9712-1-31, doi:10.1126/science.aaa8685}. This surge, largely driven by large language models (LLMs)~\cite{openai2023reasoning, liu2024deepseek, team2023gemini, Touvron2023LLaMAOA} and vision-language models (VLMs)~\cite{Radford2021LearningTV, li2022grounded}, has substantially changed how intelligent systems interpret, reason over, and respond to human instructions~\cite{xi2025rise, 10433480, ZHANG2023100131}. LLMs, trained on extensive corpora sourced from the web~\cite{10.5555/3666122.3669586}, are typically autoregressive transformer-based architectures~\cite{10.5555/3295222.3295349, NEURIPS2023_3eb7ca52}. In principle, given an input command sequence, \(c = (c_1, c_2, \dots, c_T) \in \mathcal{C}_T\), where $\mathcal{C}_T$ represents the space of all possible user commands, these models predict the corresponding output sequence \(y = (y_1, y_2, \dots, y_T) \in \mathcal{Y}_T\) by factorising the conditional sequence probability as:
\begin{equation}\label{eqn_autoreg}
    \begin{split}
        p_{\theta}(y_1, y_2, \dots, y_T \mid c) = p_{\theta}(y_1 \mid c)\cdot p_{\theta}(y_2 \mid y_1,\; c) \dots\,\\ p_{\theta}(y_T \mid y_{1:T-1},\; c) = \prod_{t=1}^T p_{\theta}(y_t \mid y_{1:t-1},\; c),
    \end{split}
\end{equation}
    where $\mathcal{Y}_T$ is the space of all possible output sequences of sequence length $T$, $\theta$ denotes the learned model parameters, and $y_{<t}$ represents the previously generated tokens. The formulation in Eq.~\eqref{eqn_autoreg} allows the model to produce context-dependent outputs conditioned on both the user instruction and the generated response.
   
    Although these LLMs were originally developed as general-purpose language processing systems~\cite{yang2024harnessing, 10.1145/3641289}, their reasoning, instruction-following, and multilingual capabilities have been extensively studied across diverse benchmarks~\cite{srivastava2022beyond, he2024multiifbenchmarkingllmsmultiturn,qin2024infobench, li2026xifbench}. Works such as~\citet{he2024multiifbenchmarkingllmsmultiturn, shilanguage, ahuja2023mega, lai2023chatgpt}, and~\citet{zhang2023m3exam} have shown that these models can achieve exceptional generalisation across languages, beyond the high-resource languages that traditionally dominate the natural language processing benchmarks~\cite{faisal2024dialectbenchnlpbenchmarkdialects, liang2020xgluenewbenchmarkdataset, costa2022no}. Thus, this cross-lingual prowess makes them compelling candidates for interaction in linguistically heterogeneous environments, where users may issue task instructions in different languages, scripts, dialects, or conversational modalities.
    
    On the other hand, VLMs~\cite{Radford2021LearningTV, li2022grounded} pre-trained on large-scale image-text pairs have emerged as a groundbreaking approach to integrate visual observations and natural-language descriptions. These models learn joint or comparable visual-textual representations, often through contrastive learning objectives~\cite{NEURIPS2020_d89a66c7, Radford2021LearningTV}, and can support open-vocabulary recognition, visual question answering, object grounding, and scene-level semantic reasoning. For robotics, this capability is important, as a user instruction rarely maps directly to a control command. Instead, the robot must abstract linguistic references to perceived objects, spatial relations, environment context, and feasible robot actions. Thus, the integration of LLMs and VLMs provides a powerful computational substrate for transforming natural-language instructions into situated robot actions.
    
\subsection{Language-conditioned Robotics and HRI}
    Grounding natural language into robot action has become a central direction in language-conditioned robotics and HRI~\cite{nwankwo2026sil,10182264, team2025gemini}. In this context, an agent must abstract human instructions into task-relevant entities, action primitives, parameters, and execution constraints. Recent advances in LLMs and VLMs have broadened the scope of this abstraction from predefined command grammars to unconstrained, free-form natural language instructions, enabling robots to interpret goals, decompose tasks, query the environment, and generate executable behaviours~\cite{wei2022chain, kojima2022large, brohan2023can, nwankwo2024multimodalhumanautonomousagentsinteraction}.
    Translating from natural language to real-world action is the most common form of language-to-action reasoning in embodied robotic systems~\cite{10182264, 10160396, team2025gemini}. For example, several approaches~\cite{10610434, brohan2023can, nwankwo2024conversation, Zeng2023DemonstratingLL, pmlr-v229-zitkovich23a, liang2023code,huang2022inner, zitkovich2023rt,kim2024openvla,nwankwo2026sil} have demonstrated that, with VLMs and LLMs combined, robots can perceive their surroundings, perform multi-step reasoning, and execute long-horizon tasks specified in unconstrained natural language in a manner akin to human-like reasoning. Together, these works show that LLMs and VLMs provide a flexible interface between high-level user goals and low-level robot execution. They also show that natural language can serve not only as a communication modality, but also as an operational specification from which task plans and concrete action sequences can be derived. However, despite these advances, grounding these models in multilingual robotic capabilities remains an open challenge.
    
    To date, most language-instructible~\cite{brohan2023can, nwankwo2026sil,10802251, nwankwo2024conversation} and vision-language-conditioned HRI frameworks~\cite{10610090, kim2024openvla, li2024visionlanguagefoundationmodelseffective, yuan2024robopoint, zitkovich2023rt} focused mainly on grounding unilingual task instructions or a limited set of high-resource languages~\cite{ku-etal-2020-room}. They are not designed for cross-lingual instructions or complex tasks specified in diverse languages, and thus struggle to translate such commands into robotic actions. Despite the strong real-world performance, their lack of multilingual capability limits deployment in cross-linguistic operational domains. In this work, we address these challenges with a novel natural language-driven approach that combines the inherent strengths of both language and visual foundation models to realise a more inclusive approach to human-agent interaction, one in which, regardless of the modality of the conversation or the language of the task instruction, the conversation is the robot’s executable commands.

\subsection{Multilingual Language-to-Action Grounding for Embodied Agents}
    Multilingual large language models and machine translation frameworks~\cite{wu2016google,johnson2017google,costa2022no,shilanguage} have demonstrated substantial cross-lingual transfer capabilities across a wide range of NLP tasks. These frameworks can represent and process textual or audio information from many languages within a shared modelling space, enabling knowledge transfer from high-resource to low-resource languages. However, in isolation, multilingual linguistic competence is insufficient to guarantee effective performance in embodied robotic tasks. For autonomous robotic systems, natural language instruction must be grounded in task-relevant entities, spatial relations, continuous and discrete parameters, executable action sequences, and dynamically changing environmental states. 
    
    Despite recent progress, research on multilingual embodied agents remains relatively limited. For instance, the work Room-Across-Room~\cite{ku-etal-2020-room} offers multilingual vision-and-language navigation data in English, Hindi, and Telugu within simulated environments.
    While such benchmarks are valuable for analysing multilingual instruction following, they do not directly support open-ended spoken human–robot interaction (HRI), bidirectional user dialogue, language-conditioned scene querying, or low-level execution via physical robot-control interfaces. More generally, many embodied-agent benchmarks primarily emphasise navigation or simulation-constrained instruction following, whereas real-world HRI deployments must integrate language understanding, perception, state estimation, action generation, execution monitoring, and user-oriented response generation into a coherent framework.
    ReLI extends this research direction by explicitly addressing cross-lingual language-to-action grounding in both simulated and physical robot environments. The primary objective of ReLI is not multilingual translation per se, but the end-to-end transformation of task instructions into grounded robotic behaviours. This includes implicit language conditioning, interpretation of task semantics, resolution of visual and spatial references, synthesis of action primitives, execution of the corresponding robot behaviours, and communication of task outcomes in the user’s interaction language. 
 
\section{Method: ReLI}\label{sec:method}
    We addressed the problem of grounding multilingual, free-form and unconstrained human instructions into robot-executable actions. We postulate that, to drive inclusive HRI and equitable robot access beyond technically trained users, embodied autonomous agents must interpret task goals expressed across languages and modalities through a unified action interface that mirrors human-to-human cross-lingual communication. This section presents the formal details, and Fig.~\ref{fig:ReLIArchitect} shows the architectural overview of the ReLI framework.
    
\subsection{Problem Description: Monolingual Physical Agents in Multilingual Reality}
    State-of-the-art language-conditioned HRI frameworks~\cite{team2025gemini,brohan2023can, nwankwo2024conversation, Zeng2023DemonstratingLL, pmlr-v229-zitkovich23a, liang2023code,huang2022inner, zitkovich2023rt,kim2024openvla,nwankwo2026sil} predominantly employ monolingual grounding architectures, in which high-level user-instructible linguistic commands, expressed in human language are mapped to task plans or robot actions through language-specific assumptions, prompts, or command interfaces. This design limits deployment in multilingual environments, where users may express equivalent task goals using different languages, scripts, dialects, or input modalities. Formally, we consider a high-level user-instructible linguistic command $c \in \mathcal{C}_T$ expressed in language $\ell \in \mathcal{L}$ that is a priori unknown to the system. We assume that $\ell$ is generalisable by state-of-the-art large-scale pre-trained foundation models (e.g., GPT-models~\cite{openai2023reasoning}, Gemini~\cite{team2023gemini}, DeepSeek~\cite{liu2024deepseek}). We further assume access to high-dimensional sensory observations \(\mathcal{V}_s\) (e.g., synchronised RGB-D data, odometry) from the robot's onboard perception sensors that capture the visual, spatial, and geometric state of the task environment. Our primary objective, therefore, is to realise a grounding function \(\mathcal{F}_{\mathrm{ReLI}}: \mathcal{C}_T \times \mathcal{V}_s \mapsto \mathcal{A}\) which maps the command–observation pair $(c\,,\,\mathcal{V}_s)$ into a sequence of executable robot actions \(\mathcal{A}\). Critically, we require the resulting output \(\mathcal{F}_{\mathrm{ReLI}}(\cdot)\) to generalise across languages and input modalities, to allow task instructions to be interpreted and executed regardless of their linguistic origin, and without task-specific retraining of the orchestrating models.
    
\begin{figure*}[htp]
    \centering
    \includegraphics[width=\linewidth]{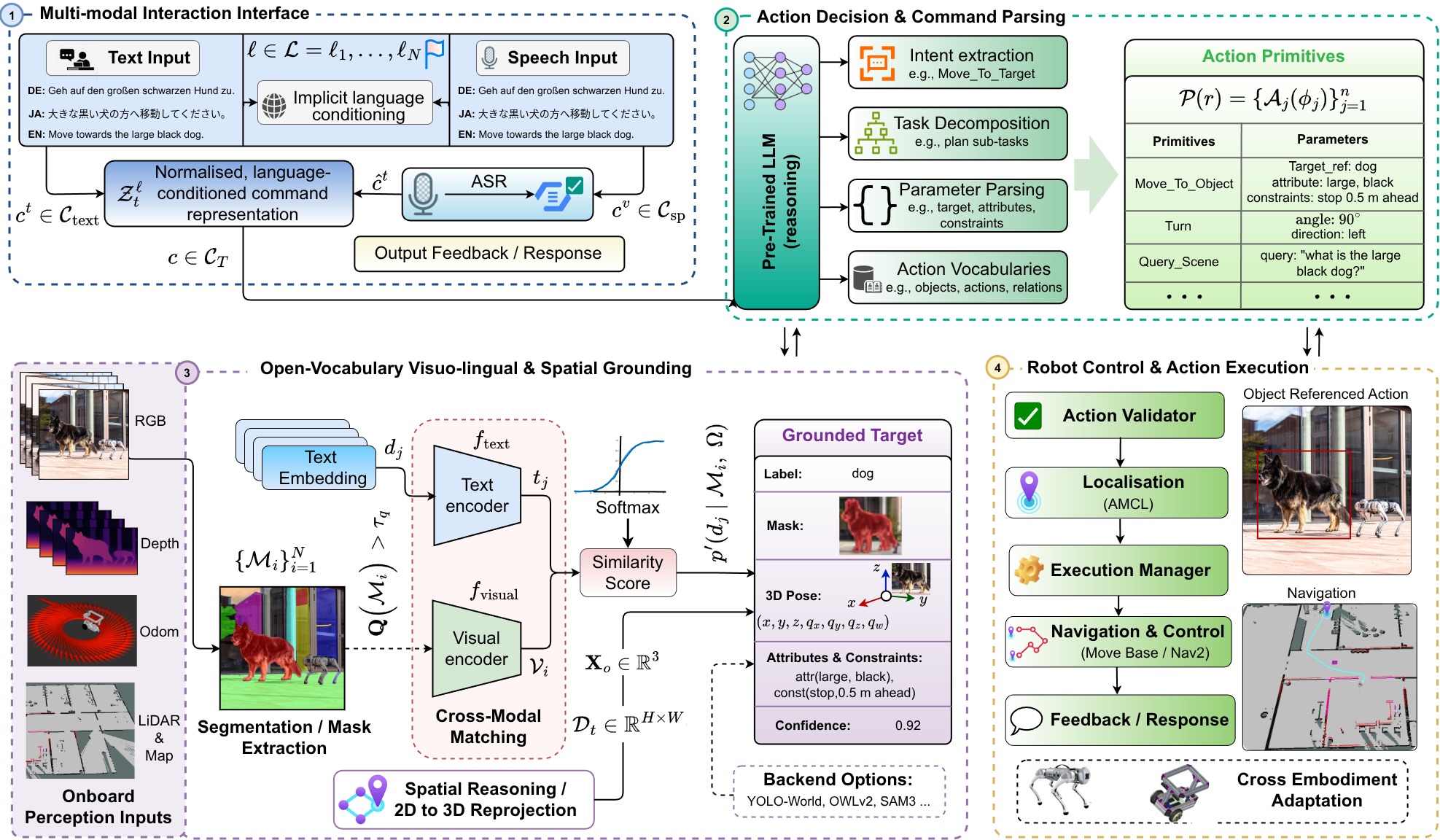}
    \caption{Overview of the ReLI framework for cross-lingual language-to-action grounding. (1) User instructions are received through speech or text and normalised into a language-conditioned command representation. (2) A pre-trained foundation model reasons over the instruction, extracts task intent, decomposes the instruction, and maps it to robot-action primitives. (3) The visuo-lingual grounding pipeline resolves object references and spatial relations via open-vocabulary perception, segmentation, RGB-D observations, odometry, and map information. (4) The grounded action representation is then passed to a backend-modular execution stack for navigation, control, and user feedback. 
    See Sections~\ref{subsec3A2}-\ref{subsec3D} for the formal details.}
    \label{fig:ReLIArchitect}
\end{figure*}
    To accomplish these novel objectives, we decomposed our approach into four architectural taxonomies based on the individual functions, as shown in Fig.~\ref{fig:ReLIArchitect}. First, we present the multimodal interaction interface, where the user's input modalities and task instructions are detected, processed, and transcribed (i.e., in the case of vocal or audio instructions $c^v$) into textual representations (Section~\ref{subsec3A2}). Second, we exploit the inherent cross-lingual capabilities of a large-scale pre-trained foundation model~\cite{openai2023reasoning} to reason over the high-level natural language instructions and parse them into robot-actionable commands (Section~\ref{subsec3B}). Third, we ground the linguistic and visual context of the agent's task environment through a contrastive language-image pre-training representation~\cite{Radford2021LearningTV}, alongside self-supervised open-vocabulary computer vision models~\cite{carion2025sam,cheng2024yolo} and geometric depth fusion~\cite{lin2025depth} (Section~\ref{subsec3C}). Finally, we abstract the high-level understanding from the decision and command parsing pipeline (Section~\ref{subsec3B}) into the physical robot actions through the action execution mechanism (Section~\ref{subsec3D}).

\subsection{Multimodal Interaction Interface}\label{subsec3A2}
\begin{figure*}[htp]
    \centering 
    \begin{subfigure}[t]{1.0\linewidth}
    \centering 
    \includegraphics[width=\linewidth]{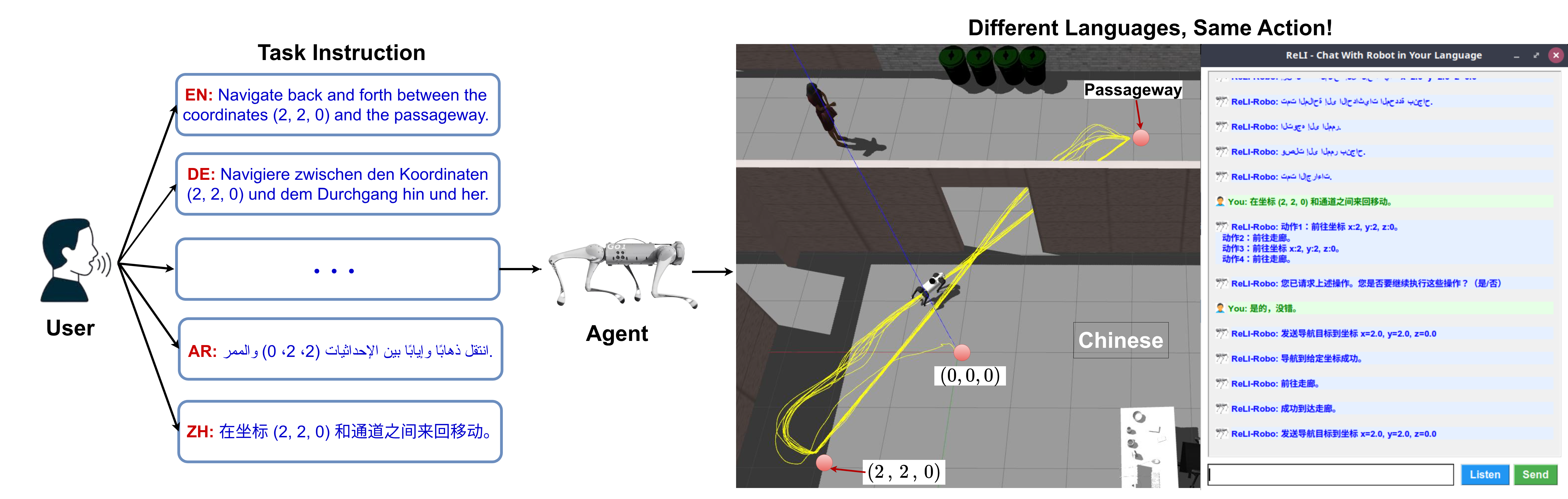}
    \caption{Label-free cross-lingual human-robot interaction.}
    \label{fig:interact-mechanism}
    \end{subfigure}
    \hfill
    \begin{subfigure}[t]{0.5\linewidth}
    \centering
    \includegraphics[width=\linewidth]{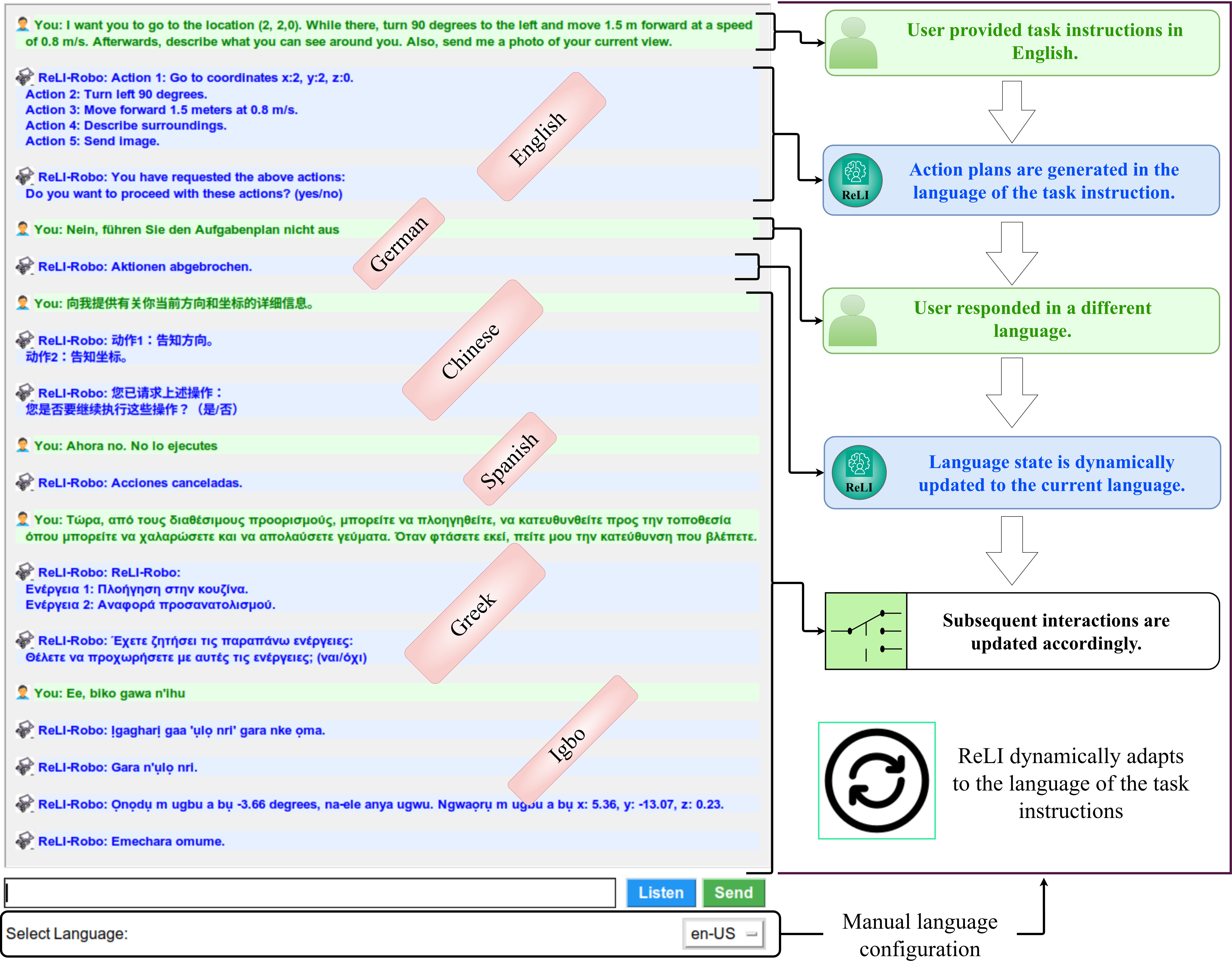}
    \caption{Event-driven multilingual interaction mechanism.}
    \label{fig:interact-interface}
    \end{subfigure}
    \caption{Multimodal interaction in ReLI. \textbf{(a)} Users can express semantically equivalent task instructions through different languages, scripts, modalities, or code-mixed utterances. ReLI normalises these inputs into a shared interaction-conditioned robot-action representation. \textbf{(b)} Interaction in ReLI is also event-driven. Each user input triggers command normalisation, action parsing, and the corresponding robot action in the user's input language.}
\label{fig:interact}
\end{figure*}

    The multimodal interaction interface, illustrated in Fig.~\ref{fig:ReLIArchitect} (top-left), serves as the entry point through which users provide task instructions to ReLI. We developed a lightweight chat-based interface (an example screenshot shown in Fig.~\ref{fig:interact}) to support bidirectional natural-language interaction between the user and the robot. User natural language instructions often arise through two primary input modalities: plain text, \(c^t \in \mathcal{C}_\text{text}\), and audio or vocal instructions, \(c^v \in \mathcal{C}_{\text{sp}}\). To process both modalities uniformly, we normalise each input into a text-based, language-conditioned command representation, \(\mathcal{Z}^{\ell}_t\), suitable for downstream linguistic reasoning, action parsing, and execution.
    \begin{equation}\label{eq:norm-rep}
\begin{split}
    \mathcal{Z}^{\ell}_t \,=\, (\bar{c}_t,\xi_t),\;\; \bar{c}_t \,=\,\Pi_{\text{norm}}(c) =
    \begin{cases}
        \Pi_{\mathrm{norm}}(c^t), & c_t \in \mathcal{C}_{\mathrm{text}},\\
        \Pi_{\mathrm{norm}}(\hat{c}^{t}), & \hat{c}^t =  \mathcal{T}_{\mathrm{ASR}}\bigl(c^v, \hat{\ell}_i\bigr),  \;\;
    \end{cases}\\
    \xi_t \,=\,
    \begin{cases}
        \bar{c}_t, & \text{if } \bar{c}_t \text{ is a substantive utterance},\\
        \xi_{t-1}, & \text{otherwise},
    \end{cases}
\end{split}
\end{equation}

    We adopt an implicit language conditioning mechanism, where the user utterance itself provides the linguistic context required for semantic interpretation and response generation. Thus, ReLI does not require an explicit language identification prior to command parsing.
    For textual input, we use the provided command directly as the linguistic observation. For speech input, we employ an automatic speech recognition (ASR) front end~\cite{10.1109/ICASSP.2018.8462105, 10.5555/3618408.3619590} to capture high-level audio input and transcribe it into textual representations, i.e., \(\hat{c}^t = \mathcal{T}_{\mathrm{ASR}}\bigl(c^v, \hat{\ell}_i\bigr)\), where $\hat{\ell}_i$ denotes an element of the finite set $\{\hat{\ell}_1, \hat{\ell}_2, \dots, \hat{\ell}_n\}$ of the ASR backend supported languages, and $\mathcal{T}_{\mathrm{ASR}}$ denotes the multilingual speech-to-text transformation. Therefore, the resulting interaction-conditioned command representation is defined as depicted in eqn~\eqref{eq:norm-rep},
    where $\Pi_{\text{norm}}$ performs modality unification and orthographic normalisation (e.g., encoding, whitespace, and punctuation regularisation), \(\xi_t\) is an implicit interaction exemplar derived from the most recent substantive user utterance, and the superscript \(\ell\) in \(\mathcal{Z}^{\ell}_t\) denotes an implicit language-conditioned interaction context.

    In our design, minimal confirmation or rejection tokens, such as \textit{yes}, \textit{no}, \textit{ja}, or \textit{sí}, do not overwrite \(\xi_t\). We adopt this representation to allow ReLI to preserve the user's language, dialectal form, or code-mixed register during plan summaries, confirmation requests, execution updates, and feedback. Hence, from Eq.~\eqref{eq:norm-rep}, regardless of whether the user interacts through speech or text, the downstream pipelines (Section~\ref{subsec3B}) receive a consistent interaction-conditioned textual command representation.

    We developed the interaction interface using the Tkinter library~\cite {moore2018python}, and integrated it into our framework through the ROS~\cite{quigley2009ros} message-passing communication mechanism\footnote{We employed the standard ROS~\cite{quigley2009ros} publish–subscribe communication for bidirectional message exchange between the interface and the action decision pipeline. Normalised user inputs are published to the decision pipeline, while generated responses, plan summaries, confirmation requests, and execution updates are subscribed to and relayed back to the interface. This event-driven architecture ensures that user operations, such as command issuance, trigger the appropriate interface updates and direct message publications.}. Nonetheless, the interaction layer is not constrained to ROS~\cite{quigley2009ros}. The same representation \(\mathcal{Z}^{\ell}_t\) can be instantiated by web-based, speech-only, or dry-run interfaces.

\subsection{Action Decision and Command Parsing}\label{subsec3B}
    The action decision and command parsing pipeline (Fig.~\ref{fig:ReLIArchitect}, top right) abstracts the normalised command representation \(\mathcal{Z}^{\ell}_t\) (Section~\ref{subsec3A2}) into robot-action primitives. Given \(\mathcal{Z}^{\ell}_t=(\bar{c}_t,\xi_t)\) and the current robot state \(\mathbf{s}_t\), our objective is to infer the user's task intent, decompose the instruction into executable sub-goals, and extract the parameters required for downstream grounding and execution. These parameters include target entities, object attributes, spatial constraints, distances, orientations, velocities, tolerances, stopping conditions, and response requirements.
    Formally, we model this process as an LLM-driven semantic mapping conditioned on the command, implicit language context, and robot state: \(\mathcal{F}_{\mathrm{LLM}} :\left(\mathcal{Z}^{\ell}_t,\mathbf{s}_t\right) \mapsto \left(r,\tilde{\mathcal{A}},\rho^{\ell}_t\right)\), where \(r\in\mathcal{R}_{\mathrm{int}}\) denotes the inferred semantic interpretation, \(\tilde{\mathcal{A}}=\{\tilde{a}_1,\tilde{a}_2,\ldots,\tilde{a}_k\}\) denotes the candidate high-level action sequence, and \(\rho_t^\ell\) is the response generated in the corresponding input language.
    The probability of generating this candidate action sequence and response over possible semantic interpretations is:
\begin{equation}
    p_{\theta}(\tilde{\mathcal{A}},\rho_t^\ell \mid \mathcal{Z}^{\ell}_t,\mathbf{s}_t)
    =
    \sum_{r\in\mathcal{R}_{\mathrm{int}}}
    p_{\theta}(\tilde{\mathcal{A}},\rho_t^\ell \mid \mathcal{Z}^{\ell}_t,\mathbf{s}_t,r)
    p_{\theta}(r\mid \mathcal{Z}^{\ell}_t,\mathbf{s}_t),
    \label{eq:llm-prob}
\end{equation}
    where \(\theta\) denotes the frozen parameters of the pretrained LLM. We further factorise the conditional action distribution autoregressively to enforce contextual consistency across sequentially generated action tokens as:
\begin{equation}
    p_{\theta}(\tilde{\mathcal{A}} \mid \mathcal{Z}^{\ell}_t,\mathbf{s}_t,r)
    =
    \prod_{i=1}^{k}
    p_{\theta}
    \left(
    \tilde{a}_i
    \mid
    \tilde{a}_{1:i-1},
    \mathcal{Z}^{\ell}_t,
    \mathbf{s}_t,
    r
    \right),
    \label{eq:action-factorisation}
\end{equation}
    With the decomposition in Eq.\eqref{eq:action-factorisation}, we ensure that each generated high-level action is conditioned on the current command, implicit linguistic context, robot state, inferred intent, and previously generated actions.
    Further, to produce a deterministic robot-executable representation, we apply a hierarchical semantic parser \(\mathcal{P}\). The parser maps the inferred interpretation and candidate action sequence into parameterised action primitives as follows:
\begin{equation}\label{eqn_primit}
    \mathcal{P}(r,\tilde{\mathcal{A}}) = \{\mathcal{A}_1(\phi_1), \dots, \mathcal{A}_n(\phi_n)\} = \{\mathcal{A}_j(\phi_j)\}_{j=1}^{n}, \,
\end{equation}
    where \(\mathcal{A}_j\) denotes a discrete action primitive and \(\phi_j \in \mathbb{R}^{m_j}\) denotes its associated parameter vector, with $n \geq k$ to account for potential high-level actions that may require expansion to multiple primitives (e.g., \textit{``move in a square pattern''} which translates to multiple linear and angular motions), with the associated physical parameters (e.g., distance, angle, speed, etc).
    The action vocabulary, on the other hand, defines the admissible primitive set used by ReLI. Typical primitives include, but are not limited to, navigation actions, object-referenced actions, scene-query actions, response actions, and multi-step task operations. We represent each primitive by a structured set of arguments: $\mathcal{A}_j(\phi_j) = \bigl(\alpha_j, \mathbf{o}_j, \mathbf{s}_j, \mathbf{q}_j,\mathbf{b}_j\bigr)$ where \(\alpha_j\) is the action type, \(\mathbf{o}_j\) contains object or landmark references, \(\mathbf{s}_j\) encodes spatial constraints, \(\mathbf{q}_j\) contains execution parameters such as distance, angle, velocity, tolerance, or stopping condition, and and \(\mathbf{b}_j\) specifies behavioural constraints such as whether grounding, confirmation, or user feedback is required. This representation provides a language-independent interface between instruction interpretation, visual grounding, and robot execution.

    Furthermore, to guarantee reliability, particularly for long-horizon or safety-critical tasks, we introduce a safety user-approval or rejection mechanism that validates whether the generated action plans accurately reflect the user’s intent before being deployed for physical execution. We modelled this as a binary decision problem, \(\rho_d \in \{0, 1\}\), inferred by applying a linear classifier to the embedding of the user's confirmation response as:
\begin{equation}
\rho_d =
  \begin{cases} 
   1 & \text{if } \mathbf{v}^\top \psi(c_\text{resp}) > 0 \implies \text{execute the plan}\\
   0 & \text{otherwise} \implies \text{discard the plan}
  \end{cases}.
\label{eq_confirm}
\end{equation}
    where \(\psi(c_{\mathrm{resp}})\) is the semantic embedding of the confirmation response and \(\mathbf{v}\) is the decision vector. A positive decision \(\rho_d=1\) authorises execution of the parsed action sequence, whereas \(\rho_d=0\) discards the plan. Because the decision is based on semantic embeddings rather than fixed lexical yes/no templates, the confirmation mechanism supports multilingual affirmative and negative responses such as \textit{``proceed''}, \textit{``that is correct''}, \textit{``cancel''}, or \textit{``that is not what I meant''} across the generalisable languages.

\subsection{Open-Vocabulary Visuo-Lingual and Spatial Grounding}\label{subsec3C}
    ReLI’s visuo-lingual and spatial grounding pipeline, shown in Fig.~\ref{fig:ReLIArchitect} (bottom left), grounds parsed linguistic references from Section~\ref{subsec3B} to spatially localised entities in the robot's operating environment. Specifically, given an action primitive that contains an object reference, attributes, and constraints, this pipeline identifies candidate objects, segments their image regions, estimates their 3D poses, and returns grounded action arguments for execution.
    
    Formally, let \(\mathcal{V}_{1:T} = \{(\mathcal{I}_t, \mathcal{D}_t, \mathbf{u}_t, \mathcal{G}_t)\}_{t=1}^{T}\) denote the time synchronised sensory stream, where \(\mathcal{I}_t\in \mathbb{R}^{H \times W \times 3}\) is the RGB frames, \(\mathcal{D}_t \in \mathbb{R}^{H \times W}\) is the corresponding depth maps, \(u_t\) encodes the transformations in the robot’s local frame, and $\mathcal{G}_t$ is the available map representation. From the command parser (Section~\ref{subsec3B}), ReLI receive a set of language-derived object references: $\mathcal{Q}_c = \{d_j\}_{j=1}^{M}$, where each \(d_j\) encodes a target label, attributes, and spatial constraints extracted from the instruction.
    For each \(\mathcal{I}_t\), we employ a visual segmementation model, SAM~3~\cite{carion2025sam} to generate $N$ candidate masks \(\{\mathcal{M}_i\}_{i =1}^{N}\) through both prompt-driven and automatic segmentation, conditioned on \(\mathcal{Q}_c\): $ \mathcal{B}_t = \mathcal{G}_{\mathrm{ov}}(\mathcal{I}_t,\mathcal{Q}_c) = \{(b_i,\mathcal{M}_i,s_i^{\mathrm{det}})\}_{i=1}^{N}$, where \(b_i\) is a bounding region, \(\mathcal{M}_i\) is the corresponding mask, and \(s_i^{\mathrm{det}}\) is the detector confidence.
    
    For automatic mask generation, we employ convex hull analysis to evaluate the quality of the candidate masks $\mathcal{M}_i$ before semantic matching. The ratio of the mask area to the convex hull area $\mathbf{q}(\mathcal{M}_i)$ determines its validity, and low-quality masks with $\mathbf{q}(\mathcal{M}_i) < \mathbf{q}_{\text{thresh}}$ are discarded, where $\mathbf{q}_{\text{thresh}}$ is the quality threshold. For retained valid masks with $\mathbf{q}(\mathcal{M}_i) > \mathbf{q}_{\text{thresh}}$, we encode the masked image patches and the parsed textual reference into a joint visual-lingual embedding space: \(\mathbf{v}_i = f_\text{visual}(\mathcal{I}_t \odot \mathcal{M}_i)\;,\; t_j=f_\text{text}(d_j)\) and then compute their semantic alignment as \(S_{ij} = \cos(\mathbf{v}_i, t_j),\). Further, we apply a temperature-scaled, \(T_{\mathrm{temp}}\), class-reference distribution as follows:
\begin{equation}\label{eqn_vlm}
    p(d_j \mid \mathcal{M}_i) 
    = 
    \frac{\exp\!\bigl(\tau\,S_{ij}\bigr)}{\sum_{k=1}^M \exp\!\bigl(\tau\,S_{ik}\bigr)},
    \quad
     \tau = \frac{1}{T_{\mathrm{temp}}}, \quad T_{\mathrm{temp}} > 0,
\end{equation}
    where larger \(\tau\) (lower \(T_{\mathrm{temp}}\)) produces a sharper distribution, and thus increases the model's confidence, whereas lower \(\tau\) yields a smoother distribution, with greater uncertainty. To ensure that only strong visual-language predictions propagate downstream, we filter uncertain detections through an energy-based uncertainty quantification score, $\mathbf{e}_\tau(\mathcal{M}_i) = -\tau^{-1}\log \sum_{j=1} \exp\!\bigl(\tau S_{ij}\bigr) > \mathbf{e}_{\text{thresh}}$ by rejecting masks that exceeds the defined energy threshold, $\mathbf{e}_{\text{thresh}}$. We retain a candidate prediction only if its energy score and detector confidence satisfy the configured acceptance criteria.

    Since perception quality often degrades under adverse environmental conditions (e.g., low illumination, occlusion, or motion blur), we introduced a degradation-aware reliability weighting to modulate the contribution of each mask to the final grounding decision. We downweight probabilities for masks in degraded regions using \(\Theta_{ij}(\Omega_t) = \exp\bigl(-\beta \,\eta_{ij}\bigr)\,,\; \eta_{ij} \in \mathbb{R}_{\geq 0}\), where  \(\Omega_t\) denotes frame-level degradation indicators, \(\eta_{ij}\) measures the descriptor-specific degradation for mask \(\mathcal{M}_i\) and reference \(d_j\), and \(\beta > 0\) controls sensitivity. Therefore, Eq.~\eqref{eqn_vlm} can be rewritten as:
\begin{equation}\label{eqn_deg}
    p'(d_j \mid \mathcal{M}_i,\;\Omega_t) 
    = 
    \frac{p(d_j \mid \mathcal{M}_i) \cdot \Theta_{ij}(\Omega_t)}
    {\sum_{k=1}^{M} 
     \Bigl(
        p(d_k \mid \mathcal{M}_i) \cdot \Theta_{ik}(\Omega_t)
     \Bigr)}.
\end{equation}

    To spatially ground and track detected objects, we employed the aligned depth map \(\mathcal{D}_t\). At the mask's centroid $(u_c\,, v_c)$, we compute the depth value \(z_c\) as the median of valid depth measurements within the local neighbourhood:
\begin{equation}\label{eqn:med}
    z_c =
    \begin{cases}
    \mathrm{med}\!\left(\mathcal{N}_r(u_c,v_c) \cap \mathcal{M}_i \odot \mathcal{D}_t\right),
    & \text{if valid RGB-D samples exist},\\[2mm]
    \mathrm{med}\!\left(\mathcal{M}_i \odot \hat{\mathcal{D}}_{\mathrm{mono}}\right),
    & \text{otherwise},
    \end{cases},
\end{equation}
    where \(\mathcal{N}_r(u_c,v_c)\) denotes a local pixel neighbourhood around the centroid, and \(\hat{\mathcal{D}}_{\mathrm{mono}}\) represents an optional monocular depth estimate. In physical deployment, we derive the monocular depth from Depth Anything~\cite{lin2025depth}, and we use it only when its metric scale can be aligned with available depth measurements; otherwise, the target object is re-queried.

    Given camera intrinsics \(\mathbf{K}\), we back-project the centroid into 3D space, as \(\mathbf{x}_o = z_c \mathbf{K}^{-1} [u_c \; v_c  \;1]^{\top}  \in \mathbb{R}^3\).
    We then transform $\mathbf{x}_o$ to the robot’s base frame using iterative TF lookups to handle temporal synchronisation (i.e., for ROS~\cite{quigley2009ros} implementation). In non-ROS deployments, the same operation is performed through the configured camera-to-base calibration and pose-estimation backend. For temporal consistency, we use a Kalman filter to track the object poses, with state dynamics as $\mathcal{X}_{t + 1} = \mathrm{F}\mathcal{X}_{t} + \mathbf{w}_t$ to smooth pose estimates and account for motion uncertainty. $\mathrm{F}$ is the motion model, \(\mathcal{X}_t = [x,y,z,\dot{x},\dot{y},\dot{z}]^{\top}\) is the object's state vector at time $t$, and $\mathbf{w}_t \sim \mathcal{N}(0, \mathrm{Q})$ is the process noise with covariance $\mathrm{Q}$. The observation $\mathcal{Z}_t = \mathcal{H}\mathcal{X}_t + \mathbf{v}_t$ are obtained from the depth-based back-projection (Eq.~\eqref{eqn:med}), where $\mathcal{H}$  extracts the position components and $\mathbf{v}_t \sim \mathcal{N}(0, \mathrm{R})$ is the measurement noise with covariance $\mathrm{R}$ estimated from the depth sensor.
    
    Finally, given an object-referenced linguistic command, we determine the grounded target \(o^*\) by combining visual confidence, semantic alignment, spatial consistency, and relation constraints as follows:
\begin{equation}\label{eq_refObj}
\begin{split}
    o^* = \arg\max_{i,j}
    \Big[
        \lambda_1 \log p'(d_j \mid \mathcal{M}_i,\Omega_t)
        +
        \lambda_2 \cos(\mathbf{v}_i,\mathbf{t}_j)
        + \\
        \lambda_3 s_i^{\mathrm{det}}
        +
        \lambda_4 \Gamma(d_j,\mathcal{M}_i,\mathcal{D}_t,\mathcal{G}_t)
    \Big],
\end{split}
\end{equation}
    where \(\Gamma(d_j,\mathcal{M}_i,\mathcal{D}_t,\mathcal{G}_t)\) measures whether candidate \(\mathcal{M}_i\) satisfies the spatial relation or constraint in \(d_j\) and $\lambda_1, \ldots, \lambda_4 > 0$ are weighting coefficients to control the relative contribution of calibrated reference probability, semantic similarity, detector confidence, and spatial-relation consistency.
    The resulting output of Eq.~\eqref{eq_refObj} corresponds to the grounded target representation, $g_t = \left(o^*, \mathcal{M}^*, \hat{\mathbf{x}}^*, q^*, \gamma^* \right)$, where \(o^*\) is the selected object, \(\mathcal{M}^*\) is its segmentation mask, \(\hat{\mathbf{x}}^*\) is the filtered 3D pose, \(d^*\) is the matched linguistic reference, and \(\gamma^*\) is the final grounding confidence. This grounded representation is passed to the execution mechanism (Section~\ref{subsec3D}) as the metric target and constraint set for robot action execution.

\subsection{Robot control and Action Execution Mechanism}\label{subsec3D}
    We operationalise the high-level intents derived from the action decision (Section~\ref{subsec3B}) and the visuo-lingual pipelines (Section~\ref{subsec3C}) into physical robot behaviours through the action execution mechanism (AEM) ( see Fig.~\ref{fig:ReLIArchitect}, bottom right). Generally, given a parsed action sequence \(\mathcal{P}(r)=\{\mathcal{A}_j(\phi_j)\}_{j=1}^{n}\), a grounded target representation \(g_t\), and the current robot state \(\mathbf{s}_t\), the AEM manages the action execution mapping, \(\mathcal{E}_{\mathrm{AEM}}: \left(\mathcal{A}_j(\phi_j), g_t, \mathbf{s}_t\right) \mapsto \mathbf{u}_{1:T_j},\) where \(\mathbf{u}_{1:T_j}\) represents the control commands issued over the action horizon \(T_j\). Depending on the action primitive type, these commands may correspond to navigation goals, velocity profiles, sensor queries, or user-facing response actions.
    
    For navigation, the target can be specified as explicit coordinates \((x_g,y_g,z_g)\), a symbolic destination, or an object-relative goal produced by the grounding pipeline (Section~\ref{subsec3C}). In our ROS-based implementation, we rely on a highly efficient Rao-Blackwellized particle filter-based SLAM~\cite{4084563} to learn occupancy representations from the robot's operational environment. We then localise the robot within the learned occupancy map, utilising the Adaptive Monte Carlo Localisation algorithm~\cite{THRUN200199}, which maintains a particle-based distribution over the probable state of the robot in the environment.
    Given the current robot pose \(\mathbf{x}_t\), target pose \(\mathbf{x}_g\), and the map \(\mathcal{G}_t\), we employ the navigation backend to compute a feasible path \(\pi_{1:T}=\Pi_{\mathrm{nav}}(\mathbf{x}_t,\mathbf{x}_g,\mathcal{G}_t)\) to the goal. For object-referenced navigation, we derived the target pose from the grounded object. Thus, a command such as \textit{``move to the black dog and stop $0.5 m$ ahead''} becomes a metric object navigation objective, interpretable and executable by the AEM.
    
    Beyond large-scale navigation, the AEM also supports low-level motion and query-oriented action primitives that do not require mapping, path planning, or obstacle avoidance. Commands like \textit{``move in a geometric pattern of length $3~m$ and breadth $2~m$ at $0.5~m/s$''} or \textit{``perform a $180^\circ$ arc of radius $2~m$''} are directly mapped into continuous linear and angular velocity profiles through twist messages, i.e., \(\Lambda: \mathcal{A}_j(\phi_j) \mapsto \{(\mathbf{v}(t), \mathbf{\omega}(t))\}_{t=1}^{T_i}\), where \(\mathbf{v}(t)\) and \(\mathbf{\omega}(t)\) are the linear and angular velocities profiles. Further, for query-oriented commands that do not involve physical movements, e.g.,  \textit{``what is in front of you?''}, we directly invoke the visuo-lingual pipeline (Section~\ref{subsec3C}) to generate the corresponding outputs.

\section{Experiments and Results}\label{sec2}
    We evaluated ReLI in both simulated and real-world environments in order to validate its full potential to ground multilingual task instructions into executable robot behaviours. In our experiments, we examine four core aspects of the ReLI framework: (i) cross-lingual instruction parsing, (ii) visuo-lingual and spatial grounding, (iii) robot action execution, and (iv) interaction quality under both text- and speech-based user instructions. In this section, we report the quantitative results obtained across multilingual task trials, and the qualitative analysis of human-rater interactions in which users issued commands in their native languages, as well as the comparison to existing interactive baselines.

\subsection{Experiment Platforms and Setup}\label{subsec4A}
    In our experiments, we utilised two robotic embodiments: (i) a wheeled differential drive robot, and (ii) a quadruped. Both platforms were equipped with RGB-D cameras and LiDAR sensors to provide synchronised visual and spatial observations. All simulations were conducted in both web-based and ROS environments with an NVIDIA GeForce RTX-4090 ground-station PC. The Gazebo ROS simulation world comprised $11$ interconnected rooms and an external corridor, mirroring a typical indoor office layout with furniture (tables, chairs, shelves) and standing obstacles. For audio-based interaction, the PC's onboard microphone array was employed to capture vocal instructions.

    For real-world deployment, we used a Lenovo ThinkBook with an Intel Core i7 CPU and Intel Iris integrated graphics. Experiments were conducted in our laboratory environment of approximately $28.72 \times 12.75 \, \text{m}^2$, containing standard furnishings comparable to those in simulation.
    We benchmarked multiple LLM backends, including the open-source LLaMA~3.2~\cite{Touvron2023LLaMAOA}, Gemini~\cite{team2023gemini}, and GPT-series~\cite{singh2025openai,openai2023reasoning}. In our preliminary comparisons, GPT-series~\cite{singh2025openai,openai2023reasoning} produced the most consistent contextual interpretation and action parsing. Thus, all quantitative and qualitative results reported herein (Sections~\ref{sec:quant} \& ~\ref{sec:qualitative}) were obtained leveraging GPT models~\cite {singh2025openai,openai2023reasoning} as the instruction-reasoning backends. Further details about the key model backends, decoding settings, perception thresholds, execution parameters, and other relevant experimental hyperparameters are provided in Appendices~\ref{append:hyper-param}.

\subsection{Benchmark Design and Dataset Construction}\label{subsec4B}
    \textbf{Languages.} Ultimately, we are mostly interested in the number of languages that ReLI can ground into real-world robotic affordances. For this, we conducted an extensive multilingual evaluation of ReLI to investigate its generalisation across the global languages. 
    We randomly chose $140$ representative languages from the ISO 639~\cite{ISO639} language catalogue, distributed across the continents, language families, and resource tiers.
\begin{figure*}[htp]
    \centering
    \includegraphics[width=0.8\linewidth]{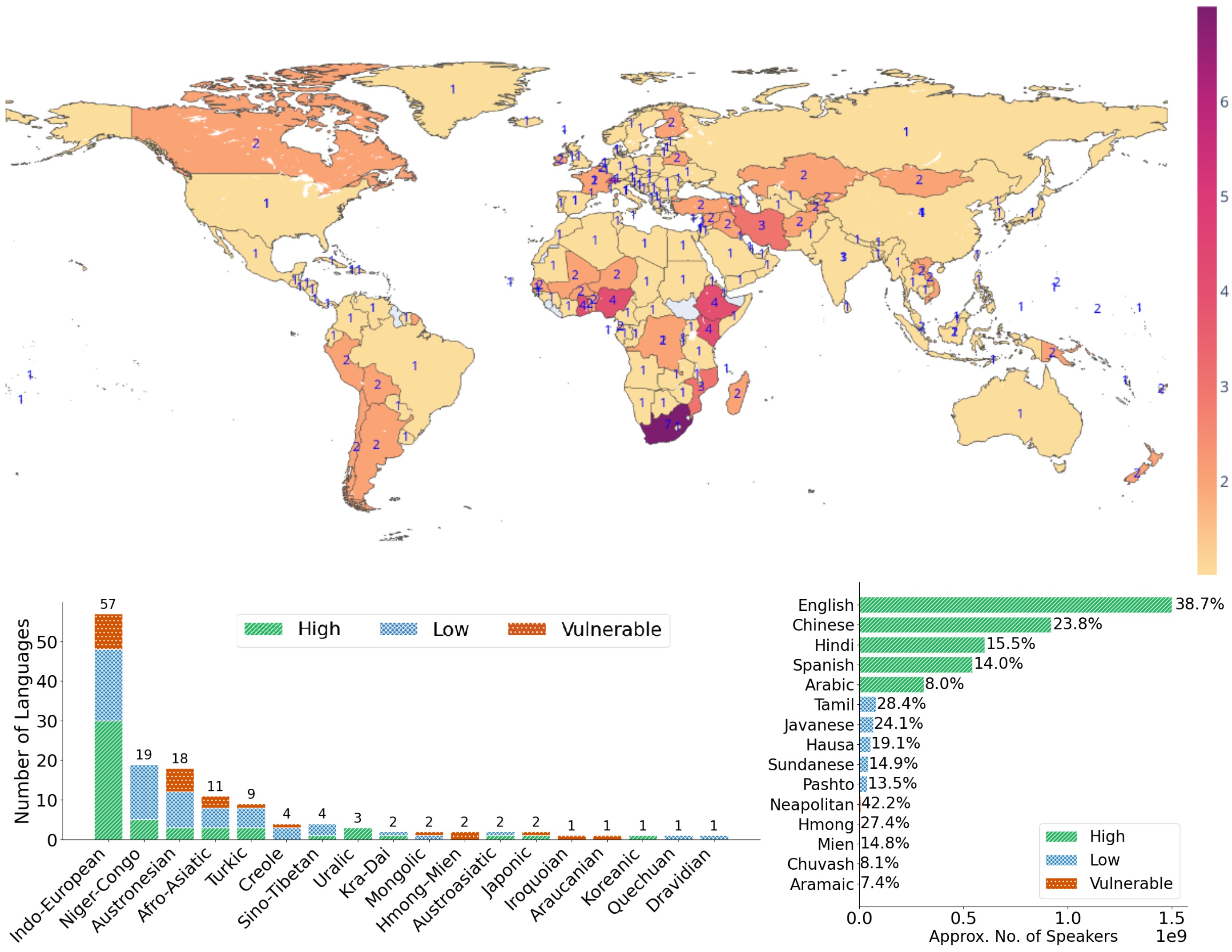}
    \caption{Distributions of the $140$ representative languages utilised for ReLI benchmarking. We prioritise the inclusion of low-resource and vulnerable languages in our selection criteria (bottom left), as we posit that this will rigorously evaluate the robustness of ReLI's cross-lingual language-to-action grounding beyond high-resource settings. Further, to promote inclusive and accessible HRI, we ensured that our selected languages are strategically distributed across the world's continents (top).}
    \label{fig:lang-distrib}
\end{figure*}
    Following similar resource-oriented taxonomies used in NLLB~\cite{costa2022no} and Joshi et al.~\cite{joshi-etal-2020-state}, we categorised languages with strong digital presence, large-scale corpora, and well-established tokenisers as high-resource languages (HRL). We considered those with limited digital presence, low-scale training corpora, and weaker institutional support as low-resource languages (LRL).
    Further, we grouped creoles, vernaculars, rare dialects, and languages with minimal recognised computational support or documented endangerment status~\cite{moseley2010atlas,wiki:endangeredlanguage}, yet are decodable by the LLM backend as vulnerable languages (VUL).
    Fig.~\ref{fig:lang-distrib} shows the distribution of our chosen languages across continents, language family, and resource tier. The full details are provided in Appendix~\ref{append:a}.

\textbf{Task instructions and rationales.}
    We designed benchmark tasks (see Appendix~\ref{append:c}) that target ReLI's core capabilities: multilingual parsing, action-primitives generation, spatial grounding, numeric-parameter extraction, contextual reasoning, object-referenced navigation, and query-oriented interaction.
\begin{figure}[htp]
\centering
    \includegraphics[width=1.0\linewidth]{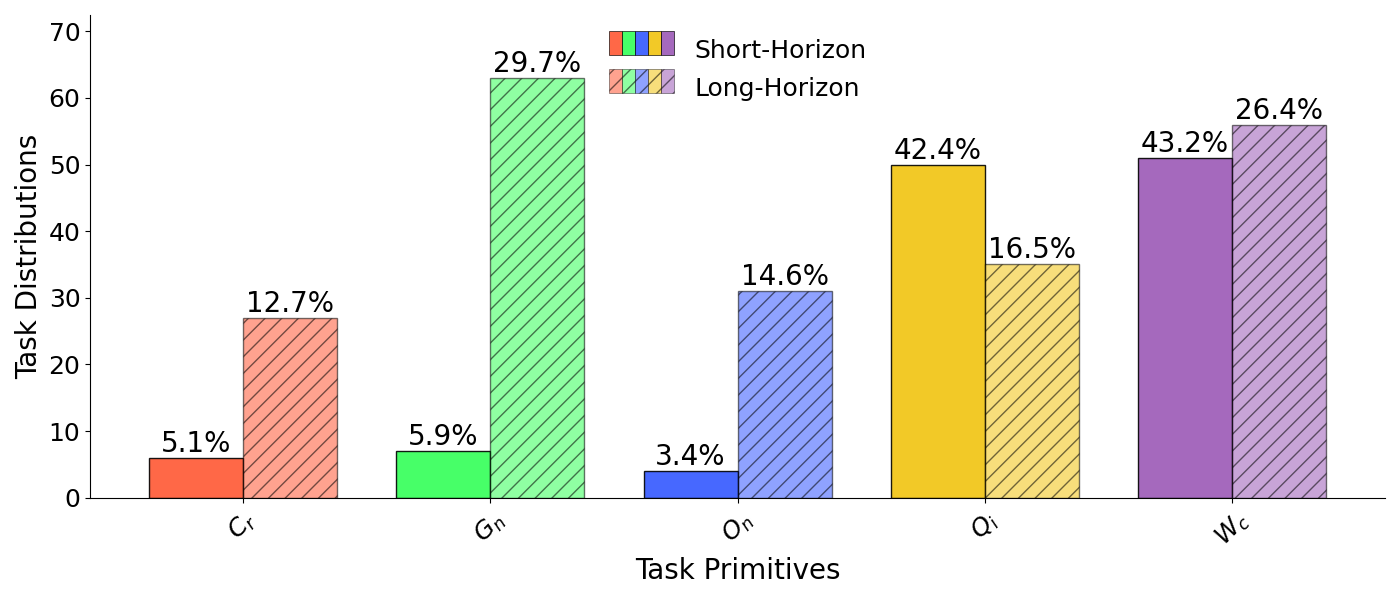}
    \caption{Overview of task instructions distribution. Short-horizon tasks involve atomic actions that require no user approval, path planning, or obstacle avoidance, whereas long-horizon tasks require multi-step action decomposition and user approval or rejection before execution. The labels correspond to \(G_n\) (zero-shot spatial and goal-directed tasks), \(W_c\) (movement commands without location targeting), \(Q_i\) (information and perception queries), \(O_n\) (object-based navigation), and \(C_r\) (contextual and descriptive reasoning).}
    \label{examp-task-distr}
\end{figure}
    Since we were unable to quantify all open-ended language-conditioned task instructions that ReLI can ground in real time, we instead structured the task space as: \(\mathcal{T}_{T}^{Re} = (\mathcal{G}_n, \mathcal{W}_c, \mathcal{Q}_i, \mathcal{O}_n, \mathcal{C}_r)\), where $\mathcal{G}_n$ denotes all zero-shot spatial or goal-directed navigation tasks (e.g., \textit{``navigate to the coordinates \((x_g, y_g, z_g)\)''} or to a named destination, \textit{``head to the kitchen''}); $\mathcal{W}_c$ are low-level motion control commands that do not require map-based navigation (e.g., \textit{``move forward $d$ meters at a speed of $v~m/s$''}, \textit{``rotate $\theta$ degrees''}, etc); 
    \(\mathcal{Q}_i\) represents instructions that probe general knowledge, causal reasoning, or sensor-based queries (e.g., \textit{``what are your capabilities?'', ``send me photos of your surroundings'', etc}); \(\mathcal{O}_n\) denotes object-referenced navigation tasks (e.g., \textit{``go towards the detected chair''}); and \(\mathcal{C}_r\) are task instructions that demand contextual or descriptive reasoning. For example, the command \textit{``head to the location where one can cook food''} implies navigating to the kitchen, while \textit{``go to where administrative tasks are handled''} should be mapped to the secretary's office.

\begin{figure*}[htp]
    \centering
    \begin{subfigure}[b]{0.5\textwidth}
    \centering
    \includegraphics[width=\linewidth]{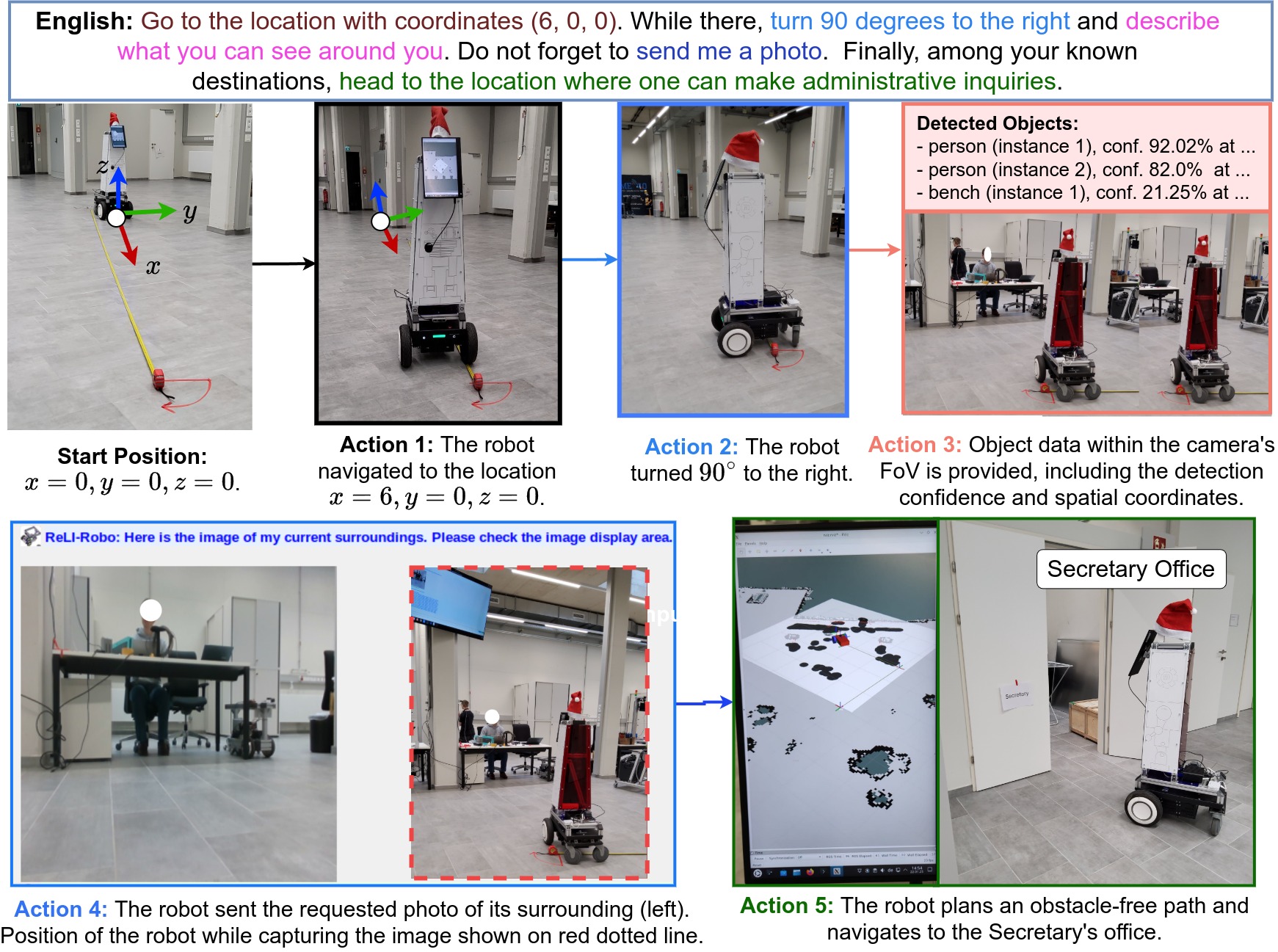}
    \caption{English task involving coordinate-based navigation, scene understanding, object detection, and contextual reasoning.}
    \label{examp-task-eng}
    \end{subfigure}
    \hspace{0.2cm}
    \vspace{0.3cm}
    \begin{subfigure}[b]{0.42\textwidth}
    \centering
    \includegraphics[width=\linewidth]{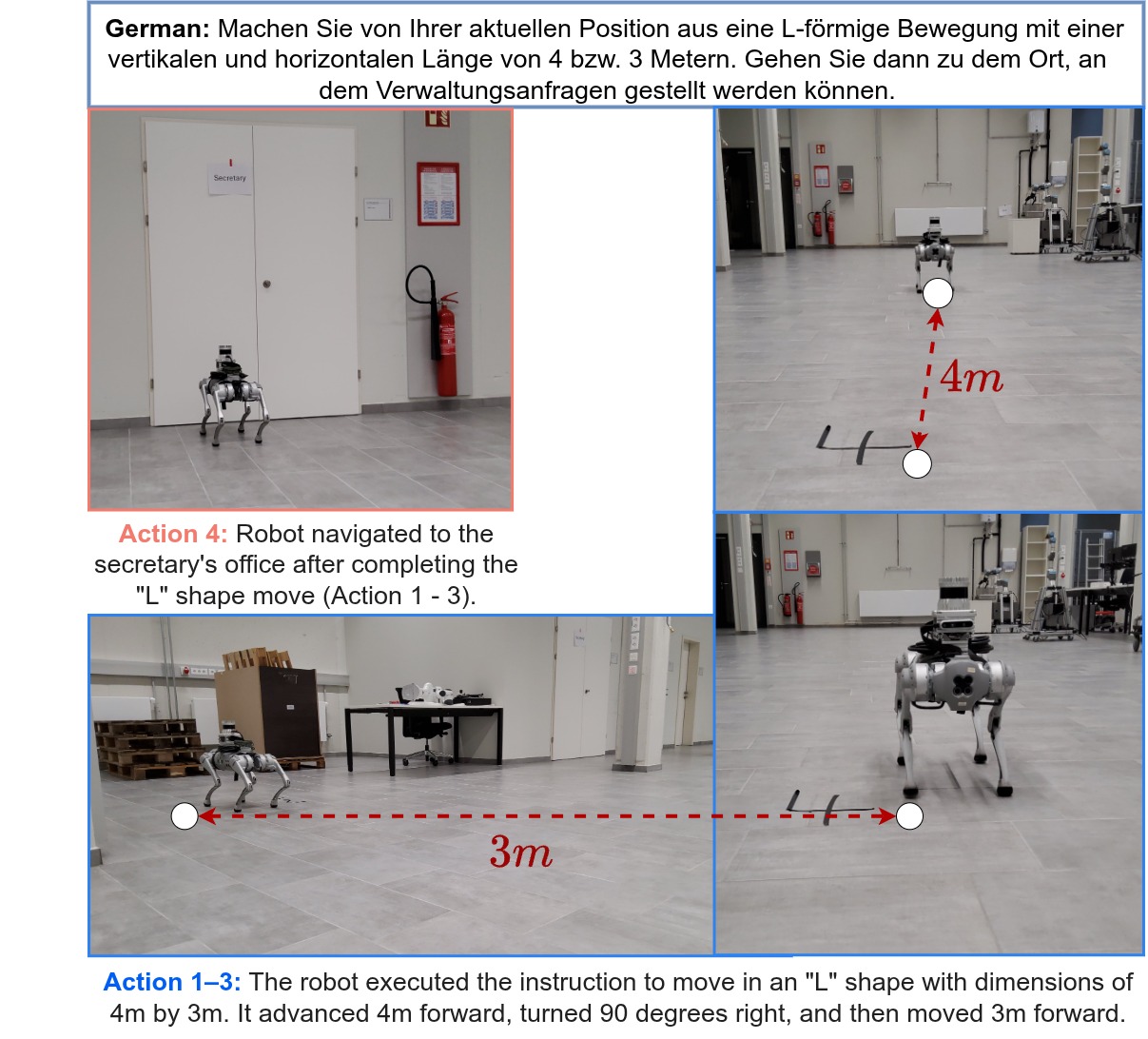}
    \caption{German task involving geometric motion, patterned trajectories, and coordinate-based navigation.}
    \label{examp-task-germ}
    \end{subfigure}
    \begin{subfigure}[b]{0.37\textwidth}
    \centering
    \includegraphics[width=\linewidth]{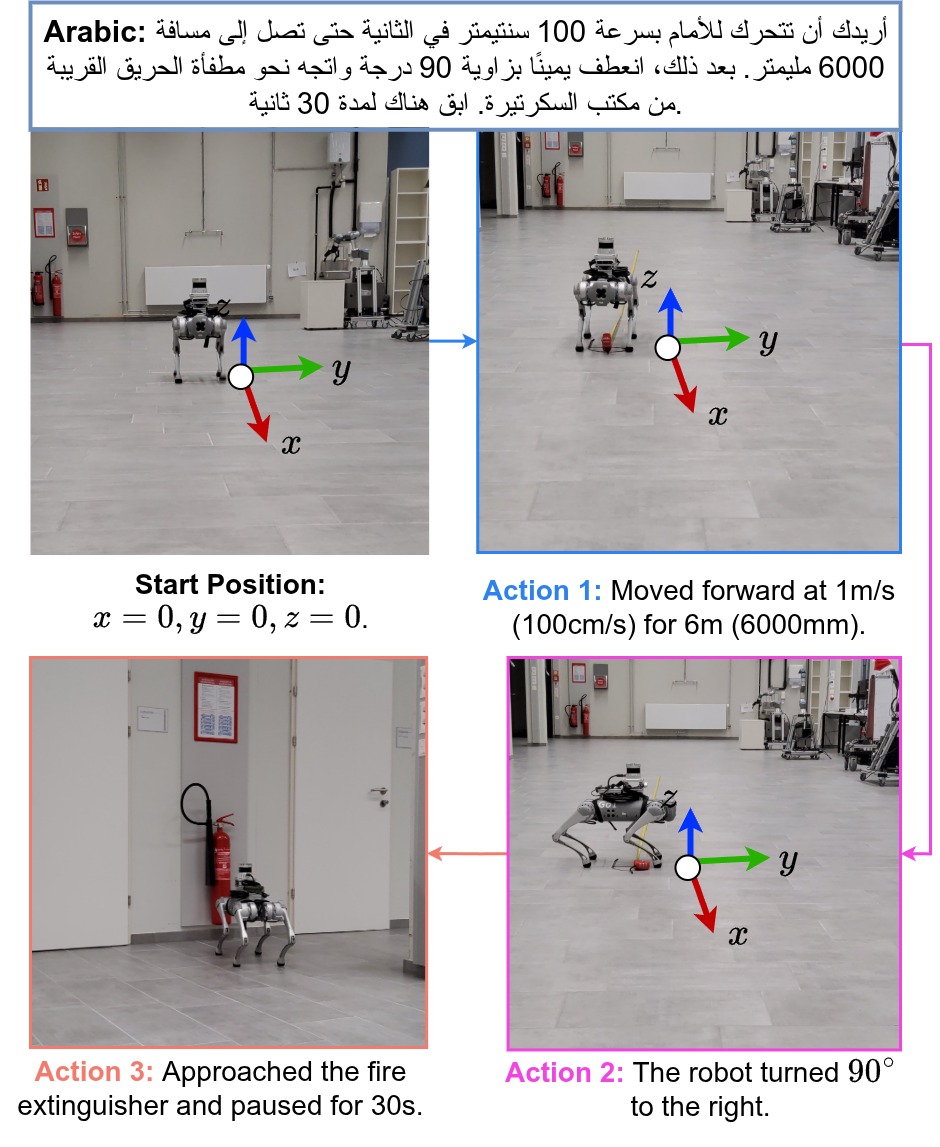}
    \caption{Arabic task involving SI-unit constraints, object detection, and object-referenced action.}
    \label{examp-task-arab}
    \end{subfigure}
    \hspace{0.3cm}
    \begin{subfigure}[b]{0.53\textwidth}
    \centering
    \includegraphics[width=\linewidth]{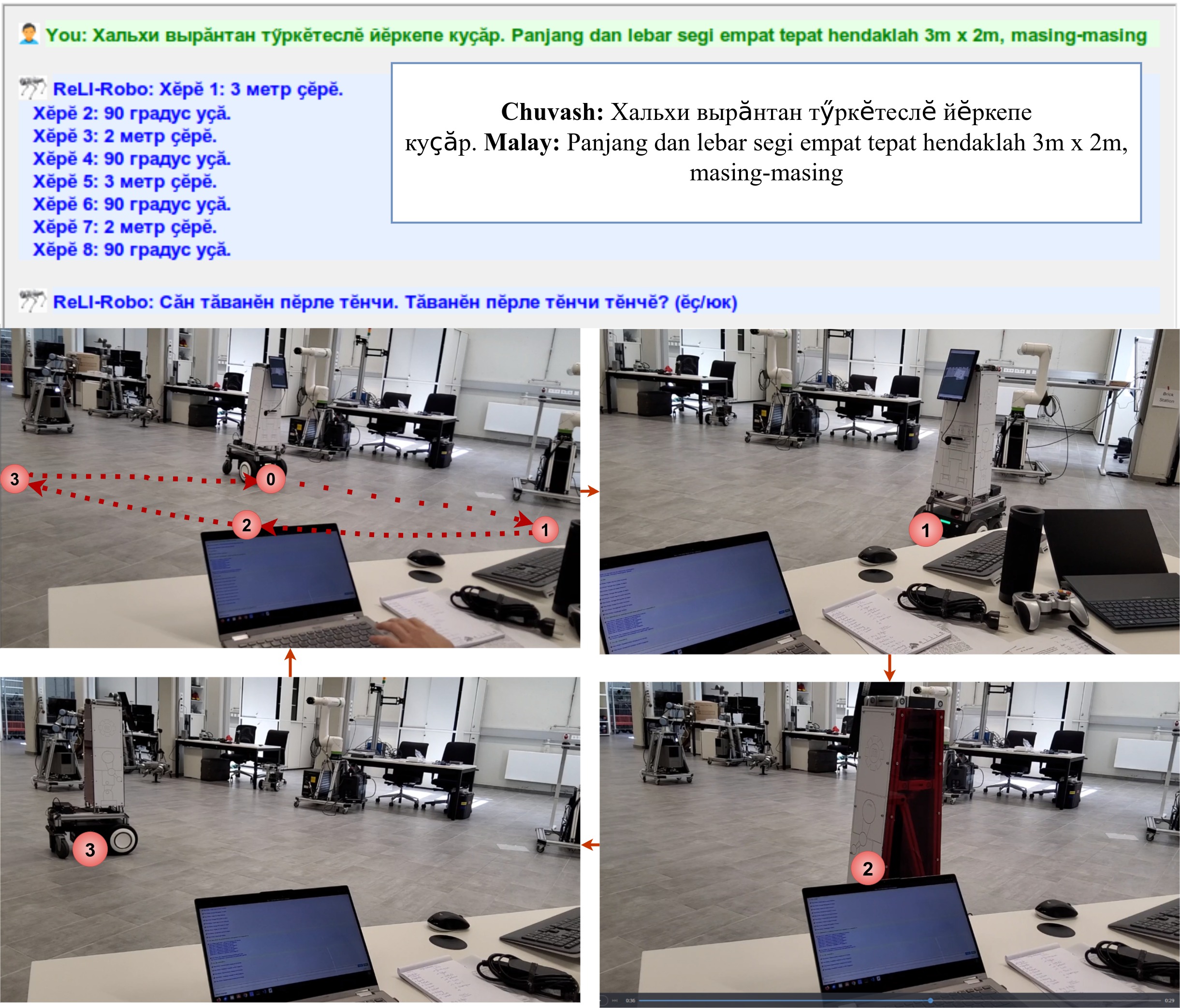}
    \caption{Code-switched Chuvash-Malay instruction testing intermixed-language parsing within a single command.}
    \label{examp-task-chu-mal}
    \end{subfigure}
    \caption{Representative multilingual task executions. These examples illustrate ReLI's ability to parse, ground, and execute task instructions across different languages, scripts, and task categories. Additional examples are provided in Appendix~\ref{append:b}.}
    \label{fig:task-primitives}
\end{figure*}
    Fig.~\ref{examp-task-distr} shows the distribution of the task instructions utilised in our benchmark, and  Fig.~\ref{fig:task-primitives}(a - d) illustrates example task deployments across different languages. For each language, we conducted multiple random short- and long-horizon task trials distributed across the five aforementioned task categories. 
    In addition to inter-lingual translated task instruction trials, we intermittently recruited $34$ external human raters (mean age: $25 \pm3$; gender distribution: 65\% male, 32\% female, 3\% other), each fluent in one or more of the evaluated languages (see Appendix~\ref{append:b}, Table~\ref{tab:raters-demong}). We instructed the raters to interact with the robots through text or speech and to issue commands that involved navigation, object identification, scene queries, and robot-status inquiries in their native language. Similar to~\citet{nwankwo2024conversation}, we logged the interaction datasets: \(\mathcal{D} = \{( c_n, t_{n}^{\mathrm{ins}}, t_{n}^{\mathrm{res}}, \mathcal{A}_n, \hat{\mathcal{A}}_n, s_n)\}_{n = 1}^N\), where $c_n$ is the issued command, $t_{n}^{\mathrm{ins}}$ is the instruction timestamp, $t_{n}^{\mathrm{res}}$ is the timestamp at which the robot began responding or executing, \(\mathcal{A}_n\) and \(\hat{\mathcal{A}}_n\) are the ground-truth and predicted action sequences, and \(s_n\in\{0,1\}\) is the execution success indicator. This resulted in over $70$K logged multi-turn interaction data, which allowed us to evaluate ReLI's instruction understanding, action-sequence agreement, response latency, and end-to-end task success under both interlingual translated and native user task interactions.

\subsection{Interlingual Translation and        Quality}\label{sec:trans-qual}
\begin{figure*}[htp]
    \centering
    \begin{subfigure}[b]{0.46\textwidth}
        \includegraphics[width=\linewidth]{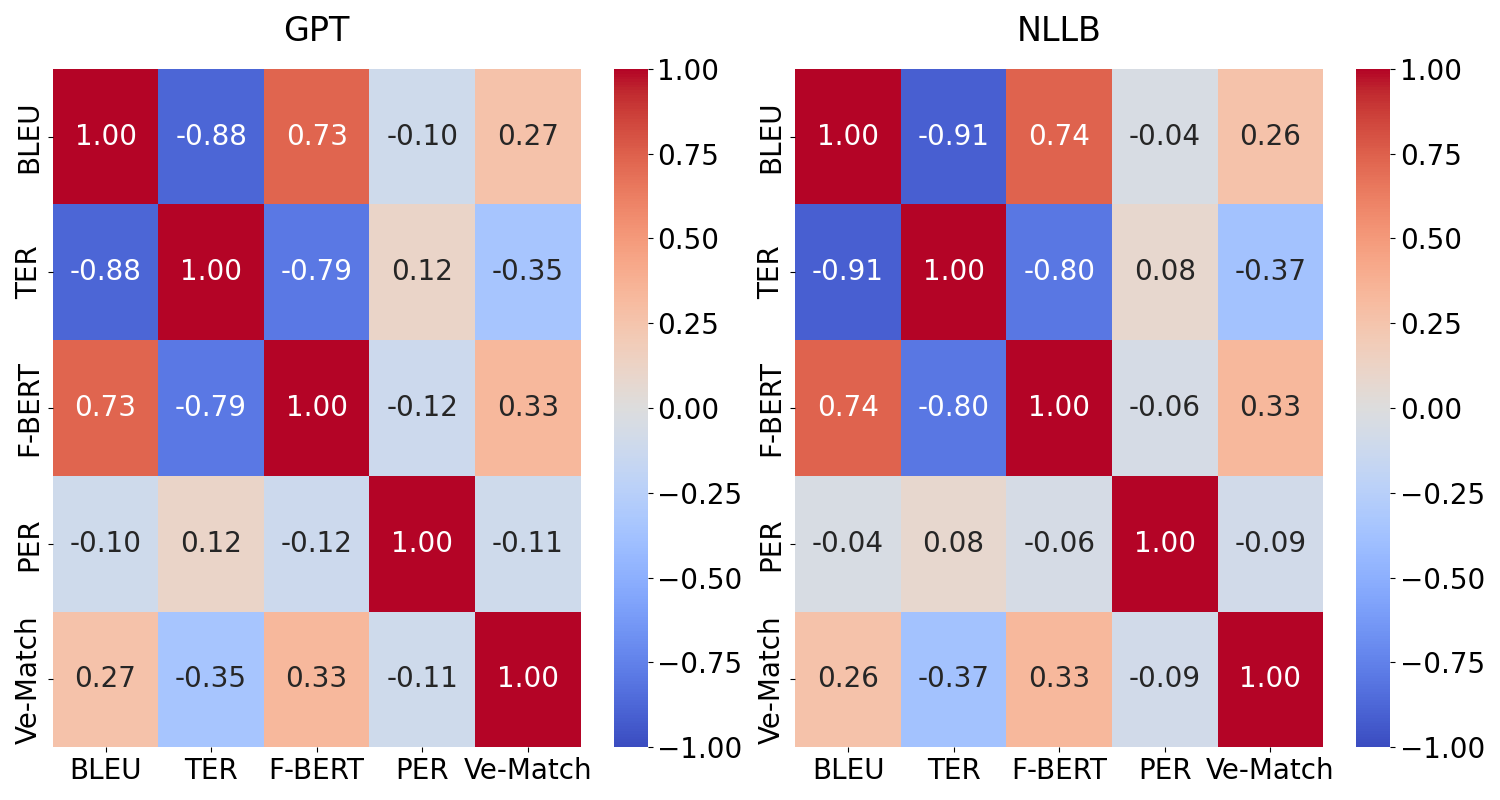}
        \caption{Metric correlations}
    \end{subfigure}
    \hspace{0.7em}
    \begin{subfigure}[b]{0.46\textwidth}
        \includegraphics[width=\linewidth]{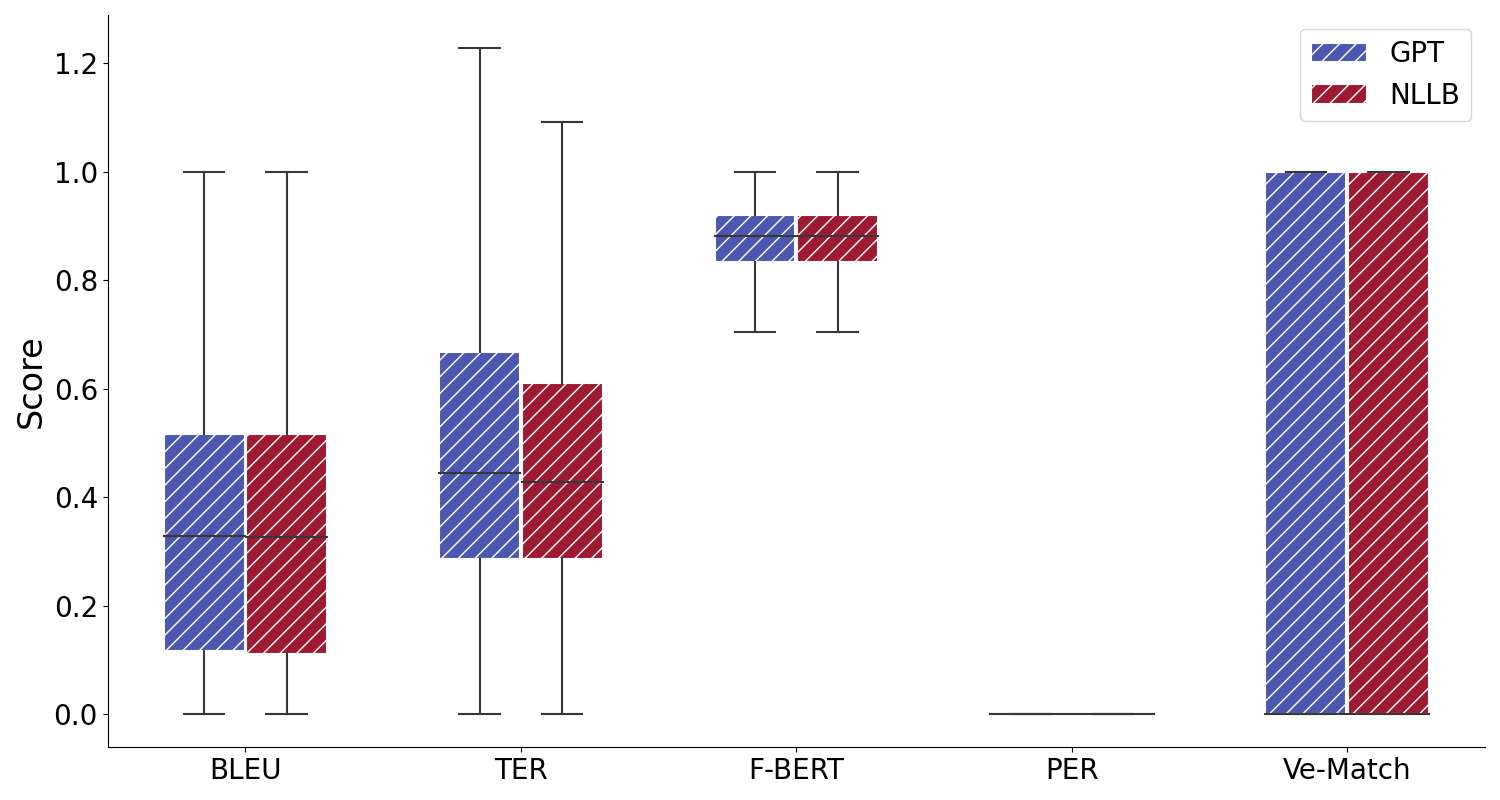}
        \caption{Aggregate score}
    \end{subfigure}
    \begin{subfigure}[b]{0.46\textwidth}
        \includegraphics[width=\linewidth]{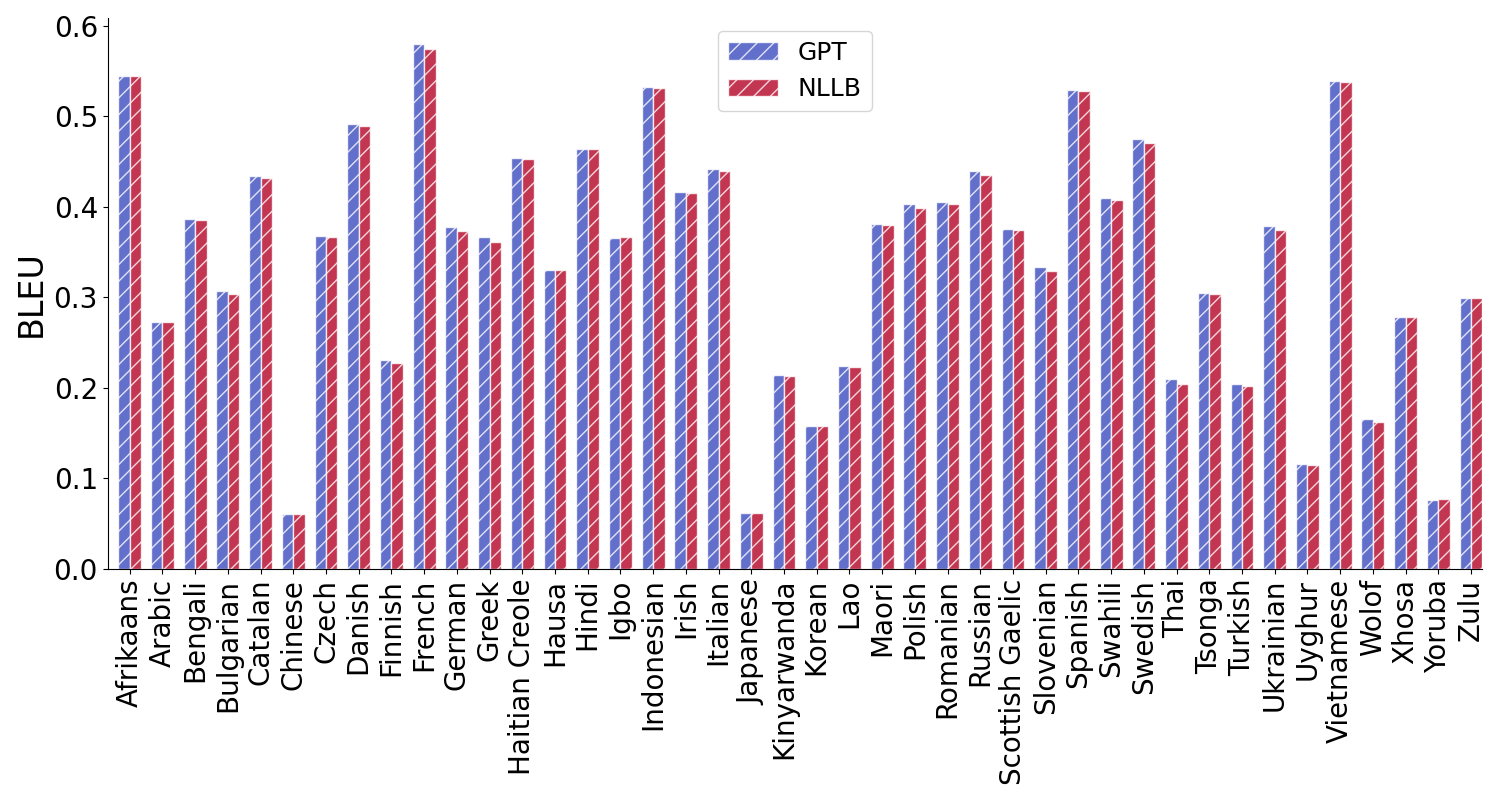}
        \caption{BLEU}
    \end{subfigure}
    \hspace{0.9em}
    \begin{subfigure}[b]{0.46\textwidth}
        \includegraphics[width=\linewidth]{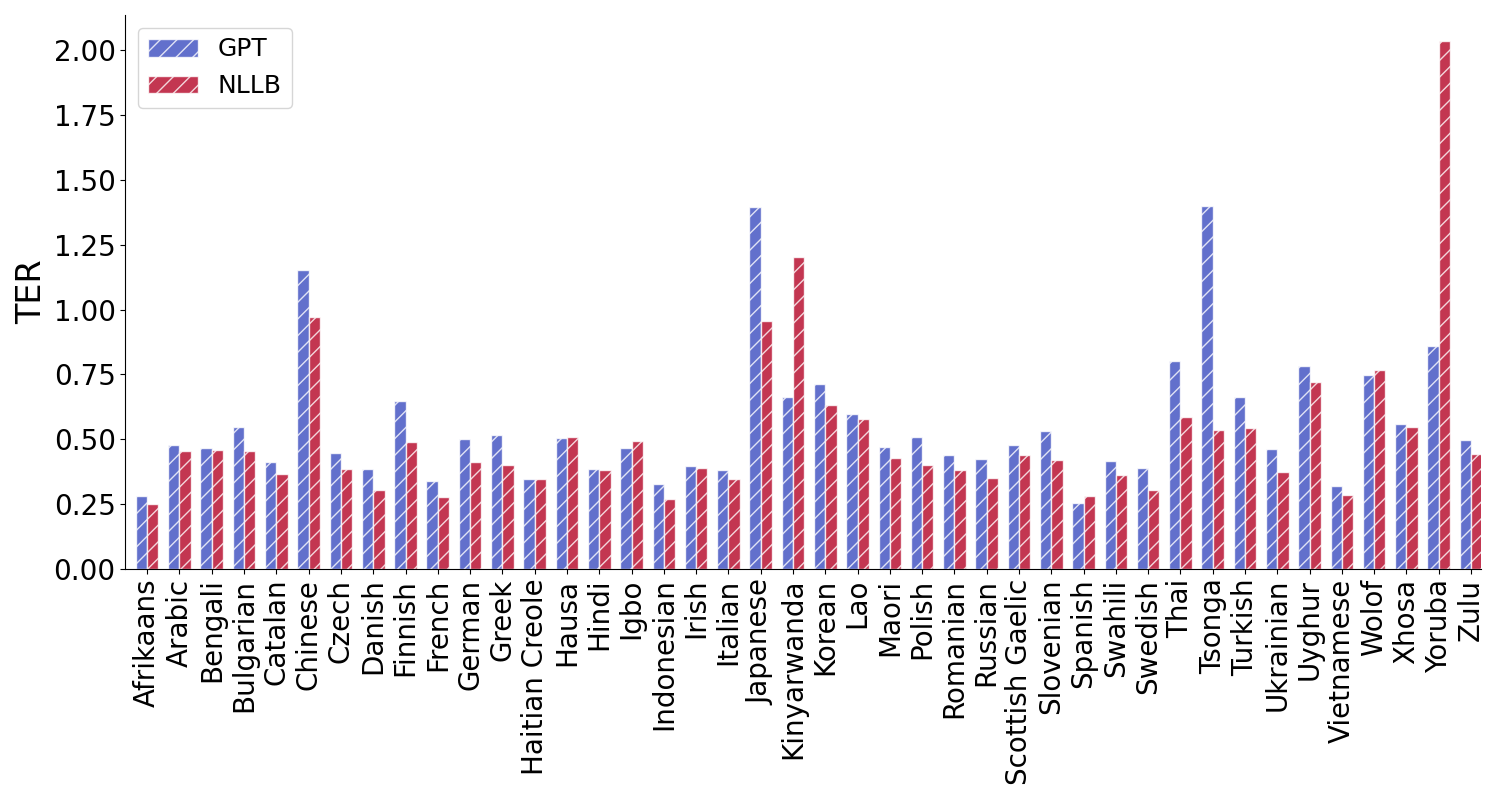}
        \caption{TER}
    \end{subfigure}
    \begin{subfigure}[b]{0.46\textwidth}
        \includegraphics[width=\linewidth]{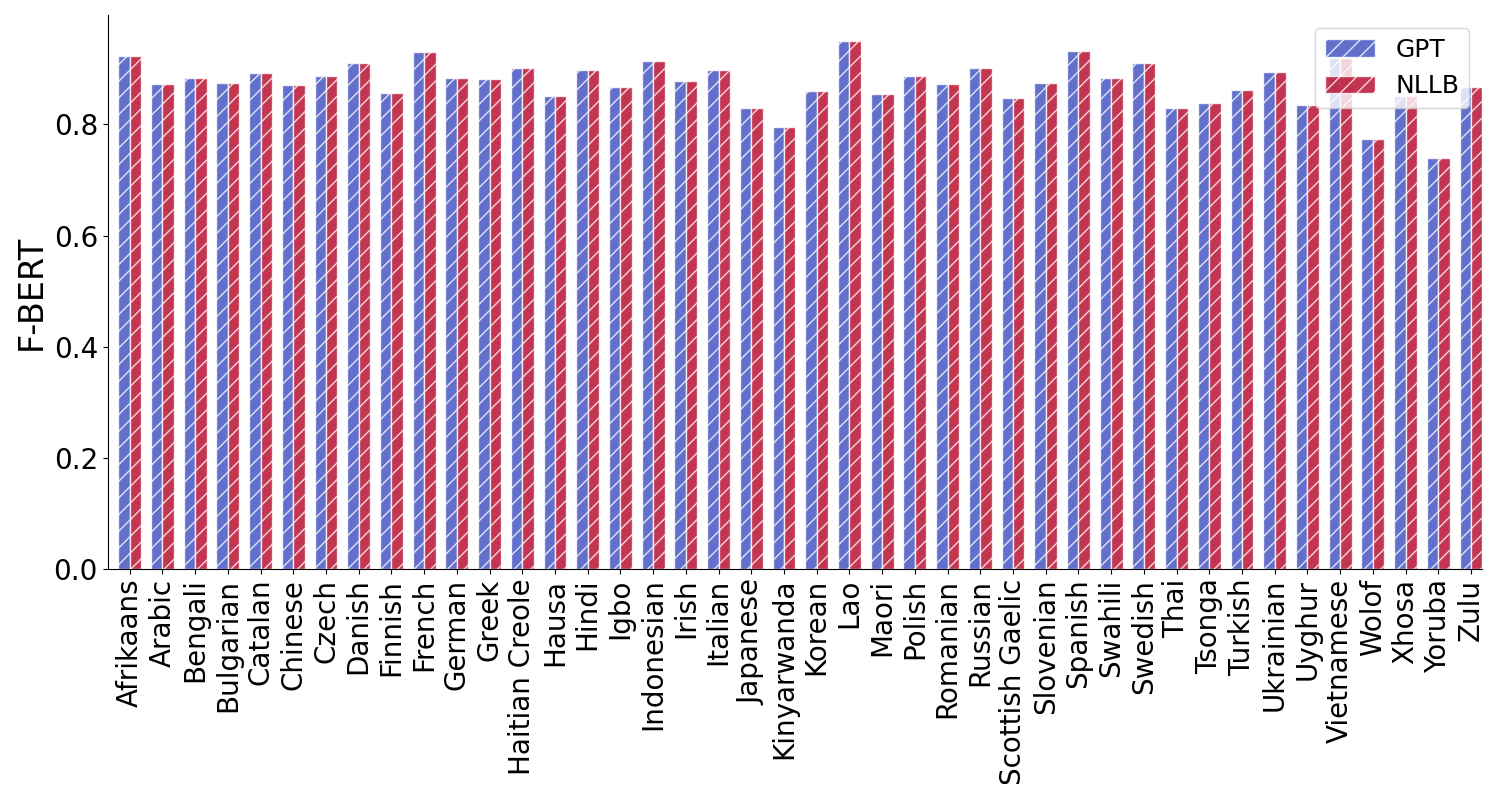}
        \caption{BERTScore}
    \end{subfigure}
    \hspace{0.9em}
    \begin{subfigure}[b]{0.46\textwidth}
        \includegraphics[width=\linewidth]{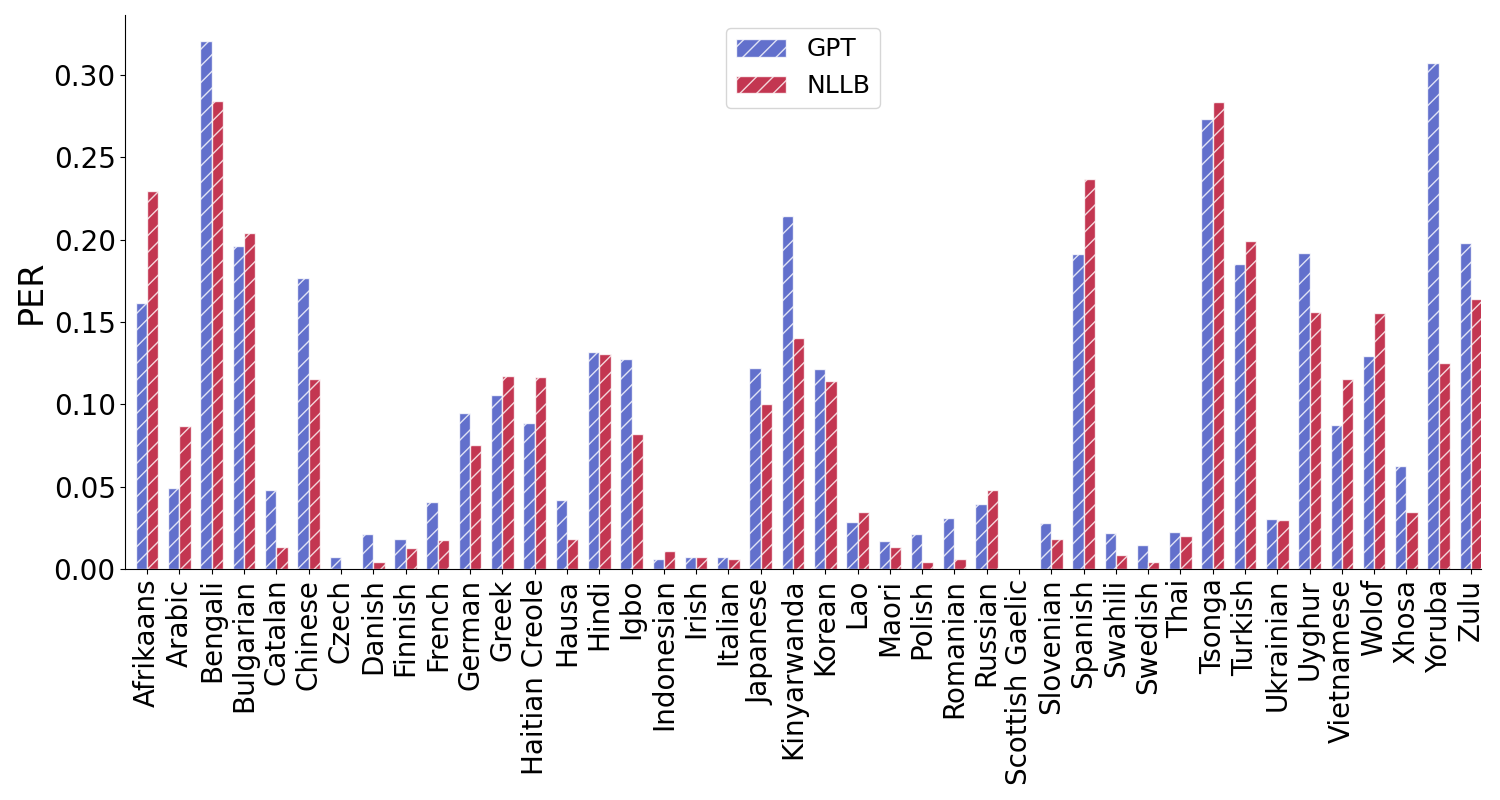}
        \caption{PER}
    \end{subfigure}
    \begin{subfigure}[b]{0.46\textwidth}
        \includegraphics[width=\linewidth]{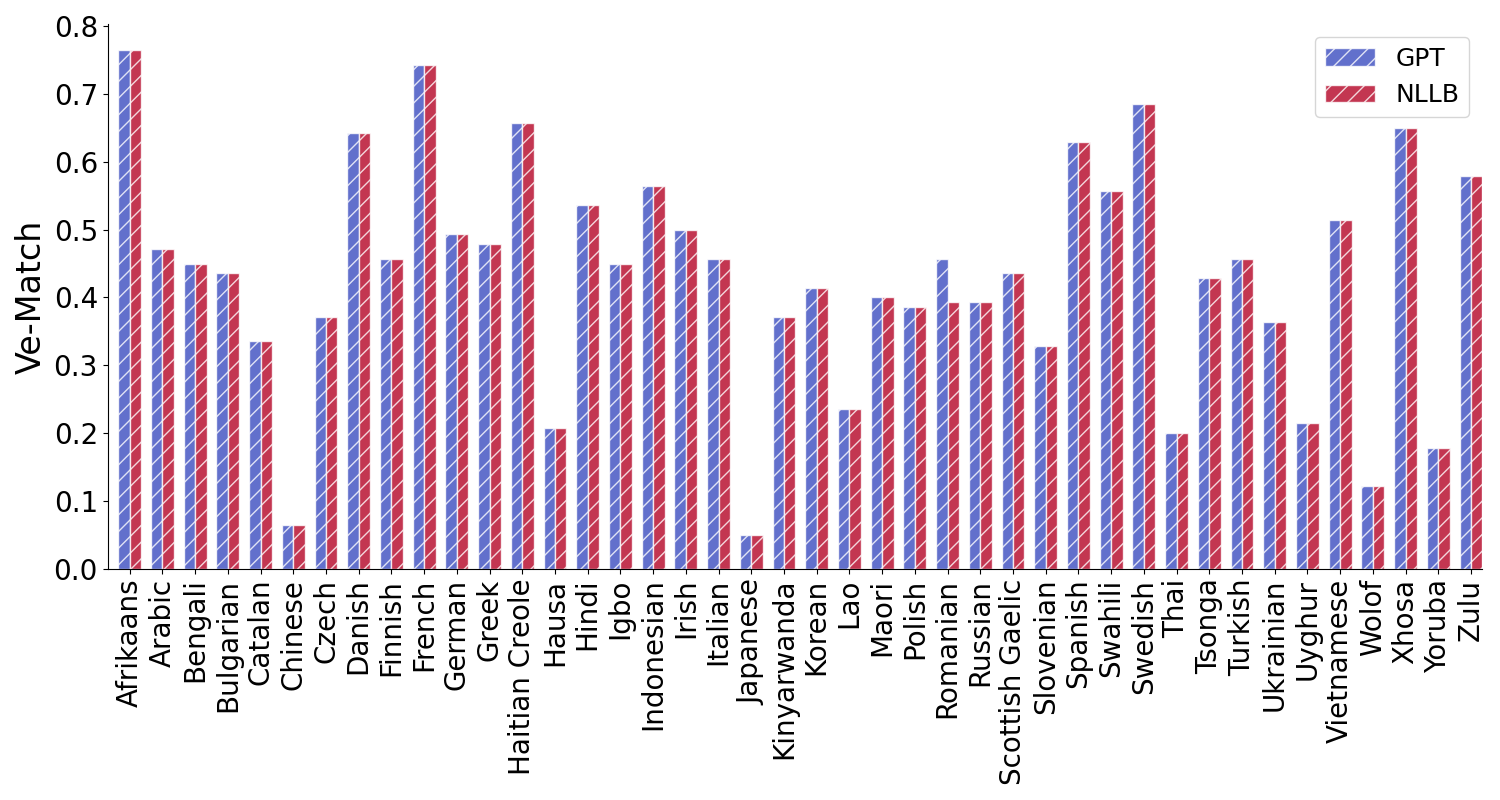}
        \caption{VeMatch}
    \end{subfigure}
    \caption{Translation quality comparison between GPT-models~\cite{singh2025openai, openai2023reasoning} and NLLB~\cite{costa2022no} across $42$ languages. We report lexical similarity, edit distance, semantic fidelity, parameter preservation, and command-verb matching to assess whether translated task instructions preserve robot-relevant semantics and execution parameters.}
    \label{fig:trans_qual}
\end{figure*}

    Modern neural machine translation (NMT) frameworks are trained on vast multilingual corpora to generate high-quality translations~\cite{robinson-etal-2023-chatgpt, jiao2023chatgpt}.
    In addition to task instructions provided by native speakers, we obtained the remaining non-English language task instructions by utilising GPT models~\cite {singh2025openai,openai2023reasoning} for interlingual translations. We made this choice to accommodate under-resourced and vulnerable languages that are inconsistently and insufficiently supported by widely used multilingual NMT frameworks such as Google's MNMT~\cite{johnson2017google} or NLLB~\cite{costa2022no}, e.g., Cherokee, Bislama and African Pidgin. Because translation quality can influence subsequent command parsing and action execution, we therefore conducted a separate evaluation of the GPT models' translation output against NLLB-based reference translations across $42$ languages (see Fig.~\ref{fig:trans_qual}), in order to systematically validate the quality of the interlingual translations.

    We employed multidimensional empirical methods to measure the lexical similarity, semantic fidelity, and safety scores. Specifically, we adopted the BLEU~\cite{papineni2002bleu} metric to assess the lexical/syntactical similarities through n-gram precision. Further, we utilised the translation edit rate (TER)~\cite{snover2006study} metric to quantify the edits required to align the translations with the reference. For semantic fidelity, we employed the BERTScore~\cite{Zhang2020BERTScore} metric to compare meaning. Additionally, we defined parameter error rates (PER) to measure preservation of numerical values and command parameters, and verb-matching accuracy (VeMatch) as a simple diagnostic of command-verb preservation.

    Formally, we considered the input data comprising the source instruction $x_i$ and the translated instruction $y_i$ in the target language $\ell_i$. First, we aligned the data $\{x_i, \ell_i\}$ to produce a unified dataset: \(\mathcal{D}_\mathrm{txn} = \{(x_i, \; \ell_i, \; y^{\mathrm{GPT}}_i, \; y^{\mathrm{NLLB}}_i)\}^N_{i = 1}\), where $y^{\mathrm{GPT}}_i$ is the GPT-models~\cite{openai2023reasoning,singh2025openai} translated instruction (herein referred to as the hypothesis, $\mathcal{H}_{\mathrm{txn}}$) and $y^{\mathrm{NLLB}}_i$ is the reference NLLB~\cite{costa2022no} translations, $\mathcal{R}_\mathrm{txn}$. 
    Since different languages exhibit varying syntactic and morphological features, tokenisation is crucial to maintain consistent scoring criteria. Thus, for BLEU~\cite{papineni2002bleu} and TER~\cite{snover2006study} metrics, we tokenised the instructions using per-language MosesTokenizer~\cite{koehn2007moses} to ensure consistent lexical segmentation across the languages, and for BERTScore~\cite{Zhang2020BERTScore}, we utilised the native subword multilingual tokeniser of \textit{bert-base-multilingual-cased} to remain consistent with the model's pre-training. Therefore, given the reference $\mathcal{R}_\mathrm{txn}$ and the hypothesis $\mathcal{H}_{\mathrm{txn}}$, tokenised into sequences of tokens, we compute the lexical similarity as follows:
\begin{equation}\label{eqn:bleu}
  \mathrm{BLEU}(\mathcal{R}_{\mathrm{txn}}, \;  \mathcal{H}_{\mathrm{txn}}) = \mathrm{BP} \times \exp \left( \sum_{n=1}^{4} \omega_n \log p_n \right),
\end{equation}
    where $p_n$ denotes the modified $n-$gram precision, $\omega_n$  are the \(n\)-gram weights, and \(\mathrm{BP}\) is the brevity penalty to avoid overly short outputs. Further, we compute the TER as:
\begin{equation}\label{eqn:ter}
  \mathrm{TER}(\mathcal{R}_{\mathrm{txn}}, \mathcal{H}_{\mathrm{txn}}) = \frac{\mathrm{edits}(\mathcal{H}_{\mathrm{txn}}\rightarrow \mathcal{R}_{\mathrm{txn}})}
  {|\mathcal{R}_{\mathrm{txn}}|},
\end{equation}
    where \textit{edits} include insertions, deletions, substitutions, and shifts. For more details about this metric, we refer the reader to the following works~\cite{papineni2002bleu, snover2006study}.
    For semantic closeness, we compute the BERTScore. In principle, BERTScore calculates the contextual embeddings through a pre-trained multilingual BERT model~\cite{devlin2019bert} by comparing the embeddings of tokens in $\mathcal{R}_{\mathrm{txn}}$ with $\mathcal{H}_{\mathrm{txn}}$. Let these sequence of embeddings be denoted as $\mathbf{E}(\mathcal{R}_{\mathrm{txn}})$ and $\mathbf{E}(\mathcal{H}_{\mathrm{txn}})$. Thus, we compute the final score by aligning the tokens across both sequences with a pairwise matching strategy as follows:
\begin{equation}\label{eqn:bertscore}
\begin{split}
   F_{\mathrm{BERT}}
   =
   2\frac{P_{\mathrm{BERT}}R_{\mathrm{BERT}}}
   {P_{\mathrm{BERT}}+R_{\mathrm{BERT}}},\\
   P_{\mathrm{BERT}}
   =
   \frac{1}{|\mathcal{H}_{\mathrm{txn}}|}
   \sum_{h_t\in\mathcal{H}_{\mathrm{txn}}}
   \max_{r_t\in\mathcal{R}_{\mathrm{txn}}}
   \cos(\mathbf{E}(h_t),\mathbf{E}(r_t)),\\
   R_{\mathrm{BERT}}
   =
   \frac{1}{|\mathcal{R}_{\mathrm{txn}}|}
   \sum_{r_t\in\mathcal{R}_{\mathrm{txn}}}
   \max_{h_t\in\mathcal{H}_{\mathrm{txn}}}
   \cos(\mathbf{E}(r_t),\mathbf{E}(h_t)),
   \end{split}
\end{equation}
    where $\mathbf{E}(h_t)$ and $\mathbf{E}(r_t)$ are the embeddings of tokens in the hypothesis and reference, respectively.

    To assess parameter preservation, let \(P(\mathcal{R}_{\mathrm{txn}})\) and \(P(\mathcal{H}_{\mathrm{txn}})\) denote extracted numerical or symbolic command parameters from the reference and hypothesis. We define PER as:
\begin{equation}\label{eqn:per}
    \text{PER}(\mathcal{R}_{\mathrm{txn}}, \mathcal{H}_{\mathrm{txn}}) = \begin{cases}
    \frac{\sum_{i = 1}^{k} \delta[P(\mathcal{R}_{\mathrm{txn}})_i \neq  P(\mathcal{H}_{\mathrm{txn}})_i]}{\mid P(\mathcal{R}_{\mathrm{txn}})\mid}, 
    & \text{ if } K_1, \\
    1,
    & \text{ if } K_2, \\
    0, 
    & \text{ if } K_3,
    \end{cases}
\end{equation}
    where $\delta[\cdot]$ is an indicator function that ensures that crucial numeric values or directives remain intact after translation and $k = \min(|P(\mathcal{R}_{\mathrm{txn}})|, |P(\mathcal{H}_{\mathrm{txn}})|)$,  $K_1 \rightarrow \; \mid P(\mathcal{R}_{\mathrm{txn}}) \mid \; > \; 0,\; K_2 \rightarrow \; \mid P(\mathcal{R}_{\mathrm{txn}})\mid \; = \;0 \text{ and } \mid P(\mathcal{H}_{\mathrm{txn}})\mid \; > \; 0$, and $K_3 \rightarrow \; \mid P(\mathcal{R}_{\mathrm{txn}}) \mid  \; = \; \mid P(\mathcal{H}_{\mathrm{txn}}) \mid \; = \;0$.
    Finally, we compute the verb matching (VeMatch) accuracy by assessing whether the first token in the tokenised list for both the reference and the hypothesis is identical as follows:
\begin{equation}\label{eqn:vma}
    \text{VeMatch}(\mathcal{R}_{\mathrm{txn}}, \mathcal{H}_{\mathrm{txn}}) = \begin{cases}
1, & \text{head}(\mathcal{R}_{\mathrm{txn}}) = \text{head}(\mathcal{H}_{\mathrm{txn}}), \\
0, & \text{otherwise}.
\end{cases}
\end{equation}
    
    Fig.~\ref{fig:trans_qual} shows the results obtained. Both the GPT models~\cite {singh2025openai,openai2023reasoning} and NLLB~\cite{costa2022no} exhibit comparable aggregate translation quality, but with different strengths.
    GPT-models~\cite{singh2025openai, openai2023reasoning} achieved a marginally higher BLEU score (\(\mu=0.343\) vs. \(0.341\)), whereas NLLB~\cite{costa2022no} achieved lower TER (\(\mu=0.513\) vs. \(0.556\)) and lower PER (\(\mu=0.084\) vs. \(0.095\)). Both models achieved nearly identical semantic preservation, with BERTScore approximately \(0.874\), and similar verb-matching accuracy around \(0.430\). The correlation structure is also consistent across the models: BLEU and TER were strongly negatively correlated (\(r\approx-0.88\) to \(-0.91\)), BLEU and BERTScore were moderately positively correlated (\(r\approx0.73\) to \(0.74\)), and PER showed weak correlation with the other metrics (\(|r|<0.12\)). Low VeMatch scores in languages such as Yoruba, Wolof, Chinese, and Japanese reflect the limitation of the first-token heuristic in languages with different word order, morphology, or tokenisation behaviour.

\subsection{Performance Metrics}\label{sec:performance-metrics}
    We conducted a comprehensive empirical evaluation of ReLI along two primary dimensions: quantitative and qualitative.  On the quantitative side, we assessed: (i) the accuracy and robustness of multilingual instruction parsing, (ii) the reliability of action execution, and (iii) the latency of robot responses. Qualitatively, we investigated human–robot interaction dynamics and analysed the subjective feedback obtained from human evaluators. The subsequent sections report our findings and provide an in-depth discussion of their broader implications.

\textbf{Instruction parsing accuracy (IPA).}
    We quantify the accuracy with which ReLI maps natural language commands $c_n$ to the targeted robot-action sequence. Given a predicted action sequence \(\hat{\mathcal{A}}_n\) and ground-truth sequence \(\mathcal{A}_n\), we compute: $\mathrm{IPA} = \frac{1}{N}\sum_{n=1}^{N} \delta\,(\mathcal{S}_{\mathrm{IPA}}(\mathcal{A}_n\,,\, \hat{\mathcal{A}}_n)\, \geq\, \gamma)$, where $\gamma = 0.9$ represents the correctness threshold. The composite scoring function $\mathcal{S}_{\mathrm{IPA}}$ integrates both semantic alignment and parameter correctness through weighted fusion: $\mathcal{S}_{\mathrm{IPA}} = w_1.\mathcal{S}_{\mathrm{BERT}} + w_2.\mathcal{S}_{\mathrm{PER}}, \;
    w_1=0.4,\; w_2=0.6$, where \(\mathcal{S}_{\mathrm{BERT}}\) is the BERTScore~\cite{Zhang2020BERTScore}  \(F_1\) alignment between the predicted and ground-truth action sequences, while \(\mathcal{S}_{\mathrm{PER}}\) measures the correctness of extracted action parameters such as coordinates, distances, angles, and velocities. In our experiments, we weight the parameter correctness more heavily because numerical or spatial errors can make a semantically plausible command operationally incorrect.

\textbf{Task success rate (TSR).}
    We used the TSR to measure the proportion of trials in which the robot completes the intended task successfully. For a task \(n \in \{1,\dots, N\}\), we compute: \(\mathrm{TSR} = \frac{1}{N}\sum_{n=1}^{N} \delta_\mathrm{task}(\hat{\mathcal{A}}_n, \mathcal{A}_n)\), where \(\delta_\mathrm{task}(\cdot)\) is a task-specific success indicator. For navigation, success requires the robot to reach the goal within the configured spatial tolerance, e.g., \(\pm0.2\,\mathrm{m}\). For query-oriented tasks, success requires the robot to return the requested information. Minor deviations in speed, travelled distance, or stopping pose are accepted only when they remain within the configured execution tolerance and do not alter the intended task outcome.

\textbf{Average response time (ART).}
    We measure the latency from command issuance to the robot's response with the ART metric. Formally, we compute: \(\mathrm{ART} = \frac{1}{N}\sum_{n=1}^{N}(t_{n}^{\mathrm{res}} - t_{n}^{\mathrm{ins}})\), where \(t_{n}^{\mathrm{ins}}\) is the command timestamp and \(t_{n}^{\mathrm{res}}\) is the timestamp at which the robot begins planning and executing the parsed action sequence.

\subsection{Quantitative Results}\label{sec:quant}
    Tables~\ref{tab:hrl},~\ref{tab:lrl}, and~\ref{tab:end} show ReLI's performance across high-resource, low-resource, and vulnerable language groups, respectively.  Across the full benchmark, ReLI demonstrated strong multilingual robustness across diverse language families and resource tiers, from the mainstream Indo-European to the less-documented creoles and vernaculars.  Performance was consistently high, with $\mathrm{IPA}$ exceeding $85\%$ in nearly all the reported languages, and $\mathrm{TSR}$ surpassing \(83\%\). Importantly, $\mathrm{ART}$ remained stable between $2.1$s and $2.6$s for most languages, even for highly vulnerable ones.

\textbf{High-resource languages.}
    As shown in Table~\ref{tab:hrl}, ReLI performed best on high-resource languages, due to their substantial representation in multilingual training corpora. English and Spanish achieved near-perfect results, with over \(99\%\) IPA and TSR. French, German, Finnish, and Thai also remained above \(97\%\) for IPA and TSR. Arabic, Chinese, Japanese, Swahili, Turkish, and Korean showed lower but still strong performance. The lower scores for some non-Latin or morphologically complex languages are associated with script handling, tokenisation, and language-specific parsing complexity. Korean produced the highest ART among high-resource languages (\(2.55\,\mathrm{s}\)), consistent with the additional processing overhead from Hangul segmentation and agglutinative morphology.
\begin{table*}
    \caption{ReLI's performance on high-resource languages. Accuracies are averaged, and standard deviations are within $\pm1.9$. See Appendix~\ref{append:a} for more details.}\label{tab:hrl}
\begin{tabular*}{\textwidth}{@{\extracolsep\fill}lcccccccc}
\toprule%
& \multicolumn{7}{@{}c@{}}{\underline{\textbf{Indo-European}}} & \multicolumn{1}{@{}c@{}}{\underline{\textbf{Sin-Ti}}} \\
\textbf{Lang.} & English & Spanish & French & German & Hindi & Russian &Portug.&Chinese\\
\cmidrule{2-9}
\textbf{IPA} (\%)  & 99.3 & 99.1 & 98.8 & 97.7 & 93.8 & 96.2 & 96.9 
& 93.8\\
\textbf{TSR} (\%)  & 99.2 & 99.0 & 98.6 &  97.5 & 93.6 &  96.1 & 96.8 
& 93.7 \\
\textbf{ART} (s)  & 2.11 & 2.12 & 2.13& 2.14 & 2.19 & 2.15 & 2.15 
& 2.13 \\
& \multicolumn{1}{@{}c@{}}{\underline{\textbf{Afr-As}}} & \multicolumn{1}{@{}c@{}}{\underline{\textbf{Japonic}}} & \underline{\textbf{Nig-Co}} & \underline{\textbf{Austr.}} & \underline{\textbf{Turkic}} & \underline{\textbf{K-Dai}} & \underline{\textbf{Kor.}} & \underline{\textbf{Uralic}}\\
\textbf{Lang.} & Arabic & Japanese & Swahili & Malay & Turkish & Thai & Korean & Finnish \\
\cmidrule{2-9}
\textbf{IPA} (\%)  & 92.3 & 94.6 & 93.1 & 95.4 & 93.8 & 97.7 & 90.0 & 98.1\\
\textbf{TSR} (\%)  & 92.1 & 94.4 & 92.9 & 95.2 & 93.7 & 97.6 & 90.0 & 98.1\\
\textbf{ART} (s)  & 2.27 & 2.18 & 2.20 & 2.17 & 2.18 & 2.16 & 2.55 & 2.15\\
\bottomrule
\vspace{-7.5pt}
\end{tabular*}
\textcolor{blue}{\textbf{Legends}}: 
\textbf{Sin-Ti} \(\rightarrow\) Sino-Tibetan.
 \textbf{Afr-As} \(\rightarrow\) Afro-Asiatic.
\textbf{Nig-Co} \(\rightarrow\) Niger-Congo.
\textbf{Austr.} \(\rightarrow\) Austronesian.
 \textbf{K-Dai} \(\rightarrow\) Kra-Dai.
\textbf{Kor.} \(\rightarrow\) Koreanic.
\end{table*}

\textbf{Low-resource languages.}
    Table~\ref{tab:lrl} shows that ReLI retained strong performance across many low-resource languages. Irish, Sicilian, Shona, Yoruba, Igbo, and Javanese achieved IPA and TSR values above \(95\%\). Lower results were observed for Serbian, Fijian, Burmese, Tibetan, and Tamil, where IPA and TSR fell below the strongest HRL values. ART remained within \(2.12\)-\(2.76\,\mathrm{s}\), with the largest delays observed in languages that require more difficult script processing, tokenisation, or translation input handling. Overall, the results indicate that ReLI can ground instructions in a broad range of low-resource settings, although performance remains sensitive to language representation and structure.
    
\begin{table*}[htp]
\caption{ReLI's performance on low-resource languages. Accuracies are averaged, and standard deviations are within $\pm1.9$. See Appendix~\ref{append:a} for more details.}\label{tab:lrl}
\begin{tabular*}{\textwidth}{@{\extracolsep\fill}lcccccccc}
\toprule
& \multicolumn{4}{@{}c@{}}{\underline{\textbf{Indo-European}}} & \multicolumn{2}{@{}c@{}}{\underline{\textbf{Afro-Asiatic}}} & \multicolumn{2}{@{}c@{}}{\underline{\textbf{Niger-Congo}}} \\
\textbf{Lang.} & Irish & Serbian & Faroese & Sicilian & Hausa & Amharic & Shona & Igbo \\
\cmidrule{2-9}
\textbf{IPA} (\%) & 97.7 & 87.7 & 94.6 & 96.5 & 91.5 & 93.1 & 96.9 & 95.4 \\
\textbf{TSR} (\%) & 97.5 & 87.7 & 94.5 & 96.3 & 91.4 & 93.0 & 96.8 & 95.3 \\
\textbf{ART} (s) & 2.17 & 2.76 & 2.49 & 2.20 & 2.23 & 2.31 & 2.22 & 2.24 \\
& \multicolumn{1}{@{}c@{}}{\underline{\textbf{Nig-Co}}} & \multicolumn{2}{@{}c@{}}{\underline{\textbf{Austronesian}}} & \multicolumn{1}{@{}c@{}}{\underline{\textbf{Kra-Dai}}} & \multicolumn{1}{@{}c@{}}{\underline{\textbf{Quechua}}} & \multicolumn{2}{@{}c@{}}{\underline{\textbf{Sino-Tibetan}}} & \multicolumn{1}{@{}c@{}}{\underline{\textbf{Drav}}} \\
\textbf{Lang.} & Yoruba & Fijian & Javanese & Lao & Quechua & Burmese & Tibetan & Tamil\\
\cmidrule{2-9}
\textbf{IPA} (\%) & 96.2 & 90.8 & 96.9 & 93.9 & 92.3 & 90.2 & 87.7 & 91.5\\
\textbf{TSR} (\%) & 96.0 & 90.6 & 96.9 & 93.7 & 92.1 & 90.0 & 87.6 & 91.5\\
\textbf{ART} (s) & 2.17 & 2.29 & 2.12 & 2.32 & 2.22 & 2.74 & 2.53 & 2.71\\
\bottomrule
\vspace{-7.5pt}
\end{tabular*}
\textcolor{blue}{\textbf{Legends}}: 
\textbf{Nig-Co} \(\rightarrow\) Niger-Congo.
 \textbf{Drav} \(\rightarrow\) Dravidian.
\end{table*}

\textbf{Vulnerable languages.}
    Table~\ref{tab:end} reports results for vulnerable languages, including creoles, vernaculars, and languages with limited computational support.  ReLI achieved strong performance for Nigerian Pidgin, Tok Pisin, Haitian Creole, Cornish, Chuukese, Hmong, and Chuvash. Nigerian Pidgin reached \(98.1\%\) IPA and \(97.9\%\) TSR, while Hmong achieved \(97.7\%\) IPA and \(97.6\%\) TSR. Bislama, Tiv, Acholi, Breton, Cherokee, and Aramaic showed lower but still effective performance. The lower scores and higher response times for several vulnerable languages indicate that cross-lingual grounding remains constrained by language coverage, orthographic consistency, and the availability of reliable native command resources.
    
\begin{table*}[htp]
\caption{ReLI's performance on vulnerable languages. Accuracies are averaged, and standard deviations are within $\pm1.9$. See Appendix~\ref{append:a} for more details.}
\label{tab:end}
\begin{tabular*}{\textwidth}{@{\extracolsep\fill}lcccccccc}
\toprule
& \multicolumn{4}{@{}c@{}}{\underline{\textbf{Creoles}}} & \multicolumn{3}{@{}c@{}}{\underline{\textbf{Indo-European}}}\\
\textbf{Lang.} & Nig. Pidg. & Tok Pisin & Bislama & Haitian & Ossetian & Breton & Cornish \\
\cmidrule{2-8}
\textbf{IPA} (\%) & 98.1 & 95.0 & 91.9 & 96.2 & 94.2 & 92.3 & 95.4  \\
\textbf{TSR} (\%) & 97.9 & 94.8 & 91.7 & 96.1 & 94.0 & 92.1 & 95.2  \\
\textbf{ART} (s) & 2.14 & 2.21 & 2.38 & 2.33 & 2.23 & 2.49 & 2.71  \\
& \multicolumn{1}{@{}c@{}}{\underline{\textbf{Nig-Co}}} & \multicolumn{1}{@{}c@{}}{\underline{\textbf{Iroq.}}} & \underline{\textbf{Austr.}} & \underline{\textbf{Hm-Mi}} & \underline{\textbf{Turkic}} & \underline{\textbf{Nilo-Sa}} & \underline{\textbf{Afro-As}} \\
\textbf{Lang.} & Tiv & Cherokee & Chuukese & Hmong & Chuvash & Acholi & Aramaic\\
\cmidrule{2-8}
\textbf{IPA} (\%) & 91.5 & 93.1 & 95.8 & 97.7 & 95.4 & 91.5 & 93.1 \\
\textbf{TSR} (\%) & 91.3 & 92.9 & 95.7 & 97.6 & 95.2 & 91.3 & 93.0 \\
\textbf{ART} (s) & 2.67 & 2.53 & 2.26 & 2.28 & 2.23 & 2.57 & 2.55 \\
\bottomrule
\vspace{-7.5pt}
\end{tabular*}
\textcolor{blue}{\textbf{Legends}}:
\textbf{Nig. Pidg.} \(\rightarrow\) Nigerian Pidgin.
\textbf{Nig-Co} \(\rightarrow\) Niger-Congo.
\textbf{Iroq.} \(\rightarrow\) Iroquoian. 
\textbf{Austr.} \(\rightarrow\) Austronesian. 
\textbf{Hm-Mi} \(\rightarrow\) Hmong-Mien. 
\textbf{Nilo-Sa} \(\rightarrow\) Nilo-Saharan.
\textbf{Afro-As} \(\rightarrow\) Afro-Asiatic.
\end{table*}

\begin{figure}[htp]
    \centering
    \includegraphics[width=1.0\linewidth]{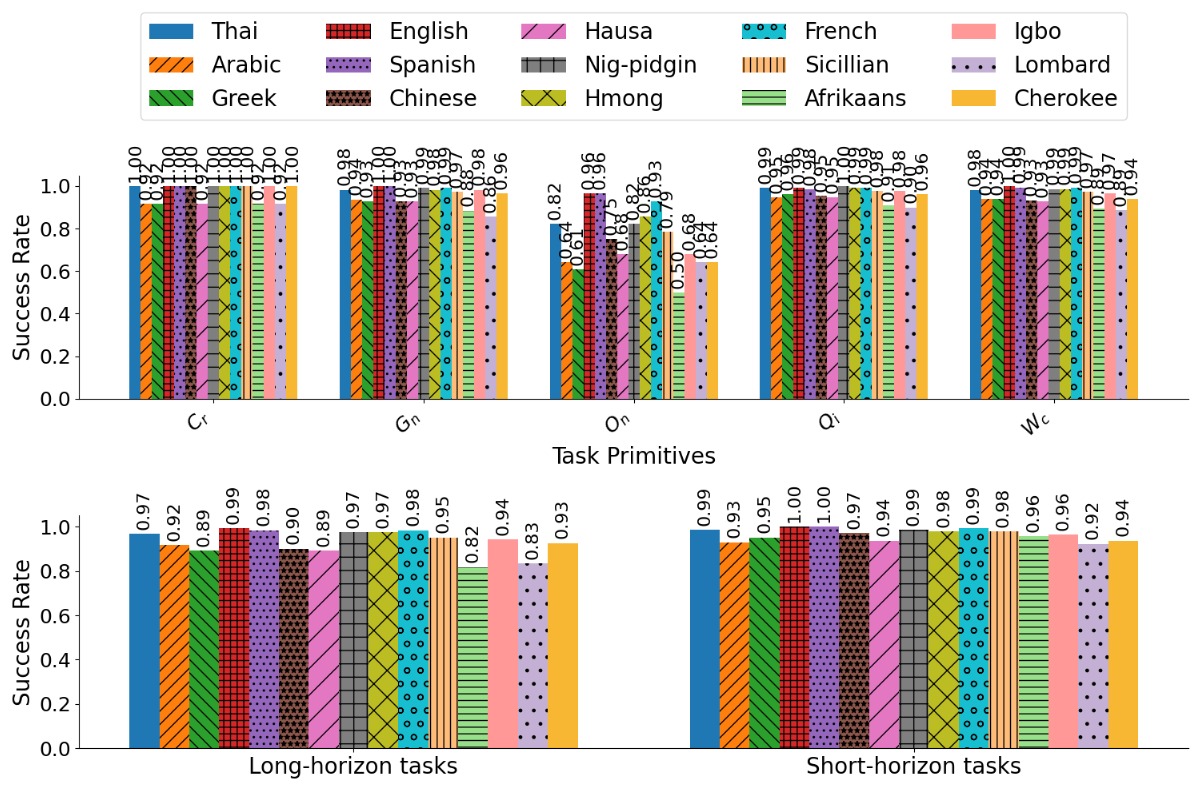}
    \caption{TSR across languages and task categories (top), along with short- and long-horizon performance comparison (bottom). ReLI maintains strong execution performance across most task categories, while failures are observed mostly in object-referenced navigation, ambiguous references, and planner errors.}
    \label{fig:task_primit_lang}
\end{figure}

\textbf{Impact of task horizon and task category.}
    We investigate whether short- and long-horizon instructions impact ReLI's capabilities. For this, we examined ReLI's action execution success rate, based on individual task instructions. Fig.~\ref{fig:task_primit_lang} summarises task success by category and horizon.
    As shown in Fig.~\ref{fig:task_primit_lang} (top), ReLI achieved its highest success in contextual and descriptive reasoning tasks \((C_r)\), where instructions often need to be mapped to an implicit description from a known semantic destination, e.g., associating ``where to cook food'' with the kitchen. 
    Query and information-retrieval tasks \((Q_i)\) also exceeded \(90\%\) success in the evaluated set. Goal-directed navigation \((G_n)\) achieved over \(86\%\) success rate, with most failures attributable to planner errors, partial SLAM inaccuracies, and goal location ambiguity. Object navigation \((O_n)\) was more error-prone because visually similar objects or close confidence scores occasionally caused incorrect reference resolution.
    In terms of task horizons, short-horizon tasks (Fig.~\ref{fig:task_primit_lang}, bottom right) exceeded 90\% success and outperformed long-horizon counterparts (bottom left). This is consistent with the larger number of parsing, grounding, confirmation, and execution steps required for multi-step commands.

\subsection{Qualitative Results}\label{sec:qualitative}
    Although the quantitative evaluation (Section \ref{sec:quant}) showed impressive results, it does not fully capture the qualitative aspects of ReLI's behaviour. To this end, we collected subjective feedback from the $34$ human raters through a 5-point Likert scale, where 1 denotes strongly unfavourable, and 5 denotes strongly favourable. Specifically, we assessed: (i) ReLI's responsiveness, i.e., perceived latency and promptness, (ii) the correctness and naturalness, and (iii) whether users observed a language-induced performance gap during interaction.
    
\begin{figure}[htp]
    \centering
   \includegraphics[width=0.9\linewidth]{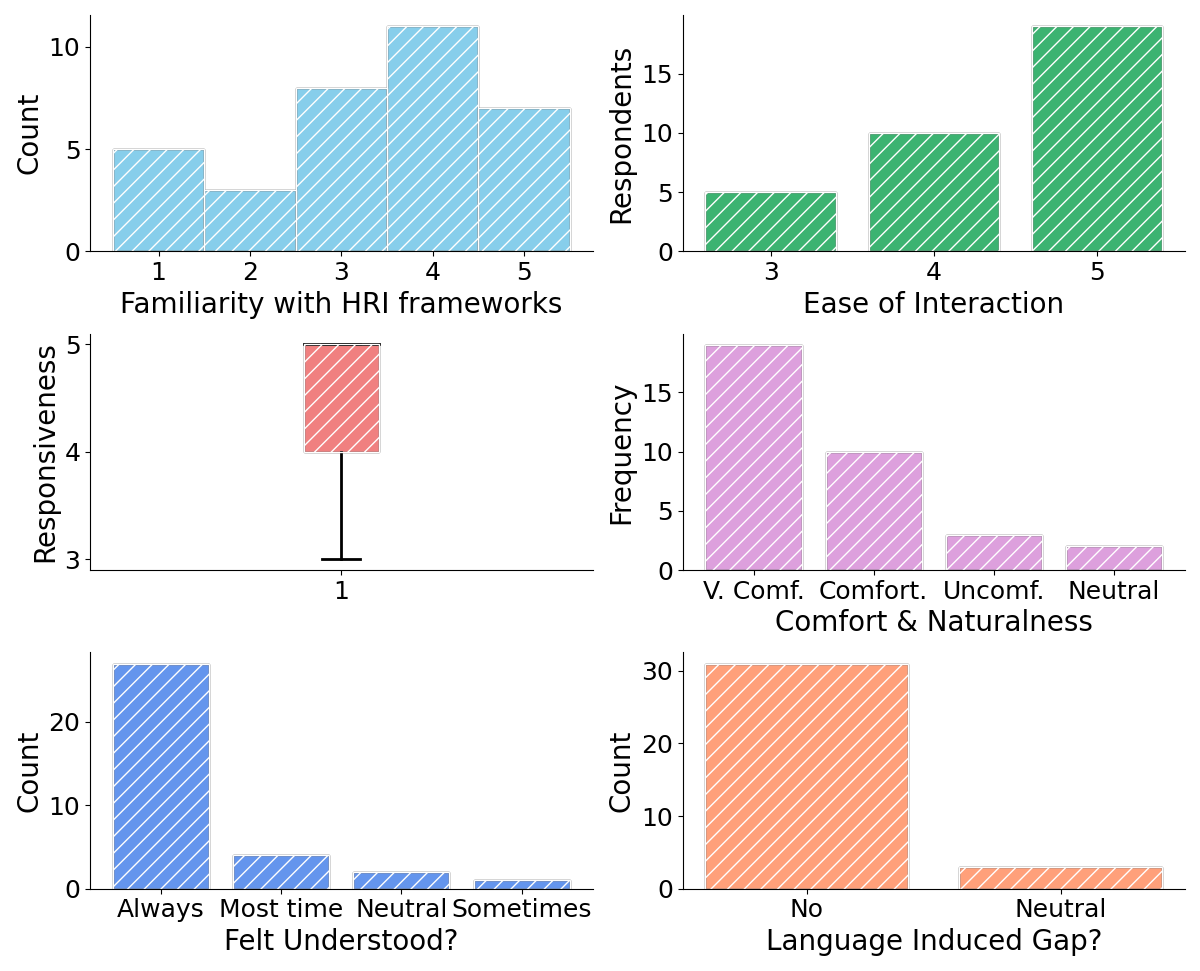}
    \caption{Human raters' feedback on ReLI. Most raters gave favourable scores for ease of interaction, comfort, naturalness, and responsiveness (V. Comf. $\rightarrow$ Very Comfortable, Uncomf. $\rightarrow$ Uncomfortable). More than \(85\%\) reported no observable language-induced performance gap during their interactions.}
    \label{fig:raters}
\end{figure} 
    Fig.~\ref{fig:raters} shows the notable open‐ended qualitative feedback and the corresponding quantitative ratings from the human raters.  With 4 and 5 ratings as the most favourable benchmarks, $75\%$ of the raters reported comfort with the naturalness of their interaction, and over $85\%$ reported high satisfaction with the robot’s responsiveness to their commands. Among the raters who expressed a view on language-induced performance differences, none perceived a language-induced gap with their instruction execution.

    In general, the raters described the interaction as ``intuitive,'' ``cool,'' and ``natural'', with some noting that it felt like talking to a person. However, some recommended extending support for more advanced behaviours, e.g., performing a specialised dance action (e.g., a quadruped robot), given verbal or textual descriptions of the dance style. Further details, including the rater demographics, contributed instructions, and visual examples, are provided in Appendix~\ref{append:b}.

\subsection{Baselines and ablations}
    We conducted baseline comparisons and component ablations to determine whether ReLI's performance arises from its cross-lingual orchestration and grounding mechanisms, or primarily from the multilingual capabilities of the underlying foundation models.
    
\textbf{Baselines.}
    Since there are no directly deployable cross-lingual language-to-action grounding baselines that leverage foundation models under the same experimental setting, we defined three representative alternatives. We compared ReLI against: (i) \textsc{MT$\rightarrow$EN}, which translates each non-English instruction into English before applying the same grounding and execution stack; (ii) \textsc{Direct-MLLM},  which directly generates robot actions from multilingual instructions without ReLI's interaction-conditioned representation or hierarchical semantic parser; and (iii) \textsc{ReLI-Open}, which retains the complete ReLI pipeline but substitutes the closed instruction-reasoning backend with an open-weight model, LLaMA~3~\cite{grattafiori2024llama}. All configurations use identical task instructions, robot observations, action spaces, and execution backends.
    
\begin{table}[htp]
    \centering
    \caption{Comparison of ReLI with representative cross-lingual language-to-action baselines using the native-speaker-provided task instructions (see Table~\ref{tab:raters-demong}).}
    \label{tab:baseline_results}
    \begin{tabular}{lcccc}
        \toprule
        Configuration &
        IPA (\%) $\uparrow$ &
        TSR (\%) $\uparrow$ &
        ART (s) $\downarrow$ &
        Invalid (\%) $\downarrow$\\
        \midrule
        MT$\rightarrow$EN & 89.21 & 86.73 & 2.66 & 15 \\
        Direct-MLLM  & 83.46 & 82.89 & 2.43 & 32 \\
        ReLI-Open  & 80.11 & 79.00 & 2.54 & 19 \\
        Full ReLI & \textbf{94.56} & \textbf{94.32} & \textbf{2.31} & \textbf{2} \\
        \bottomrule
    \end{tabular}
\end{table}
    Table~\ref{tab:baseline_results} shows the results obtained. In particular, the full ReLI pipeline outperformed all the baselines. Compared to \textsc{MT$\rightarrow$EN}, ReLI improved IPA and TSR by $5.35$ and $7.59$ points, respectively, whereas the ART decreased from $2.66$~s to $2.31$~s. This shows that explicit English translation can introduce information loss or semantic distortions that affect downstream action grounding. Relative to \textsc{Direct-MLLM}, ReLI improved IPA by $11.10$ points and TSR by $11.43$ points, with invalid actions reduced from $32\%$ to $2\%$. This also shows that multilingual reasoning alone is insufficient for reliable embodied action generation without a structured language-to-action interface. Further, \textsc{ReLI-Open} remained operational with an open-weight backend but showed a larger performance reduction, with IPA and TSR decreasing by $14.45$ and $15.32$ points relative to Full ReLI. This result shows that ReLI is backend-modular, while also showing that its end-to-end performance remains sensitive to the instruction-reasoning capability of the base foundation model.
    
\textbf{Components ablation.}
    We further ablate four principal ReLI-specific mechanisms to disentangle their individual contributions: the persistent interaction exemplar $\xi_t$, the hierarchical semantic parser $\mathcal{P}$, the grounding reliability gate combining energy rejection and degradation-aware weighting, and the user-confirmation mechanism for multi-step executable plans. Each mechanism was removed independently while retaining the remaining ReLI components unchanged. 
    
    Table~\ref{tab:ablation_results} shows our obtained results.
\begin{table}[htp]
    \centering
    \caption{Ablation of the principal ReLI components. Each variant is evaluated while other components remained active.}
    \label{tab:ablation_results}
    \begin{tabular}{lccc}
    \toprule
    Ablation Variants & IPA(\%) $\uparrow$ &
    TSR (\%) $\uparrow$ & ART (s) $\downarrow$ \\
    \midrule
    w/o implicit conditioning $\xi_t$ & 87.34 & 83.82 & 2.34 \\
    w/o hierarchical parser $\mathcal{P}$ & 78.23 & 69.48 & \textbf{2.29} \\
    w/o grounding reliability gate & 76.45 & 76.31 & 2.30 \\
    w/o user's conf. mechanism & 90.22 & 72.43 & 2.33 \\
    Full ReLI & \textbf{94.56} & \textbf{94.32} & 2.31 \\
    \bottomrule
    \end{tabular}
\end{table}
    All the four mechanisms contribute differently to ReLI's overall performance. Ablating the interaction-conditioned exemplar $\xi_t$ reduced the IPA by $7.22$ points and the TSR by $10.50$ points. This implies that preserving the user's interaction-conditioned linguistic context improves consistent cross-lingual interpretation and downstream execution. Also, ablating the hierarchical parser $\mathcal{P}$ caused the largest drop in task execution, reducing TSR from $94.32\%$ to $69.48\%$ (a $24.84$-point decrease). Excluding the grounding reliability gate produced the largest performance degradation in the grounding accuracy, from $94.56\%$ to $76.45\%$, with an $18.01$-point reduction in TSR. This result shows that uncertainty rejection and degradation-aware weighting are important for preventing unreliable perceptual hypotheses from propagating to downstream control. In contrast, removing the user-confirmation mechanism had a comparatively smaller effect on instruction-level accuracy, which decreased by only $4.34$ points, but caused TSR to fall by $21.89$ points. This divergence is expected because the confirmation mechanism does not primarily affect instruction interpretation; rather, it prevents incorrectly interpreted or unintended multi-step plans from being dispatched for execution.
    Overall, the baseline and ablation results show that ReLI's performance cannot be attributed solely to the multilingual capability of the underlying foundation model. Instead, its interaction conditioning, structured action parsing, reliability-aware visuo-lingual grounding, and execution confirmation contribute complementary improvements at different stages of the language-to-action pipeline.

\section{Conclusion}
    In this paper, we present ReLI, a cross-lingual language-to-action grounding framework for human-robot interaction. We demonstrated empirically that ReLI can abstract free-form spoken or written task instructions into robot-action primitives, ground object and spatial references through visuo-lingual perception, and execute the resulting actions through a robot-control mechanism.
    We evaluated ReLI in both simulated and real-world environments using two robotic embodiments, diverse task categories, and $140$ representative languages that span high-, low-, creole, and vulnerable-language tiers. Across more than $70$K logged multi-turn interactions, ReLI achieved consistently strong instruction parsing accuracy, task success rate, and response latency across the evaluated languages. These results show that ReLI can support practical cross-lingual HRI beyond English-centric interaction settings, common with state-of-the-art language-conditioned HRI frameworks.
    Overall, ReLI provides a technically grounded step toward more accessible multilingual robot interaction. Our future work will focus on on-robot inference and model distillation to decouple ReLI completely from partial cloud dependency while preserving performance robustness. In addition, we plan to investigate adaptive noise cancellation mechanisms to maintain reliable linguistic grounding and perception in acoustically dynamic or noisy operational domains. We believe that ReLI advances inclusive, accessible, and cross-lingual natural language-conditioned HRI to benefit global communities.

\bibliographystyle{ACM-Reference-Format}
\bibliography{sample-base}

\appendix

\section{Generalisation Across Languages}\label{append:a}
    Tables~\ref{tab:quantitative-hrl},~\ref{tab:quantitative-lrl}, and~\ref{tab:quantitative-end} provide the complete language-level benchmark results used in Section~\ref{subsec4B}. Our evaluation covers $140$ representative languages distributed across high-resource, low-resource, creole, vernacular, and vulnerable-language groups. For each language, we evaluated ReLI using the same task taxonomy defined in Section~\ref{subsec4B}, namely zero-shot spatial or goal-directed navigation, low-level motion control, query-oriented interaction, object-referenced navigation, and contextual or descriptive reasoning.
    All results in these tables were obtained using GPT models~\cite{singh2025openai,openai2023reasoning} as the instruction-reasoning backend. The model backend and decoding parameters are reported in Appendix~\ref{append:hyper-param}, while the task categories and representative benchmark instructions are given in Appendix~\ref{append:c}. The tables report instruction parsing accuracy (IPA), task success rate (TSR), and average response time (ART). Reported accuracies are averaged over the benchmark trials for each language, with standard deviations within $\pm1.9$ unless otherwise stated.
    The benchmarking of other languages currently not represented in these tables is currently being conducted, and the results will be regularly updated on the project website\footnote{We are continuously improving ReLI as the multilingual generalisation capabilities of LLMs evolve. Therefore, we have created the following website for updates on ReLI's future development:~\url{https://linusnep.github.io/ReLI/}}.
    
{\scriptsize
\setlength{\tabcolsep}{2.5pt}
\onecolumn
\begin{longtable}{@{}l c c c c c || c c c c c c@{}}
    \captionsetup{font=small}
    \caption{Complete ReLI benchmark results for high-resource languages. IPA denotes instruction parsing accuracy, TSR is the task success rate, and ART is the average response time. Accuracies are averaged over the evaluated task trials, with std. within $\pm1.9$.}
    \label{tab:quantitative-hrl}\\
    \toprule
    \textbf{Language} & \textbf{ISO 639-1} & \textbf{Family} & \textbf{IPA (\%)} & \textbf{TSR (\%)} & \textbf{ART (s)} &
    \textbf{Language} & \textbf{ISO 639-1} & \textbf{Family} & \textbf{IPA (\%)} & \textbf{TSR (\%)} & \textbf{ART (s)} \\
    \midrule
    \endfirsthead
    \caption[]{ReLI's benchmark on high-resource languages $\cdots$ (continued).}\\
    \toprule
    \textbf{Language} & \textbf{ISO 639-1} & \textbf{Family} & \textbf{IPA(\%)} & \textbf{TSR(\%)} & \textbf{ART(s)} &
    \textbf{Language} & \textbf{ISO 639-1} & \textbf{Family} & \textbf{IPA(\%)} & \textbf{TSR(\%)} & \textbf{ART(s)} \\
    \midrule
    \endhead
    \endfoot
    Afrikaans  & af & Indo-Eu & 89.2 & 89.1 & 2.24 & Kazakh     & kk & Turkic    & 90.8 & 90.8 & 2.25 \\
    Albanian   & sq & Indo-Eu & 96.2 & 96.1 & 2.57 & Korean     & ko & Koreanic  & 90.0 & 90.0 & 2.55 \\
    Arabic     & ar & Afro-As & 92.3 & 92.1 & 2.27 & Latvian    & lv & Indo-Eu   & 88.1 & 88.1 & 2.28 \\
    Bengali    & bn & Indo-Eu & 88.5 & 88.5 & 2.54 & Lithuanian & lt & Indo-Eu   & 97.7 & 97.7 & 2.23 \\
    Bosnian    & bs & Indo-Eu & 97.7 & 97.5 & 2.67 & Macedonian & mk & Indo-Eu   & 91.9 & 91.7 & 2.60 \\
    Bulgarian  & bg & Indo-Eu & 90.0 & 90.0 & 2.51 & Malay      & ms & Austro    & 95.4 & 95.2 & 2.17 \\
    Catalan    & ca & Indo-Eu & 96.2 & 96.2 & 2.56 & Maltese    & mt & Afro-As   & 94.6 & 94.4 & 2.22 \\
    Chinese    & zh & Sino-Ti & 93.8 & 93.7 & 2.13 & Persian    & fa & Indo-Eu   & 97.3 & 97.3 & 2.48 \\
    Croatian   & hr & Indo-Eu & 87.7 & 87.6 & 2.34 & Polish     & pl & Indo-Eu   & 97.7 & 97.5 & 2.29 \\
    Czech      & cs & Indo-Eu & 96.9 & 96.7 & 2.33 & Portuguese & pt & Indo-Eu   & 96.9 & 96.8 & 2.15 \\
    Danish     & da & Indo-Eu & 98.1 & 97.9 & 2.24 & Romanian   & ro & Indo-Eu   & 88.5 & 88.5 & 2.49 \\
    Dutch      & nl & Indo-Eu & 96.9 & 96.9 & 2.16 & Russian    & ru & Indo-Eu   & 96.2 & 96.1 & 2.15 \\
    English    & en & Indo-Eu & 99.3 & 99.2 & 2.11 & Sesotho    & st & Niger-Co  & 88.1 & 87.9 & 2.44 \\
    Estonian   & et & Uralic  & 89.2 & 89.0 & 2.55 & Slovak     & sk & Indo-Eu   & 96.9 & 96.7 & 2.21 \\
    Filipino   & tl & Austro  & 94.6 & 94.5 & 2.19 & Slovenian  & sl & Indo-Eu   & 94.6 & 94.4 & 2.13 \\
    Finnish    & fi & Uralic  & 98.1 & 98.1 & 2.15 & Spanish    & es & Indo-Eu   & 99.1 & 99.0 & 2.12 \\
    French     & fr & Indo-Eu & 98.8 & 98.6 & 2.13 & Swahili    & sw & Niger-Co  & 93.1 & 92.9 & 2.20 \\
    German     & de & Indo-Eu & 97.7 & 97.5 & 2.14 & Swedish    & sv & Indo-Eu   & 98.1 & 97.9 & 2.21 \\
    Greek      & el & Indo-Eu & 92.3 & 92.2 & 2.15 & Thai       & th & Kra-Dai   & 97.7 & 97.6 & 2.16 \\
    Hebrew     & he & Afro-As & 96.2 & 96.0 & 2.46 & Tswana     & tn & Niger-Co  & 89.6 & 89.6 & 2.34 \\
    Hindi      & hi & Indo-Eu & 93.8 & 93.6 & 2.19 & Turkish    & tr & Turkic    & 93.8 & 93.7 & 2.18 \\
    Hungarian  & hu & Uralic  & 97.3 & 97.3 & 2.15 & Ukrainian  & uk & Indo-Eu   & 96.2 & 96.0 & 2.18 \\
    Icelandic  & is & Indo-Eu & 93.4 & 93.2 & 2.58 & Uzbek      & uz & Turkic    & 93.8 & 93.8 & 2.46 \\
    Indonesian & id & Austro  & 96.9 & 96.7 & 2.53 & Vietnamese & vi & Austro    & 98.8 & 98.8 & 2.61 \\
    Italian    & it & Indo-Eu & 98.5 & 98.3 & 2.24 & Xhosa      & xh & Niger-Co  & 91.5 & 91.3 & 2.43 \\
    Japanese   & ja & Japonic & 94.6 & 94.4 & 2.18 & Zulu       & zu & Niger-Co  & 96.9 & 96.7 & 2.15 \\
\bottomrule
\end{longtable}
\vspace{-9.5pt}
    \textcolor{blue}{\textbf{Legend:}} 
    \textbf{IPA} \(\rightarrow\) Instruction Parsing Accuracy. 
    \textbf{TSR} \(\rightarrow\) Task Success Rate. 
    \textbf{ART} \(\rightarrow\) Average Response Time. 
    \textbf{Indo-Eu} \(\rightarrow\) Indo-European. 
    \textbf{Afro-As} \(\rightarrow\) Afro-Asiatic. 
    \textbf{Niger-Co} \(\rightarrow\) Niger-Congo. 
    \textbf{Austro} \(\rightarrow\) Austronesian. 
    \textbf{Sino-Ti} \(\rightarrow\) Sino-Tibetan. 
    \textbf{Uralic} \(\rightarrow\) Uralic. 
    \textbf{Turkic} \(\rightarrow\) Turkic. 
    \textbf{Koreanic} \(\rightarrow\) Koreanic. 
    \textbf{Japonic} \(\rightarrow\) Japonic. 
    \textbf{Kra-Dai} \(\rightarrow\) Kra-Dai.
}

{\scriptsize
\setlength{\tabcolsep}{2.5pt}
\begin{longtable}{@{}l c c c c c || c c c c c c@{}}
\captionsetup{font=small}
\caption{Complete ReLI benchmark results for low-resource languages. The table reports instruction parsing accuracy, task success rate, and response latency across language families with more limited computational resources or smaller digital corpora. Accuracies are averaged over the evaluated task trials, with standard deviations within $\pm1.9$.}
\label{tab:quantitative-lrl}\\
\toprule
\textbf{Language} & \textbf{ISO 639-1} & \textbf{Family} & \textbf{IPA (\%)} & \textbf{TSR (\%)} & \textbf{ART (s)} & \textbf{Language} & \textbf{ISO 639-1} & \textbf{Family} & \textbf{IPA (\%)} & \textbf{TSR (\%)} & \textbf{ART (s)} \\
\midrule
\endfirsthead
\caption[]{ReLI's benchmark on low-resource languages $\cdots$ (continued).}\\
\toprule
\textbf{Language} & \textbf{ISO 639-1} & \textbf{Family} & \textbf{IPA(\%)} & \textbf{TSR(\%)} & \textbf{ART(s)} & \textbf{Language} & \textbf{ISO 639-1} & \textbf{Family} & \textbf{IPA(\%)} & \textbf{TSR(\%)} & \textbf{ART(s)} \\
\midrule
\endhead
\bottomrule
\endfoot
Akan         & ak & Niger-Co  & 88.1 & 88.1 & 2.33 &
Amharic      & am & Afro-As   & 93.1 & 93.0 & 2.31 \\
Armenian     & hy & Indo-Eu   & 91.5 & 91.5 & 2.48 &
Azerbaijani  & az & Turkic    & 89.6 & 89.4 & 2.31 \\
Bamb-Dioula  & bm & Niger-Co  & 87.7 & 87.6 & 2.42 &
Belarusian   & be & Indo-Eu   & 95.4 & 95.4 & 2.64 \\
Burmese      & my & Sino-Ti   & 90.2 & 90.0 & 2.74 &
Chamorro     & ch & Austro  & 95.8 & 95.6 & 2.36 \\
Chewa        & ny & Niger-Co  & 92.3 & 92.3 & 2.21 &
Corsican     & co & Indo-Eu   & 97.3 & 97.2 & 2.20 \\
Dzongkha     & dz & Sino-Ti   & 88.1 & 87.9 & 2.48 &
Ewe          & ee & Niger-Co  & 93.8 & 93.6 & 2.50 \\
Faroese      & fo & Indo-Eu   & 94.6 & 94.5 & 2.49 &
Fijian       & fj & Austro  & 90.8 & 90.6 & 2.29 \\
Galician     & gl & Indo-Eu   & 97.7 & 97.6 & 2.25 &
Hausa        & ha & Afro-As   & 91.5 & 91.4 & 2.23 \\
Igbo         & ig & Niger-Co  & 95.4 & 95.3 & 2.24 &
Irish        & ga & Indo-Eu   & 97.7 & 97.5 & 2.17 \\
Javanese     & jv & Austro  & 96.9 & 96.9 & 2.12 &
Kannada      & kn & Dravidian & 88.1 & 87.9 & 2.47 \\
Khmer        & km & Austroas  & 88.5 & 88.5 & 2.49 &
Kikuyu       & ki & Niger-Co  & 89.2 & 89.0 & 2.26 \\
Kinyarwanda  & rw & Niger-Co  & 93.1 & 92.9 & 2.39 &
Kurdish      & ku & Indo-Eu   & 89.6 & 89.4 & 2.64 \\
Kyrgyz       & ky & Turkic    & 92.3 & 92.1 & 2.36 &
Lao          & lo & Kra-Dai   & 93.9 & 93.7 & 2.32 \\
Lingala      & ln & Niger-Co  & 90.2 & 90.0 & 2.14 &
Lombard      & n/a & Indo-Eu  & 88.1 & 87.9 & 2.32 \\ 
Māori        & mi & Austro  & 93.5 & 93.4 & 2.48 &
Malagasy     & mg & Austro  & 87.7 & 87.7 & 2.35 \\
Marshallese  & mh & Austro  & 97.7 & 97.7 & 2.32 &
Mongolian    & mn & Mongolic  & 92.3 & 92.3 & 2.29 \\
Nepali       & ne & Indo-Eu   & 89.2 & 89.2 & 2.23 &
Ndebele      & nr & Niger-Co  & 93.2 & 93.1 & 2.68 \\
Norwegian    & no & Indo-Eu   & 94.2 & 94.2 & 2.19 &
Oromo        & om & Afro-As   & 87.7 & 87.7 & 2.38 \\
Pashto       & ps & Indo-Eu   & 92.7 & 92.7 & 2.45 &
Punjabi      & pa & Indo-Eu   & 93.8 & 93.7 & 2.41 \\
Quechua      & qu & Quechuan  & 92.3 & 92.1 & 2.22 &
S. Gaelic & gd & Indo-Eu & 91.9 & 91.7 & 2.48 \\
Serbian      & sr & Indo-Eu   & 87.7 & 87.7 & 2.76 &
Shona        & sn & Niger-Co  & 96.9 & 96.8 & 2.22 \\
Sicilian     & sc & Indo-Eu   & 96.5 & 96.3 & 2.20 &
Somali       & so & Afro-As   & 96.2 & 96.1 & 2.15 \\
Sundanese    & su & Austro  & 98.1 & 98.1 & 2.42 &
Samoan       & sm & Austro  & 96.9 & 96.8 & 2.71 \\
Tajik        & tg & Indo-Eu   & 90.0 & 89.8 & 2.55 &
Tamil        & ta & Dravidian & 91.5 & 91.5 & 2.71 \\
Tatar        & tt & Turkic    & 91.5 & 91.4 & 2.56 &
Tibetan      & bo & Sino-Ti   & 87.7 & 87.6 & 2.53 \\
Tigrinya     & ti & Afro-As   & 92.7 & 92.7 & 2.41 &
Tongan       & to & Austro  & 96.2 & 96.2 & 2.54 \\
Tsonga       & ts & Niger-Co  & 90.4 & 90.4 & 2.37 &
Turkmen      & tk & Turkic    & 91.5 & 91.4 & 2.77 \\
Twi          & tw & Niger-Co  & 87.7 & 87.7 & 2.44 &
Telugu       & te & Dravidian & 93.8 & 93.7 & 2.36 \\ 
Uyghur       & ug & Turkic    & 96.9 & 96.7 & 2.15 &
Welsh        & cy & Indo-Eu   & 92.7 & 92.6 & 2.41 \\
Wolof        & wo & Niger-Co  & 90.0 & 89.9 & 2.34 &
Yoruba       & yo & Niger-Co  & 96.2 & 96.0 & 2.17 \\
\end{longtable}
\vspace{-9.5pt}
\textcolor{blue}{\textbf{Legends}}: 
\textbf{Indo-Eu} \(\rightarrow\) Indo-European.
\textbf{Afro-As} \(\rightarrow\) Afro-Asiatic.
\textbf{Niger-Co} \(\rightarrow\) Niger-Congo.
\textbf{Austro} \(\rightarrow\) Austronesian.
\textbf{Sino-Ti} \(\rightarrow\) Sino-Tibetan.
\textbf{Austroas} \(\rightarrow\) Austro-Asiatic.
}

\begin{table}[htp]
\centering
\captionsetup{font=small}
\caption{Complete ReLI benchmark results for creoles, vernaculars, and vulnerable-language settings. Accuracies are averaged over the evaluated task trials, with standard deviations within $\pm1.9$.}
\label{tab:quantitative-end}
\scriptsize %
\begin{tabular}{@{}l c c c c c@{}}
\toprule
\textbf{Language} & \textbf{ISO 639-1} & \textbf{Family} & \textbf{IPA (\%)} & \textbf{TSR (\%)} & \textbf{ART (s)}\\
\midrule
Acholi        & n/a  & Nilo-Saharan  & 91.5 & 91.3 & 2.57 \\
Aragonese     & an   & Indo-European  & 91.5 & 91.4 & 2.40 \\
Aramaic       & n/a  & Afro-Asiatic  & 89.1 & 87.3 & 2.59 \\
Bislama       & bi   & Creole   & 91.9 & 91.7 & 2.38 \\
Breton        & br   & Indo-European  & 92.3 & 92.1 & 2.49 \\
Buryat        & n/a  & Mongolic & 92.7 & 92.5 & 2.42 \\
Carolinian    & n/a  & Austronesian & 89.6 & 89.4 & 2.69 \\
Cherokee      & n/a  & Iroquoian     & 83.1 & 82.0 & 2.63 \\
Chuvash       & cv   & Turkic   & 95.4 & 95.2 & 2.23 \\
Chuukese      & n/a  & Austronesian & 95.8 & 95.7 & 2.26 \\
Cornish       & kw   & Indo-European  & 95.4 & 95.2 & 2.71 \\
Haitian Cr.   & ht   & Creole   & 96.2 & 96.1 & 2.33 \\
Hawaiian      & n/a  & Austronesian & 93.8 & 93.7 & 2.56 \\
Hiri Motu     & n/a  & Creole   & 90.0 & 89.8 & 2.72 \\
Hmong         & n/a  & Hmong-Mien & 97.7 & 97.6 & 2.28 \\
Latin         & la   & Indo-European  & 90.4 & 90.2 & 2.67 \\
Manx          & gv   & Indo-European  & 96.5 & 96.3 & 2.34 \\
Mapudungun    & n/a  & Araucani & 88.8 & 88.8 & 2.35 \\
Mien          & n/a  & Hmong-Mien & 90.0 & 89.9 & 2.43 \\
Nig.\ Pidgin  & n/a  & Creole   & 98.1 & 97.9 & 2.14 \\
Ossetian      & os   & Indo-European  & 94.2 & 94.0 & 2.23 \\
Palauan       & n/a  & Austronesian & 88.1 & 88.1 & 2.67 \\
Phoenician    & n/a  & Afro-Asiatic  & 79.2 & 78.1 & 3.54 \\
Pohnpeian     & n/a  & Austronesian & 90.8 & 90.8 & 2.54 \\
Romansh       & rm   & Indo-European  & 93.1 & 93.0 & 2.39 \\
Syriac        & n/a  & Afro-Asiatic  & 89.2 & 89.0 & 3.00 \\
Tiv           & n/a  &Niger-Congo & 90.1 & 90.0 & 2.77 \\
Tok Pisin     & n/a  & Creole   & 95.0 & 94.8 & 2.21 \\
\bottomrule
\end{tabular}
\end{table}

\section{Details of Hyperparameters}\label{append:hyper-param}
    Table~\ref{tab:hyper-param} provides details of the key hyperparameters we utilised in our experiments to obtain the results in Tables~\ref{tab:quantitative-hrl},~\ref{tab:quantitative-lrl}, and~\ref{tab:quantitative-end}. ReLI is a configuration-driven framework: model providers, decoding parameters, perception thresholds, grounding weights, execution limits, and backend interfaces can be specified externally through a configuration file. We adopted this design to improve reproducibility and allow the same language-to-action pipeline to be deployed with different LLMs, perception backends, and robot-control interfaces.

    For the instruction-reasoning pipeline (Section~\ref{subsec3B}), the LLM provider, model name, endpoint, and API key identify the selected language backend. The maximum-token parameter bounds the length of generated responses and action schemas, while the decoding temperature controls response determinism. In our experiments, we used a temperature of $0$ to obtain deterministic action plans and reduce stochastic variation in parsed robot commands. For local or self-hosted models, the endpoint field can be set to a local inference server, such as an Ollama, llama.cpp, or vLLM endpoint.

    For the visuo-lingual grounding pipeline (Section~\ref{subsec3C}), we used the softmax temperature to control the sharpness of the visual-textual reference distribution.  Lower values concentrate probability mass on the highest-scoring object candidates, whereas higher values produce smoother class-reference distributions. 
    For the segmentation models, we employed the mask-quality threshold to filter fragmented or geometrically implausible masks before semantic matching. The detector and segmentation confidence thresholds control the trade-off between recall and false positives. Additionally, we used the degradation sensitivity parameter \(\beta\) to determine how strongly low illumination, blur, occlusion, or other perceptual degradation indicators reduce the grounding score. A higher value (e.g., $\beta > 2.0$) downweights degraded regions more aggressively, and a lower value (e.g., $\beta < 2.0$) applies softer penalties. Further, the weighting coefficients \((\lambda_1,\lambda_2,\lambda_3,\lambda_4)\) control the relative contribution of calibrated reference probability, semantic similarity, detector confidence, and spatial-relation consistency.
    
    For the hyperparameters associated with the action execution mechanism (Section~\ref{subsec3D}) and the interlingual translation models (Appendix~\ref{append:c}), we primarily utilised the default parameter values specific to each model. For further information on parameters related to the ROS navigation planner, observation sources, and additional platform-specific parameters, including camera intrinsics, camera-to-base calibration, depth filtering, monocular-depth fallback, and navigation-controller settings, we refer the reader to the configuration file at ReLI's GitHub repository source code.
\begin{table*}[htp]
    \centering
    \scriptsize
    \captionsetup{font=small}
    \caption{Key configuration parameters used in our experiments. Numeric values correspond to the experimental setting used for our benchmark; backend names and model identifiers can be changed without modifying the core ReLI pipeline.}
    \label{tab:hyper-param}
    \begin{tabular}{@{}l c l c@{}}
    \toprule
    \multicolumn{2}{c}{\textbf{Numeric Parameters}} &
    \multicolumn{2}{c}{\textbf{Backend / Model Parameters}}\\
    \cmidrule(r){1-2}
    \cmidrule(l){3-4}
    \textbf{Parameter} & \textbf{Value} &
    \textbf{Parameter} & \textbf{Value}\\
    \midrule
    LLM max tokens & 500 &
    LLM provider & openai\\
    LLM temperature & 0 &
    LLM model name & GPT-4o\\
    Softmax temperature \(T_{\mathrm{temp}}\) & 0.07 &
    LLM endpoint & --\\
    Mask-quality threshold \(q_{\mathrm{thresh}}\) & 0.60 &
    Detector backend & YOLO-World~\cite{cheng2024yolo} / OWLv2~\cite{minderer2023scaling}\\
    Detector confidence threshold & 0.40 &
    Segmentation backend & SAM~\cite{kirillov2023segment} / SAM3~\cite{carion2025sam}\\
    Segmentation confidence threshold & 0.40 &
    Segmentation checkpoint & sam\_vit\_b\_01ec64.pth\\
    Energy threshold \(e_{\mathrm{thresh}}\) & 0.45 &
    Image-text encoder & CLIP ViT-B/32\\
    Degradation sensitivity \(\beta\) & 1.00 &
    Device preference & cuda\\
    Grounding weights \((\lambda_1,\lambda_2,\lambda_3,\lambda_4)\) &
    \((0.4,0.3,0.2,0.1)\) &
    Default interaction language & English\\
    Navigation tolerance & \(0.20\,\mathrm{m}\) &
    Execution backend & ROS~1 / ROS~2 / Web / Dry-run\\
\bottomrule
\end{tabular}
\end{table*}

\section{Visualisations and Human Raters Demographics}\label{append:b}
\subsection{Qualitative Visualisations}\label{append:qual-vis}
\begin{figure*}[htp]
    \centering
    \includegraphics[width=0.75\linewidth]{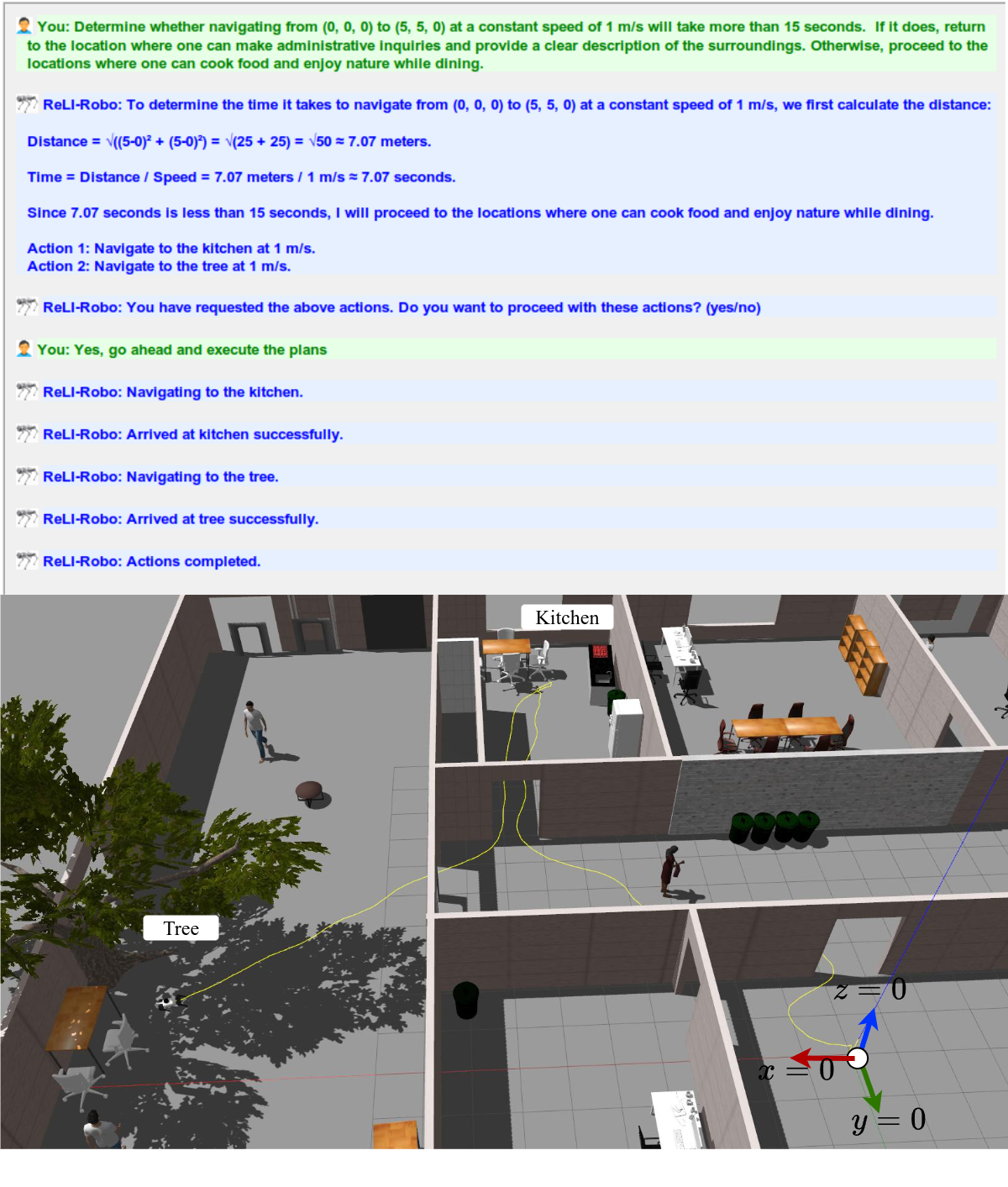}
    \caption{Example of contextual and spatial reasoning in ReLI. The instruction requires coordinate interpretation, conditional navigation logic, semantic destination mapping, and scene description. The example illustrates how ReLI abstracts natural-language instruction into action primitives and returns environment-level feedback after execution.}
    \label{fig:rationale}
\end{figure*}

\begin{figure*}[htp]
    \centering
    \includegraphics[width=\linewidth]{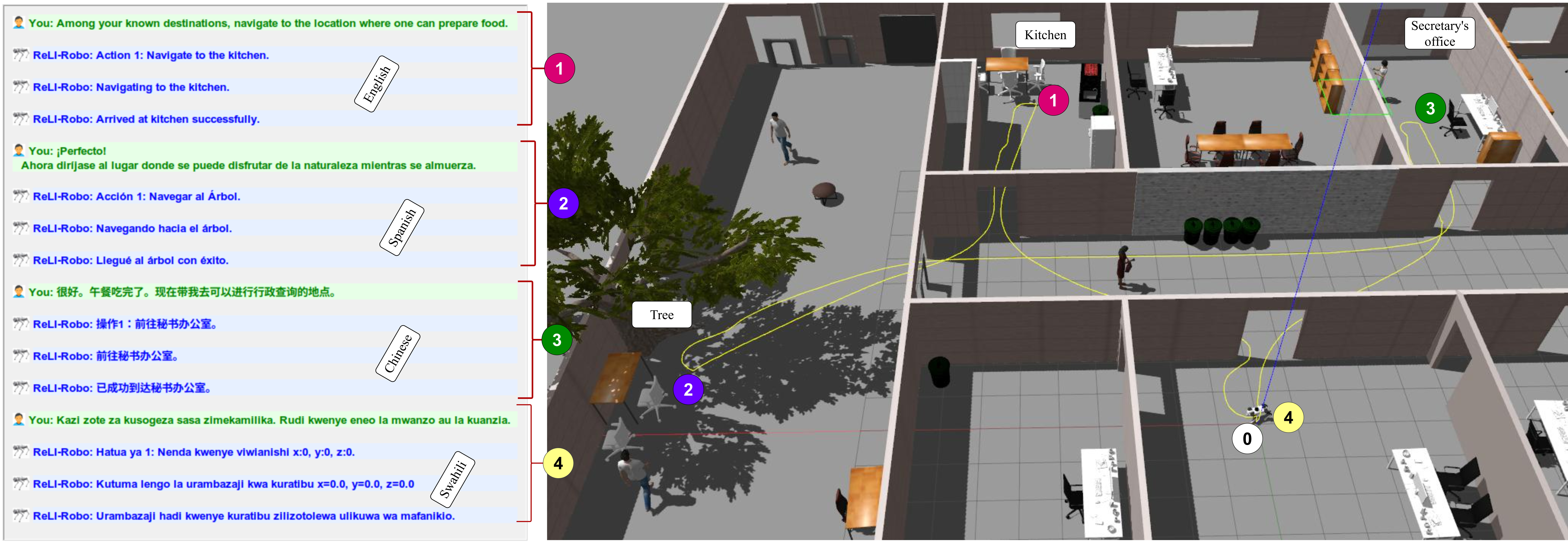}
    \caption{Multi-turn task execution in high-resource languages. The yellow path shows the robot's trajectory. The interaction interface (left) shows the multilingual dialogue history. The robot starts at the origin \((x=y=z=0)\) (0) and receives sequential task instructions in English (1), Spanish (2), Chinese (3), and Swahili (4).
    ReLI dynamically interprets and maps each command to a shared robot-action representation for navigation and contextual destination inference.}
    \label{fig:hrlChat}
\end{figure*}

\begin{figure}[htp]
    \centering
    \includegraphics[width=0.95\linewidth]{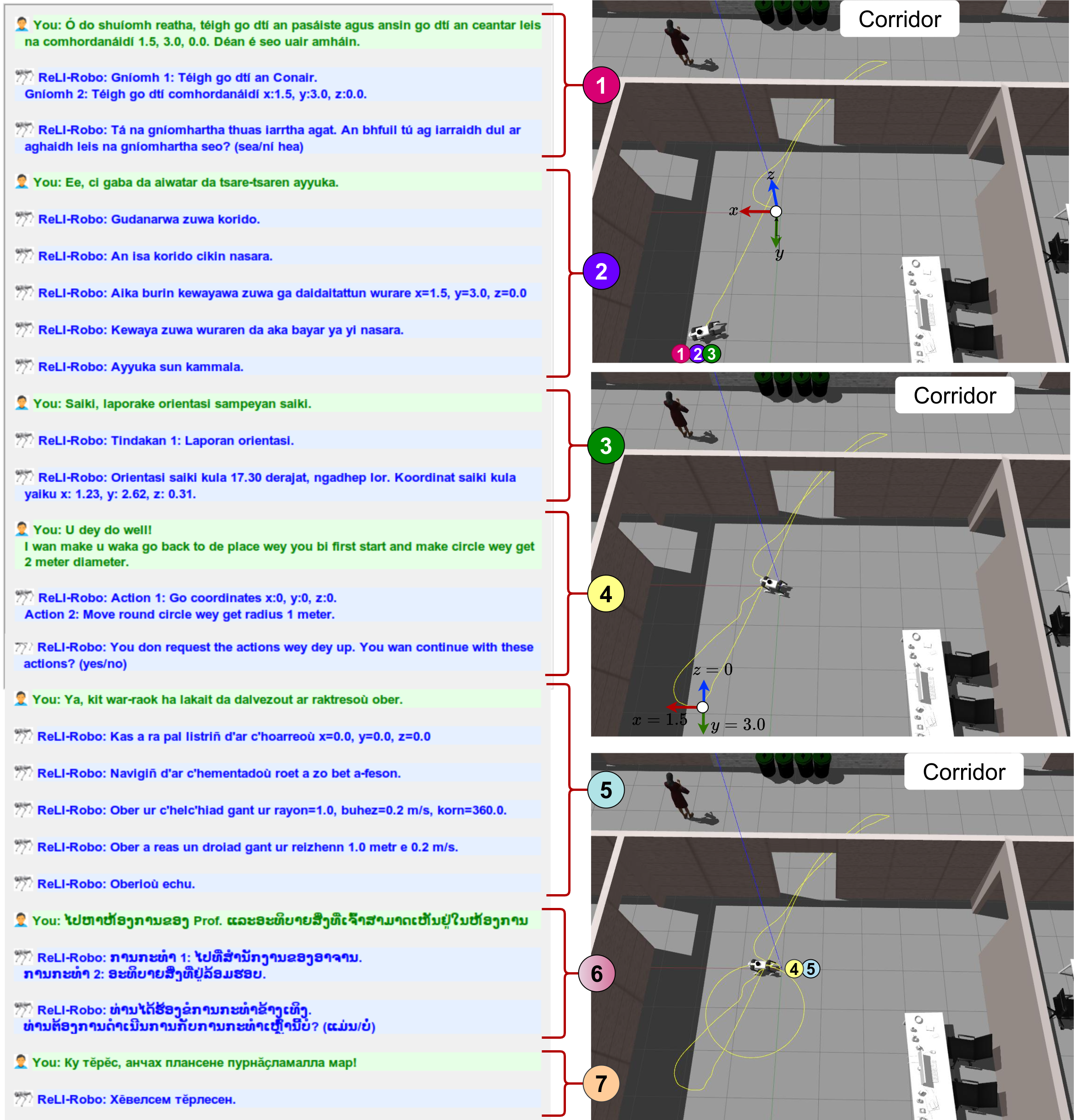}
    \caption{Multi-turn task execution in low-resource and vulnerable languages. The yellow path represents the robot’s trajectory across the sequential task steps. The interaction interface (left) shows the corresponding dialogue. The robot starts at the origin ($x = y = z = 0$).
    1: Instruction in Irish – ``\textbf{Transl.} \textit{From your current location, head to the passageway and then to the area with coordinates 1.5, 3.0, 0.0. Do this only once.}" 
    2: Action approval in Hausa – ``\textbf{Transl.} \textit{Yes, go ahead and execute the action plans.}" 
    3: Instruction in Javanese – ``\textbf{Transl.} \textit{Now, report your current orientation.}"
    4: Instruction in Nigerian Pidgin – ``\textbf{Transl.} \textit{You are doing well! I want you to go back to the place where you first started and make a circle with a diameter of 2 meters.}" 
    5: Action approval in Breton – ``\textbf{Transl.} \textit{Yes, go ahead and execute the action plans.}" 
    6: Instruction in Lao – ``\textbf{Transl.} \textit{Head to the Prof.'s office and describe what you can see in the office.}"
    7: Action rejection in Chuvash – ``\textbf{Transl.} \textit{That's correct, but do not execute the plans!}"
    In these interactions, ReLI show reliable understanding, planning, and control even in languages with limited resources.}
    \label{fig:lrl-vulChat}
\end{figure}

\begin{figure*}[htp]
    \centering
    \begin{subfigure}[b]{1.0\textwidth}
        \includegraphics[width=\linewidth]{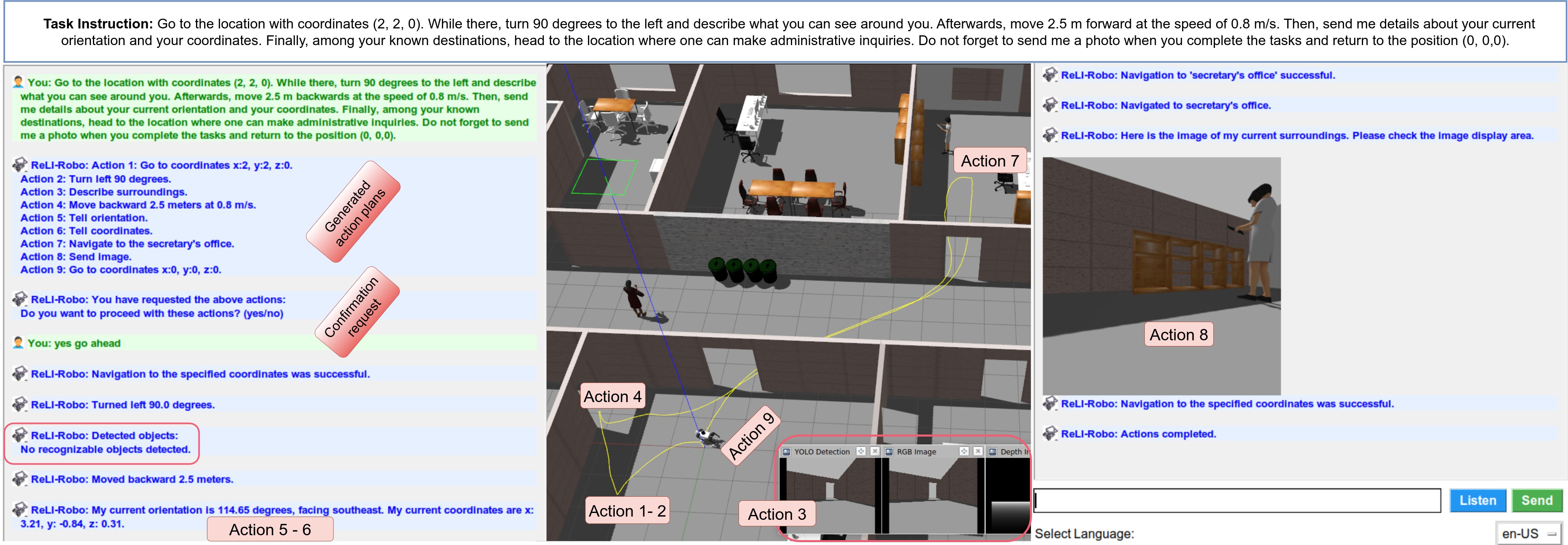}
        \caption{Long-horizon task instruction. In the action $a_3$, no objects are visible in the camera's FoV, as shown in the areas highlighted in red. The robot accurately reported that, as shown at the interaction interface.}
        \hfill
    \end{subfigure}
   \begin{subfigure}[b]{1.0\textwidth}
        \includegraphics[width=\linewidth]{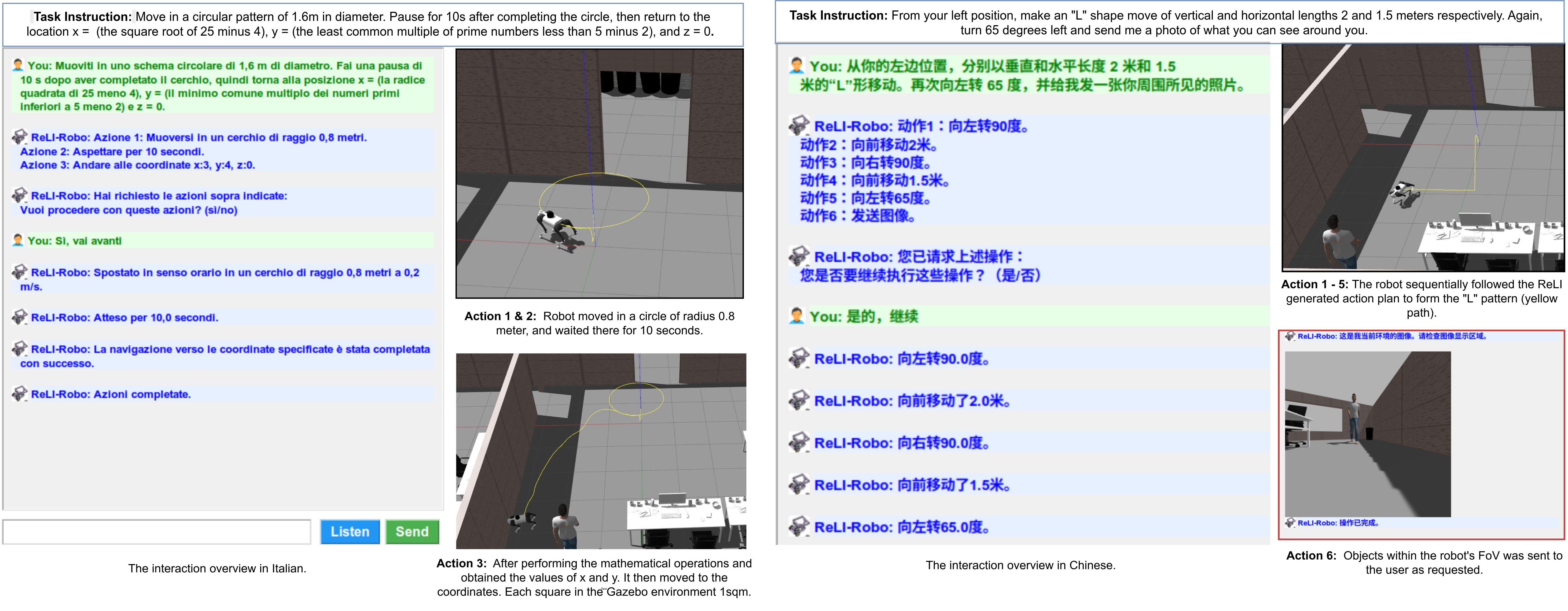}
        \caption{Geometric reasoning task in Italian and Chinese. These scenarios also validate ReLI’s numeric reasoning across languages.}
    \end{subfigure}
    \caption{Examples of long-horizon and geometry-oriented instructions. Example (a) shows a long-horizon task in which the robot reports that no objects are visible in the camera field of view for one action step. (b) shows geometric and numeric reasoning in Italian and Chinese, including patterned movement and multilingual parameter parsing.}
    \label{fig:long-horizon}
\end{figure*}

    We collected qualitative examples of ReLI's parsed instructions alongside the corresponding action execution in diverse languages. Figs.~\ref{fig:hrlChat} and~\ref{fig:lrl-vulChat} show representative multi-turn interactions in high-resource, low-resource, and vulnerable-language settings.
    Besides the multilingual, semantic, contextual, and descriptive reasoning abilities, ReLI can generalise to other advanced and complex reasoning tasks.  For instance, accomplishing some of the user's instructions in Table~\ref{tab:task_primitives} requires a high-level understanding of arithmetic reasoning, geometric motion, conditional branching, unit conversion, and long-horizon action decomposition. Refer to Fig.~\ref{fig:rationale} and~\ref{fig:long-horizon} for further examples.
    
\subsection{Human Raters and Demographics}
    As described in Sections~\ref{subsec4B} and~\ref{sec:qualitative}, we recruited external human raters to evaluate ReLI during real-world interaction. Table~\ref{tab:raters-demong} summarises the number of raters per language, the number of contributed instructions, and the resulting IPA and TSR for those contributed instructions. The contributed commands covered the same task categories used in the benchmark, including goal-directed navigation, low-level movement, scene queries, object-referenced commands, and contextual reasoning.

\begin{table}[htp]
    \centering
    \captionsetup{font=small}
    \caption{Human-rater participation by language, number of contributed instructions, and corresponding ReLI performance on those contributed commands. \(P_x\) denotes the number of fluent raters for the language; Cont.Instr. denotes contributed task instructions; Cont.IPA and Cont.TSR denote instruction parsing accuracy and task success rate measured on the contributed commands.}
    \label{tab:raters-demong}
    \scriptsize
    \begin{tabular}{@{}c c c c c@{}}
    \toprule
    \textbf{Raters} & \textbf{Language} & \textbf{Cont.Instr.} & \textbf{Cont.IPA (\%)} & \textbf{Cont.TSR (\%)} \\
    \midrule
    P$_1$   & Arabic      & 11 & 98.1 & 98.0 \\
    P$_9$   & English     & 69 & 100.0 & 99.9 \\
    P$_6$   & German      & 47 & 97.9 & 97.9 \\
    P$_1$   & Greek       & 12 & 95.8 & 95.7 \\
    P$_5$   & Hindi       & 52 & 94.4 & 94.3 \\
    P$_1$   & Igbo        & 23 & 92.3 & 92.3 \\
    P$_1$   & Italian     & 8 & 97.0 & 96.8 \\
    P$_1$   & Malay       & 15 & 97.1 & 96.9 \\
    P$_2$   & Chinese (Mandarin) & 25 & 98.0 & 97.8 \\
    P$_1$   & Nigerian Pidgin  & 29 & 98.6 & 98.5 \\
    P$_2$   & Spanish     & 28 & 99.3 & 99.2 \\
    P$_1$   & Turkish     & 16 & 96.9 & 96.7 \\
    P$_1$   & Yoruba      & 10 & 90.0 & 89.8 \\
    P$_1$   & Kannada      & 14 & 88.0 & 87.8 \\
    P$_1$   & Persian      & 17 & 82.4 & 82.3 \\
\bottomrule
\end{tabular} 
\end{table}

\section{Task Instructions and Rationales}\label{append:c}
    Table~\ref{tab:task_primitives} lists representative task instructions used in the benchmark. The instructions were designed to test ReLI's ability to parse multilingual free-form commands into robot-action primitives, preserve numerical parameters, ground object references, execute spatial navigation, and respond to query-oriented requests. The task set includes arithmetic expressions, coordinate goals, timing constraints, object-confidence thresholds, conditional branches, user-defined stopping conditions, geometric trajectories, and contextual destination references.

    Each of the $140$ selected languages was evaluated using logged interation trials based on the five task categories defined in Section~\ref{subsec4B}: zero-shot spatial or goal-directed navigation \((G_n)\), low-level motion control \((W_c)\), query-oriented interaction \((Q_i)\), object-referenced navigation \((O_n)\), and contextual or descriptive reasoning \((C_r)\). Short-horizon tasks typically contain one or two action primitives, whereas long-horizon tasks require multi-step decomposition, parameter extraction, state-dependent decisions, or explicit user confirmation before execution.

{\scriptsize
\begin{longtable}{@{}p{12cm} c c@{}}
\captionsetup{font=small}
\caption{Representative task instructions used for ReLI benchmarking. These task instructions tests ReLI's multilingual parsing, action decomposition, numerical-parameter preservation, spatial navigation, object grounding, conditional logic, and sensor-based query handling.}
\label{tab:task_primitives}\\
\toprule
\textbf{User Instructions} & \textbf{Categories} & \textbf{Horizon}\\
\midrule
\textsuperscript{1}\textbf{Task:} 
``Go to the destination with coordinates: $x$ = (the square root of 16 minus 1), $y$ = (the least common multiple of prime numbers less than 5), and $z$ = 0. While there, rotate 90 degrees to the left. Afterwards, describe the objects you can detect in front of you.''
\newline
\textbf{Rationale:}
We combine minor arithmetic reasoning (square root, number theory) and partial environment query. This tests if our framework can parse numeric expressions in various languages. In this approach, we verify the correctness in both coordinate-based navigation and object description steps.
& $G_n$, $W_c$, $Q_i$, $C_r$ & Long \\
\midrule
\textsuperscript{2}\textbf{Task:} 
``Head to the location with coordinates $(2, 0, 0)$. Stay there for 5 seconds, then circle around a 2-meter radius at 0.4\,m/s. If you detect any object with probability $\geq 80\%$, stop and send me an image.''
\newline
\textbf{Rationale:}
Here, we test the handling of coordinate-based targets, timed waiting, arc/circular motion, and object probability thresholds. Stress-test command parsing and dynamic detection for multiple languages.
& $G_n$, $W_c$, $O_n$ & Long \\
\midrule
\textsuperscript{3}\textbf{Task:}
``From your current position, calculate how many seconds it would take to reach the location $(4, -3, 0)$ if you travel at $1.0$ m/s. If it's over 15 seconds, stop and send me a photo of your surroundings; otherwise, proceed there and describe the nearest object.''
\newline
\textbf{Rationale:}
This task involves numeric logic (time calculation), conditional branching, sensor-based queries, and object references. We test ReLI's multilingual reasoning for maths plus environment-based inspection.
& $Q_i$, $C_r$, $G_n$ & Long \\
\midrule
\textsuperscript{4}\textbf{Task:}
``Perform a backwards movement of 2 meters at 0.2 m/s. While reversing, pause if you detect any obstacle closer than 0.5\,m, and describe it. Then resume until you reach 2\,m total.''
\newline
\textbf{Rationale:}
Checks partial path interruptions, user-defined distance thresholds, and object detection mid-motion. We test whether ReLI can handle sensor feedback and dynamic speed constraints in multiple languages.
& $W_c$, $Q_i$, $O_n$ & Long \\
\midrule
\textsuperscript{5}\textbf{Task:}
``Go to the location $(2,2,0)$, wait 10 seconds, then make an ``L-shape'' path of 3\,m horizontal and 2\,m vertical. Afterwards, navigate towards any detected fire extinguisher.''
\newline
\textbf{Rationale:}
Combines coordinate-based navigation, timed waiting, path drawing, and object-based motion. We validate ReLI's capacity to handle multi-step instructions and multiple movement forms.
& $G_n$, $W_c$, $O_n$ & Long \\
\midrule
\textsuperscript{6}\textbf{Task:}
``Send me your current orientation and coordinates. Next, rotate a full 360 degrees at 0.3\,m/s in place. If you see anything labelled ``chair,'' move forward 1 meter toward it.''
\newline
\textbf{Rationale:}
Here we test orientation \& coordinates queries, rotational actions, and partial object-based navigation.
& $W_c$, $O_n$, $Q_i$ & Long \\
\midrule
\textsuperscript{7}\textbf{Task:}
``Convert 500 centimetres into meters, then move that distance forward at 0.25 m/s. If you detect any ``person,'' send me a photo. Otherwise, rotate 90 degrees left and describe the surroundings.''
\newline
\textbf{Rationale:}
We explicitly test SI unit conversion (cm to m) plus object detection referencing.
& $W_c$, $Q_i$ & Long \\
\midrule 
\textsuperscript{8}\textbf{Task:}
``Head to your ``charging station'' located at (0,0,0). Remain there for 10 seconds, then return to where one can attend to personal hygiene needs among your known destinations. If no such destination exists, head to where one can cook food.''
\newline
\textbf{Rationale:}
Tests named-destination navigation (charging station, toilet, and kitchen) and fallback queries for unknown site references. This confirms that our framework can enable robots to handle environmental knowledge based on context.
& $G_n$, $Q_i$ & Long \\
\midrule
\textsuperscript{9}\textbf{Task:}
``Go to the ``Secretary's office.'' Once there, measure how many meters you have travelled from your start. Then take a snapshot. If the distance exceeds 5 meters, slow your speed to half of your maximum speed for the subsequent tasks.''
\newline
\textbf{Rationale:}
Verifies named location navigation, distance measurement, and dynamic speed changes. This tests the usefulness of our framework for large indoor environments with labelled destinations.
& $G_n$, $W_c$, $Q_i$ & Long \\
\midrule
\textsuperscript{10}\textbf{Task:}
``Drive forward at 0.5\,m/s until you've covered 3 meters, then pause for 10 seconds. Describe your surroundings.''
\newline
\textbf{Rationale:}
This task focuses on straightforward motion with an interruption clause for safety checks in an uncertain environment.
& $W_c$ & Short \\
\midrule
\textsuperscript{11}\textbf{Task:}
``I want you to identify any high-probability object in your camera feed. Then rotate to face it, and describe how far away it is from you in meters.''
\newline
\textbf{Rationale:}
Tests object-detection thresholding, orientation alignment, and distance reporting. Emphasises robust environment queries across multiple languages.
& $O_n$, $Q_i$ & Short \\
\midrule
\textsuperscript{12}\textbf{Task:}
``Calculate if your path from $(0,0)$ to $(5,5)$ at 1\,m/s will take more than 10 seconds. If yes, just return to $(0,0,0)$ and send an image. Otherwise, proceed and rotate 180 degrees upon arrival.''
\newline
\textbf{Rationale:}
Uses conditional logic, numeric comparisons, and image responses. Here, we assess our framework's capacity for minimal arithmetic in multiple linguistic forms.
& $Q_i$, $G_n$ & Long \\
\midrule
\textsuperscript{N}\textbf{Task:}
 $\cdots$ & $\cdots$ & $\cdots$\\
\bottomrule
\multicolumn{3}{@{}p{\linewidth}@{}}{
  \textcolor{blue}{\textbf{Legend:}}
  \textbf{\(G_n\)} \(\rightarrow\) zero-shot spatial or goal-directed navigation.
  \textbf{\(W_c\)} \(\rightarrow\) low-level movement commands without map-based target selection.
  \textbf{\(Q_i\)} \(\rightarrow\) information requests, causal queries, and sensor-based reporting.
  \textbf{\(O_n\)} \(\rightarrow\) visuo-spatial object-referenced navigation.
  \textbf{\(C_r\)} \(\rightarrow\) contextual or descriptive reasoning.
}
\end{longtable}

\section{Limitations}
    Although ReLI demonstrates strong cross-lingual language-to-action grounding across diverse languages and tasks, it is not without limitations. We acknowledge that ReLI relies on large-scale pre-trained LLMs~\cite{openai2023reasoning,Touvron2023LLaMAOA,liu2024deepseek, team2023gemini} and vision-language models~\cite{Radford2021LearningTV, carion2025sam} for instruction interpretation, semantic reasoning, and visuo-lingual grounding, and consequently, its performance is bounded by the capabilities of these orchestrating pre-trained models. In other words, it inherits the models' known limitations, including sensitivity to prompt formulation, uneven language coverage, and occasional hallucination or inconsistency in generated action sequences~\cite{10.1145/3703155,10569238}. However, such errors are mostly relevant for atomic actions that can be executed without explicit user confirmation. In our current implementation, the user confirmation mechanism reduces this risk for structured and long-horizon tasks.

    Secondly, our present benchmark covers $140$ representative languages, including high-resource, low-resource, creole, and vulnerable-language settings. Languages that are poorly represented in the training data of current LLMs, lack stable orthographies, or have limited ASR support may reduce ReLI's ability to infer task semantics and generate correct action primitives. This is most common with the speech-based interaction approach, where the ASR quality strongly affects downstream parsing. Code-mixed speech, accents, background noise, reverberation, and microphone quality can introduce transcription errors that propagate into the action decision pipeline. Although ReLI supports fallback language selection and text-based input, robust deployment in acoustically dynamic environments will require tighter integration with accent-robust and noise-adaptive ASR models~\cite{noisecanc,qian2016very,najafian2020automatic}.

    Finally, ReLI's current deployment depends partly on cloud-hosted LLM inference via API. This introduces latency, cost, service-availability, and network-dependence constraints that are undesirable for time-sensitive robotic applications. In highly dynamic scenarios, such as search and rescue, rapid human-robot teaming, or safety-critical navigation, network interruptions or high inference latency may degrade responsiveness. Further, ReLI's robustness also depends on the quality of onboard perception: low illumination, occlusion, motion blur, depth-sensor noise, and visually similar objects can reduce object-reference grounding accuracy. We therefore reserve these limitations for future work, which we aim to investigate through local model distillation, open-weight LLM/VLM backends, and adaptive ASR.

\end{document}